%% file: iclr2027_conference.tex
\documentclass{article} 
\usepackage{iclr2027_conference,times}

\input{math_commands.tex}

\input{preamble}

\usepackage{hyperref}
\usepackage{url}

\title{Hide and Seek: Hide Concepts to Seek Explanations for Black Box Models}

\author{Bhanu Tokas  \\
Arizona State University\\
\texttt{\{btokas\}@asu.edu} \\
\And
Rahul Nair \\
Arizona State University \\
\texttt{\{rnair21\}@asu.edu} \\
\And
Rockwell Jackson \\
Arizona State University \\
\texttt{\{rpjacks2\}@asu.edu} \\
\And
Hannah Kerner \\
Arizona State University \\
\texttt{\{hkerner\}@asu.edu} \\
}

\iclrfinalcopy 
\begin{document}

\maketitle

\input{sections/00_abstract}
\input{sections/01_intro}

\input{sections/02_related_works}
\input{sections/03_methodology}

\input{sections/04_experiments}

\input{sections/05_application}

\input{sections/06_discussion}
\input{sections/07_conclusion}

\subsection*{AI use statement}

In this work, we used generative AI tools to implement methods. We have not used generative AI tools to generate synthetic data sets, help develop theoretical models or conceptual frameworks, formulate mathematical claims, provide critical ingredients for proving mathematical claims, assist in the writing of proofs, propose or refine hypotheses, design or provide feedback on research methodology or experiments, assist with translation, clean and reformat dataset, support qualitative and thematic data analysis, or interpret results.

Additionally, we used generative AI tools to create or edit software code, create artifacts, summarize or analyze existing literature, brainstorming, sourcing/searching for information, edit a research paper to improve readability, and propose a title or keywords for a research paper. 

We have reviewed all AI-assisted work. LLM-generated code and text were verified and tested for correctness by the authors. Furthermore, we verified all existing method details from the actual paper. We take responsibility for the final content of this work, including text, claims or artifacts produced with the aid of generative AI.

\subsection*{Reproducibility statement}

We have provided the code for our method and experiments in supplementary materials. We provide parameter details and ablations in Section \ref{section:methods} and Appendix \ref{appendix:component_ablation}.

\bibliography{iclr2027_conference}
\bibliographystyle{iclr2027_conference}

\input{sections/08_appendix}

\end{document}

%% file: math_commands.tex
\usepackage{amsmath,amsfonts,bm}

\def\eqref#1{equation~\ref{#1}}

\def\1{\bm{1}}

\def\vv{{\bm{v}}}

\DeclareMathAlphabet{\mathsfit}{\encodingdefault}{\sfdefault}{m}{sl}
\SetMathAlphabet{\mathsfit}{bold}{\encodingdefault}{\sfdefault}{bx}{n}

\DeclareMathOperator*{\argmax}{arg\,max}

%% file: preamble.tex
\usepackage{array}
\usepackage{pifont}
\usepackage{comment}
\usepackage{tabularx}
\usepackage{graphicx}
\usepackage{makecell}
\usepackage{multirow}
\usepackage{amsmath}
\usepackage{subcaption}
\usepackage{booktabs}

\newcommand{\xmark}{\ding{55}}
\newcommand{\cmark}{\ding{51}}
\newcommand*\colourcheck[1]{%
  \expandafter\newcommand\csname #1check\endcsname{\textcolor{#1}{\cmark}}%
}
\newcommand*\colourx[1]{%
  \expandafter\newcommand\csname #1x\endcsname{\textcolor{#1}{\xmark}}%
}
\colourcheck{green}
\colourx{red}
\colourcheck{teal}

\newcolumntype{Y}{>{\centering\arraybackslash}X}
\newcolumntype{P}[1]{>{\centering\arraybackslash}p{#1}}
\newcolumntype{L}[1]{>{\raggedright\arraybackslash}p{#1}}

%% file: sections/00_abstract.tex
\begin{abstract}
    Interpreting machine learning model predictions is a central challenge, particularly for black-box vision models. Concept-based attribution methods interpret models using human-understandable concepts. Existing concept-based methods can only interpret classification and regression models. They require access to model weights or large concept-annotated datasets, and most methods cannot provide image-level attribution. We propose Hide and Seek (\textit{HnS}), a new method for concept-level attribution that does not require access to model weights or labeled concept datasets. \textit{HnS} can interpret a wide range of vision models. Given a concept and image set, \textit{HnS} retrieves concept-aligned images using a vision-language model, localizes the concept within each image, masks the localized region, and measures attribution from the change in the black-box model's output between the original and masked images. This enables faithful attribution by directly measuring the model’s behavior while also enabling local attributions. We validated \textit{HnS} on CelebA and COCO-Stuff. Compared to existing methods, \textit{HnS} showed better agreement with ground truth attributions. We demonstrated the applicability of \textit{HnS} on diverse tasks, including image captioning and satellite image segmentation, where no existing concept attribution method can be used. We provided our anonymous code in the supplementary material.
\end{abstract}

%% file: sections/01_intro.tex
\section{Introduction}

Interpreting why a model made a certain prediction is important for deploying it responsibly, especially in high-stakes settings such as medicine or hiring. Concept attribution is a popular approach for interpreting the behavior of vision models ~\citep{TCAV, ConceptSHAP, PACE, conceptWhiten, PCBM-h}. It measures how much a human understandable concept (e.g., smiling, long hair) contributed to a model's prediction. Suppose we have a classification model $M$ that predicts whether an image contains a male or a female, and $M$ showed a high attribution score for the concept ``smiling'' for male images. This means that $M$ relies on smiling-related features to predict males. This insight is valuable for model refinement, since we know $M$ has learned to rely on a spurious feature in the dataset to classify males. 

To be practically and widely useful, a concept attribution method should be applicable to black-box models (no access to model weights), not only for white-box models (models with publicly accessible weights). Second, it should measure both global attribution (the overall attribution of a concept across all images in a dataset) and local attribution (the attribution of a concept for a single image). Third, it should not require a labeled concept dataset, since curating such datasets is expensive and impractical to do for every application. Fourth, it should be able to interpret a wide range of vision models. No existing method meets all of these criteria.

\begin{figure*}[htbp]
    \vspace{-0.4cm}
    \centering
    \includegraphics[width=\textwidth]{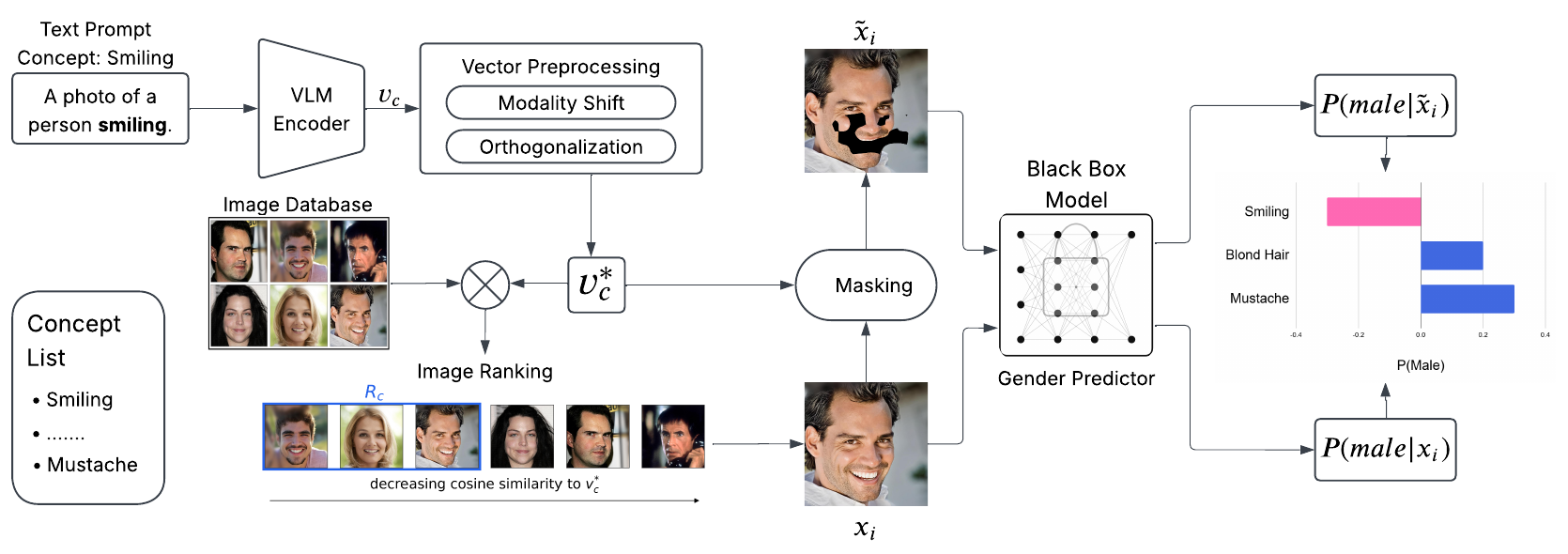}
    \caption{
    Overview of Hide and Seek (\textit{HnS}). For a concept $c$ (smiling), we encode its text prompt (``photo of a person smiling'') into a text vector $v_c$ using SigLIP (a vision-language model). We pre-process $v_c$ (using modality correction and orthogonalization) to obtain $v_c^*$. From an image database, we select a set of images $R_c$ that show the highest cosine similarity with $v_c^*$. For each image $x_i$ in $R_c$, \textit{HnS} identifies the region $c$ in $x_i$ and hides this region to create a masked image $\tilde{x}_i$. We pass $x_i$ and $\tilde{x}_i$ to our model $M$ to obtain predicted probabilities for a class $y$ (male): $P(x_i)$ and $P(\tilde{x}_i)$, respectively. We measure attribution for the concept $c$ (on class $y$) as $P(x_i)$ - $P(\tilde{x}_i)$.}


    \vspace{-0.4cm}
    \label{fig:conceptMask_teaser}
\end{figure*}

TCAV~\citep{TCAV} was one of the first methods to measure concept attribution. Several subsequent works developed extended versions of TCAV ~\citep{ConceptSHAP, PACE, conceptWhiten}. TCAV and its derivatives only work for white-box models, as they require access to model weights. More recent attribution methods, such as Post-hoc Concept Bottleneck Models (PCBMs)~\citep{PCBM-h}, work for both white-box and black-box models. However, PCBMs require labeled concept datasets, measure only global attribution, and can only interpret classification and regression models.   
Motivated by these gaps in prior work, we present a new concept attribution method called Hide and Seek or \textit{HnS} (Figure \ref{fig:conceptMask_teaser}).

To understand how Hide and Seek (\textit{HnS}) works, say we want to compute the attribution of the concept ``smiling'' for male predictions, for model $M$ (discussed in our example earlier). We use a vision-language model (VLM) to retrieve images that best represent the concept (smiling faces). In each of these images, \textit{HnS} automatically identifies the region where the target concept is present (mouth region). It creates a corresponding masked image where that concept is hidden. To compute attribution, \textit{HnS} passes the actual image (a smiling face) and its masked version (mouth region hidden) to the model $M$. We define concept attribution as the difference in the predicted probability (of the male class) between the original and the masked image. A high attribution score indicates that $M$ relies on smiling as a cue to predict males.

\textit{HnS} does not need a labeled concept dataset, since it uses a VLM to identify concepts. Since \textit{HnS} uses a model's predicted probability to measure attribution (only forward-pass outputs, not internal representations), it works with both white-box and black-box models. \textit{HnS} measures both global (overall) and local (image-level) attribution. It can interpret a wide range of vision models (e.g., classification, regression, segmentation, captioning). It is also the fastest method to measure attributions (refer to Appendix \ref{appendix:Computational_Cost}).

We validated \textit{HnS} on two distinct types of datasets: CelebA (a dataset of celebrity faces) and COCO-Stuff (an object- and scene-based dataset). \textit{HnS} achieved the highest correlation with ground-truth attributions compared to previous methods such as TCAV and PCBM. 
We also tested \textit{HnS} in real-world use cases such as image captioning and field boundary segmentation, where existing methods cannot be applied directly. We found that several state-of-the-art captioning models had gender bias (e.g., models generated male-centric captions when images had a tie or a toaster, and female-centric captions when images had a cake or a hair dryer). Using \textit{HnS}, we also investigated why PRUE is the current state-of-the-art model to segment crop fields \citep{muhawenayo2026pruepracticalrecipefield}. We found that PRUE has not only learned to identify concepts useful for segmenting cropland (e.g., arable land) but also concepts irrelevant to cropland (e.g., urban areas). 

Our contributions are as follows: (1) We propose \textit{HnS}, the first concept attribution method that can interpret any vision model, without a labeled concept dataset. (2) Compared to previous methods (TCAV, PCBM), we show that \textit{HnS} has the best correlation with ground-truth attributions on facial datasets (CelebA) and object-based datasets (COCO-Stuff). (3) We applied \textit{HnS}s on real-world vision tasks like image captioning and field boundary segmentation to show how our method can be used to interpret a wide variety of vision models.

%% file: sections/02_related_works.tex
\section{Related Works}

Concept-based interpretability methods aim to explain model predictions in terms of human-understandable concepts rather than low-level features like pixels. Previous works proposing concept attribution techniques can be organized into methods for whitebox models (in which the model's internals are accessible), interpretable concept models, and methods for black-box models.

\textbf{Concept attribution for whitebox models:} 
\citet{TCAV} introduced TCAV (Testing with Concept Activation Vectors), the first method for concept-level attributions. They start by finding vectors that represent a concept in a model's internal representations. These vectors are known as Concept Activation Vectors (CAV). The method works by finding the sensitivity of the output with respect to the CAV. This sensitivity quantifies the influence of the concept on the model's prediction.

Further work by \citet{PACE} and \citet{conceptGradient} improved the original TCAV method, making it more robust and better suited for local attribution. Works like \citet{ConceptSHAP} and \citet{ConeSHAP} combine the ideas of concept-level attribution with Shapley values. These methods have two major practical limitations: they do not work for black-box models, since they require access to the model's internal representation, and they require labeled concept datasets.

\textbf{Interpretable concept models:} Building on TCAV, \cite{conceptWhiten} introduced ``Concept Whitening", a method for training interpretable models by aligning internal representations with disentangled individual concepts. \citet{CBM} later proposed Concept Bottleneck Models (CBMs), which decompose prediction into concept prediction followed by target prediction as a linear function of the predicted concepts. Subsequent work, including \citet{ProbabilisticCBMs, StochasticCBMs}, improved the reliability and robustness of CBMs. However, these methods train interpretable models rather than explain existing ones.

\textbf{Concept attribution for black box models:} PCBM~\citep{PCBM-h} works by training a surrogate CBM to mimic the black box predictions. The final-layer weights of the surrogate CBM are considered concept attributions. It leverages CLIP~\citep{CLIP} to construct concept classifiers from text, avoiding concept annotations. However, CBMs may not faithfully approximate the black-box model and cannot provide local attributions~\citep{tapli2026rethinkingconceptbottleneckmodels}. Also, CBMs are designed for classification and regression. They cannot be used to explain models for tasks like segmentation or captioning.

In contrast to prior work, we propose Hide and Seek (\textit{HnS}), a concept attribution method that directly attributes predictions of black-box models without surrogate approximation, does not require concept-labeled datasets, captures local attributions, and supports a broad class of tasks. Table \ref{tab:concept_attribution_comparison} shows a comparison of existing concept attribution techniques with \textit{HnS}.

\begin{table*}[htbp]
\centering
\scriptsize
\vspace{-0.2cm}
\caption{Comparison of Concept Attribution Techniques}
\vspace{-0.3cm}
\label{tab:concept_attribution_comparison}
\begin{tabularx}{0.95\linewidth}{L{3.5cm}YYYYY}
\toprule
\textbf{Method} & \textbf{Black Box} & \textbf{Concept Label-Free} & \textbf{Surrogate Free} & \textbf{Local Attribution} & \textbf{Wide Task Support} \\
\midrule
TCAV~\citep{TCAV}    & \redx & \redx  & \tealcheck & \tealcheck      & \redx \\
ConceptSHAP~\citep{ConceptSHAP} & \redx & \redx  & \tealcheck & \tealcheck  & \redx \\
PACE~\citep{PACE} & \redx & \redx  & \tealcheck & \tealcheck  & \redx \\
PCBM~\citep{PCBM-h}  & \tealcheck   & \tealcheck & \redx      & \redx & \redx \\
Hide and Seek (ours)    & \tealcheck & \tealcheck & \tealcheck      & \tealcheck & \tealcheck      \\
\bottomrule
\end{tabularx}
\vspace{-0.4cm}
\end{table*}

%% file: sections/03_methodology.tex
\section{HIDE AND SEEK}
\label{section:methods}

\subsection{Problem Setup}

Consider a classifier $M$, which accepts an input image and outputs a predicted probability for a class $y$ (e.g., male). Consider a concept $c$ described using natural-language text (e.g., smiling). We measure concept attribution or $\mathrm{Attr}(c)$: how much does the model $M$ rely on concept $c$ to predict class $y$? We measure attribution at two granularities: global and local $\mathrm{Attr}(c)$. Global $\mathrm{Attr}(c)$ is the overall influence of concept $c$ on $y$ across all images in a dataset. Local $\mathrm{Attr}(c)$ is the influence of $c$ on $y$ for a particular image. 

\subsection{Hide and Seek Methodology}
\label{subsec:concept-vector}

To measure attribution for concept $c$, we extract its text representation vector $\vv_c$ using the vision-language model SigLIP. To ensure that $\vv_c$ accurately represents concept $c$, we perform additional pre-processing steps on $\vv_c$ (modality correction and orthogonalization) to get a processed vector $\vv^{*}_c$. We discussed these pre-processing steps in Section \ref{subsec:pre_processing}. Hide and Seek (\textit{HnS}) works with any VLM. With SigLIP, \textit{HnS} showed the best correlation with ground-truth attributions. Refer to our ablation with different VLMs in Appendix \ref{appendix:subsec_encoder_ablations}. 

We identify a set of images $R_c$ that best represent $c$. To extract $R_c$, we use an image database (e.g., a database of human faces). We compute cosine similarity between $\vv^{*}_c$ and each image from the database. We define $R_c$ as the set of the top $K$ images with the highest cosine similarity to $\vv^{*}_c$. For our experiments, we set $K = 50$. Refer to our ablation with different $K$ values in Appendix \ref{appendix:subsec_k_ablation}.

To compute the influence (or attribution) of concept $c$ on class $y$, we assess how model $M$ predicts $y$ when $c$ is visible in an image vs. when $c$ is hidden. Suppose the predicted probability of $y$ changes when $c$ is present compared to being hidden in the image. This means that concept $c$ has a strong influence on class $y$: the attribution of $c$ is high for $y$. If the predicted probability of $y$ changes only minimally, $c$ does not influence $y$: the attribution of $c$ to $y$ is low.

For each image in $R_c$, we create two versions: the original image $x$ and its masked version $\tilde{x}$, in which we hide concept $c$ in $x$. To create $\tilde{x}$, we first identify the region in $x$ where $c$ is present. We split the image $x$ into $L$ square patches using SigLIP's encoder. We compute an embedding vector for each patch. We compute cosine similarity between $\vv^{*}_c$ and the embedding vector of each of the $L$ patches. Concept $c$ is present in the patches that show high cosine similarity with $\vv^{*}_c$. To identify these patches, we use a cosine similarity threshold $\tau_c$.
\vspace{-0.1cm}
\begin{equation}
    \tau_c = \mu_c + \alpha\, \sigma_c
\end{equation}
We assume that $L$ cosine similarity scores (for $L$ patches) follow a Gaussian distribution, where $\mu_c$ and $\sigma_c$ are the mean and standard deviation, respectively, of these scores. We set $\alpha = 1.5$ (refer to our ablation with different $\alpha$ values in Appendix \ref{appendix:subsec_alpha_ablation}). To create $\tilde{x}$, we select patches in $x$ that show a cosine similarity $> \tau_c$. At $\alpha = 1.5$, around $7\%$ of the patches in $L$ show a cosine similarity $> \tau_c$.

To create the masked image $\tilde{x}$, we hide the selected patches using a patch filling strategy. We have $7$ filling strategies: blur, zero-fill, white-fill, mean-fill, hue-shift, zero-fill-noise, and noise-then-blur (refer to Appendix ~\ref{appendix:perturbation_methods} for a description of each strategy). We use only the strategy that best hides $c$. We offer a variety of strategies, as different strategies are useful for different types of images.

To select the best filling strategy for $\tilde{x}$, we use vector differences. Assume that a filling strategy produces a masked image that effectively hides $c$. If we subtract the image vector of $x$ (where $c$ is present) from the image vector of this masked image (where $c$ is absent), the resultant vector should predominantly have $c$'s information. The resultant vector should be closely aligned with $\vv^{*}_c$. We therefore select a strategy that maximizes the cosine similarity between the resultant vector and $\vv^{*}_c$.   
\vspace{-0.1cm}
\begin{equation}
    \tilde{x} = \argmax_{f \in \mathcal{F}} \left\langle
        \frac{\phi(x) - \phi(\tilde{x}_{f})}
             {\lVert \phi(x) - \phi(\tilde{x}_{f}) \rVert},
        \; \vv^{*}_{c} \right\rangle
\end{equation}
$\left\langle \right\rangle$ denotes the cosine similarity operation. $\phi(x)$ is the embedding vector of the original image, and $\phi(\tilde{x}_{f})$ is the embedding vector for its corresponding masked image ($f$ is a specific filling strategy). $\frac{\phi(x) - \phi(\tilde{x}_{f})} {\lVert \phi(x) - \phi(\tilde{x}_{f}) \rVert}$ is the resultant vector for strategy $f$. 
             
We pass images $x$ and $\tilde{x}$ to model $M$. We compute the predicted probability of class $y$ for images $x$ and $\tilde{x}$. We call these predicted probability values $M(x)$ and $M(\tilde{x})$, respectively. We define local attribution of $c$ on $y$ as the difference between these probabilities.
\vspace{-0.1cm}
\begin{equation}
    \label{eq:Attribution_local}
    \mathrm{Local\ Attr}(c) = M(x) - M(\tilde{x})
\end{equation}
To compute global attribution, we average local attribution values across all $K$ images in our set $R_C$. We exclude images for which the corresponding masked image is poor (e.g., the entire image is hidden), as they can affect global attribution scores.
\vspace{-0.1cm}
\begin{equation}
    \label{eq:Attribution_1}
    \mathrm{Global\ Attr}(c) = \frac{1}{K}\sum_{i}^{R_{c}}
        \Big[ M(x_i) - M(\tilde{x}_i) \Big]
\end{equation}
In rare cases, a concept can have noisy local attribution scores (either the concept vector created by SigLIP is poor, or the concept is absent in the image database). We must report a low global attribution score for such concepts, as we cannot reliably measure their influence on our model. To penalize noisy concepts, we perform a Cohen's d normalization step across all concepts. We compute $s_c$, which is the standard deviation across local attribution scores of $c$. We normalize the global attribution score with $s_c$.
\vspace{-0.3cm}
\begin{equation}
    \mathrm{Global\ Attr}(c) = \frac{\mathrm{Global\ Attr}(c)}{s_{c}}
\end{equation}
If local attribution scores of $c$ are noisy, $s_c$ will be large, making the normalized global attribution score small for that noisy concept.

To interpret a model with \textit{HnS}, users should compare the relative rankings of concepts rather than individual attribution values. For example, if a gender classifier assigns a higher global attribution to \textit{smiling} than to \textit{baldness} when predicting \textit{male}, it relies more strongly on smiling. For an image-level explanation, users should rank concepts by their local attribution score for that image. 

TCAV and PCBM define concept attribution as $\Delta y / \Delta c$: where \(\Delta y\) is the change in the model output induced by a change \(\Delta c\) in the concept. \textit{HnS} instead uses only $\Delta y$, as $\Delta c$ is hard to define in practical use cases. The way existing methods compute $\Delta c$ for a concept (e.g., cat) is unreliable. They use the $c$'s concept vector: one end of this vector points toward a positive set (cat-like images) and the other end toward a negative set (non-cat images). This makes $\Delta c$ unstable: if we change the positive or negative set, $c$'s concept vector shifts, and $\Delta c$ shifts along with it, even though $\Delta y$ (the model's actual output) has not changed. For stable attribution scores, measuring $\Delta y$ is sufficient.

\subsection{Pre-processing: Modality Correction and Orthogonalization}
\label{subsec:pre_processing}

\paragraph{Modality-shift correction.} We would expect an image vector of a concept and its corresponding text vector to be closely aligned. However, in VLMs, these vectors occupy different regions in the vector space \citep{levi2025double}. Due to this text-image modality gap, we cannot directly compare the text vector of $c$ ($\vv_c$) to its corresponding image vector. To account for this modality gap, we perform a modality shift correction using the following equation: 
\vspace{-0.15cm}
\begin{equation}
    \vv_{c}' = \vv_{c} - \mu_{\text{text}}, \qquad
    \mu_{\text{text}} = \frac{1}{|\mathcal{W}|}\sum_{c' \in \mathcal{R}} \vv_{c'},
\end{equation}
$\mu_{\text{text}}$ is the mean text embedding computed across a large, diverse word set $\mathcal{W}$ (this word set is independent of the concept set). We give intuition for this correction in Appendix \ref{appendix:Modality_Shift_Intuition}.

\paragraph{L\"owdin orthogonalization.} In VLMs, concepts can often be entangled. For instance, VLMs can give similar concept vectors for two different concepts like ``baldness'' and ``male'' (due to some inherent bias). To ensure we get a unique concept vector for each concept, we performed orthogonalization using the following equation:
\vspace{-0.2cm}
\begin{equation}
    V^{*} = V S^{-1/2},
\end{equation}
V is a matrix of all $\vv'_{c}$ stacked together. $S$ is the Gram Matrix, given by $S=V^{T}V$. $S$ expresses the overlap between the vectors in $V$. The matrix $V^*$ represents the orthogonalized vectors across all concepts. A single row $\vv^{*}_{c}$ represents the orthogonalized vector for the concept $c$. 

Depending on the VLM, these pre-processing steps may be unnecessary or may even introduce noise. In Appendix~\ref{appendix:subsec_corrections}, we performed ablations on these steps across multiple VLMs.

%% file: sections/04_experiments.tex
\section{Experiments and Results}

\subsection{Global Attributions compared to Ground-Truth}
\label{subsec:Masking_Experiment}

\subsubsection{Experiment Setup}
We compute concept attribution on two datasets: CelebA (a facial dataset) and COCO-stuff (an object-and scene dataset). We compared three attribution techniques: TCAV, PCBM, and Hide and Seek or \textit{HnS} (ours). We used three ImageNet-pretrained classification models: ViT-B-16 ~\citep{dosovitskiy2020image}, ConvNeXt-Tiny~\citep{liu2022convnet} and ResNet-18~\citep{resnet}.

We fine-tuned these models on both CelebA and COCO-Stuff. CelebA provides annotations for $40$ binary facial attributes (e.g., bald, blond hair). For CelebA, we used $2$ of these attributes (Attractive and Male) as classes and fine-tuned the models. From the $38$ remaining attributes, we used $26$ attributes as our concepts. For COCO-stuff, we fine-tuned models on a subset of $20$ classes \cite{singh2020don}. For COCO-stuff, we used $103$ concepts from Broden~\citep{Broden_filtered}. 



To compute ground truth attributions for a concept $c$ (against which each attribution method is compared), we follow the masking strategy of \citet{meister2023gender}. For each ground-truth image, we created two versions: the original image and a version in which concept $c$ is hidden from the image. We identified the region where $c$ is present in the ground truth and masked that region using a zero-fill strategy. To mask $c$ in the image, we used the segment annotations for our $26$ CelebA concepts, from the CelebA-HQ dataset~\citep{lee2020maskgan}. Because COCO-Stuff lacks annotations for Broden concepts, we generated segments for Broden concepts (on COCO-stuff images) using Grounded-SAM~\citep{ren2024grounded}. We used these generated segments as the ground truth segments. More details on ground truth generation are available in Appendices \ref{appendix:coco_stuff_ground_truth} and \ref{appendix:Concept_Retrieval}.

We passed the original ground-truth image and its masked version to the model. For a target class, we defined the ground-truth local attribution of $c$ as the difference in the predicted probability (of that class) between the ground-truth image and its masked version. To compute ground-truth global attribution of $c$ on a class, we averaged the local attributions across $100$ images in which $c$ is present. For each class-concept pair, we computed the Spearman rank correlation ($\rho$) between ground-truth global attributions and the global attributions provided by each of the three attribution methods. We computed $\rho$ across $52$ class-concept pairs in CelebA ($26$ concepts $\times$ $2$ classes) and $2060$ pairs in COCO-stuff ($103$ concepts $\times$ $20$ classes). 

Our approach to creating ground-truth attributions is similar to \textit{HnS}. One might argue that this similarity can inflate the correlation scores for \textit{HnS}. We addressed this concern with an additional experiment on global attribution in Appendix \ref{appendix:Shortcut_Experiment}. 

To create the representative set $R_c$ (for \textit{HnS}), we used CelebA's validation set as the image database for CelebA. We used COCO-Stuff's validation set as the image database for COCO-Stuff. To create the concept dataset for CelebA (for PCBM and TCAV), we used CelebA's training set. For COCO-Stuff, we used the Broden dataset as the concept dataset.

\begin{table*}[htbp]
    \centering
    \footnotesize
    \vspace{-0.1cm}
    \caption{For three models, we reported the correlation ($\rho$) between ground-truth global attribution scores and the global attribution scores from \textit{HnS}, TCAV, and PCBM. A higher $\rho$ value for a method indicates that it shows better agreement with ground-truth attributions.}
    \label{tab:masking_experiment}
    \vspace{-0.3cm}
    \begin{tabularx}{\textwidth}{L{3.1cm}YP{0.8cm}YYP{0.8cm}Y}
        \toprule
         & \multicolumn{3}{c}{CelebA} & \multicolumn{3}{c}{COCO-Stuff} \\
        \cmidrule(lr){2-4} \cmidrule(lr){5-7}
        Method & ResNet18 & ViT & ConvNeXt & ResNet18 & ViT & ConvNeXt \\
        \midrule
        TCAV & 0.319 & 0.213 & 0.192 & 0.030 & 0.085 & 0.117 \\
        PCBM (ResNet-18) & 0.180 & 0.164 & 0.258 & 0.058 & 0.018 & 0.059 \\
        PCBM (SigLIP) & 0.208 & 0.197 & 0.291 & 0.163 & 0.211 & 0.214 \\
        PCBM (CLIP-RN50) & 0.149 & 0.142 & 0.201 & 0.168 & 0.199 & 0.214 \\
        \textit{HnS} (ours) & \textbf{0.615} & \textbf{0.507} & \textbf{0.421} & \textbf{0.315} & \textbf{0.342} & \textbf{0.257} \\
        \bottomrule
    \end{tabularx}
    \vspace{-0.3cm}
\end{table*}

\subsubsection{Global Attribution Results}

\textbf{\textit{HnS} shows the best agreement with global ground truth attributions}. In Table \ref{tab:masking_experiment}, we reported the Spearman rank correlation for all attribution methods (across three models). \textit{HnS} showed the highest correlation across all models for both CelebA and COCO-Stuff. 

For \textit{HnS}, we used an image database of in-distribution images (CelebA's validation set for CelebA). In Appendix \ref{appendix:Dataset_Ablation}, we showed that \textit{HnS} provides accurate attributions for CelebA, even if we use an image database of out-of-distribution images (e.g., UTKFace~\citep{zhifei2017cvpr}.

\subsection{Local Attributions}
\label{subsec:local_attribution_experiment}

We visually examine whether concept attribution methods can retrieve sensible images for a class-concept pair. For image retrieval, we only used the two methods that provide local (image-level) attributions: \textit{HnS} and TCAV. We did not use PCBM as it only provides global attributions. 


\subsubsection{Experiment Setup} 

For CelebA, we selected the following class-concept pairs: \{class: attractive, concept: eyeglasses\}, and \{class: attractive, concept: blond hair\}. For COCO-stuff: \{class: apple, concept: black\}, and \{class: skateboard, concept: blurry\}. For each pair, we computed local attribution scores (using both \textit{HnS} and TCAV) for all images in which the class was present. For each attribution method, we sorted images (in both CelebA and COCO-stuff) from top to bottom by the magnitude of the local attribution scores. For each class-concept pair, we visually examined the top $3$ and bottom $3$ images retrieved by \textit{HnS} and TCAV.

\subsubsection{Local Attribution Results} 

\begin{figure}[htbp]
    \vspace{-0.2cm}
    \centering
    \begin{subfigure}[b]{0.48\linewidth}
        \includegraphics[width=\linewidth]{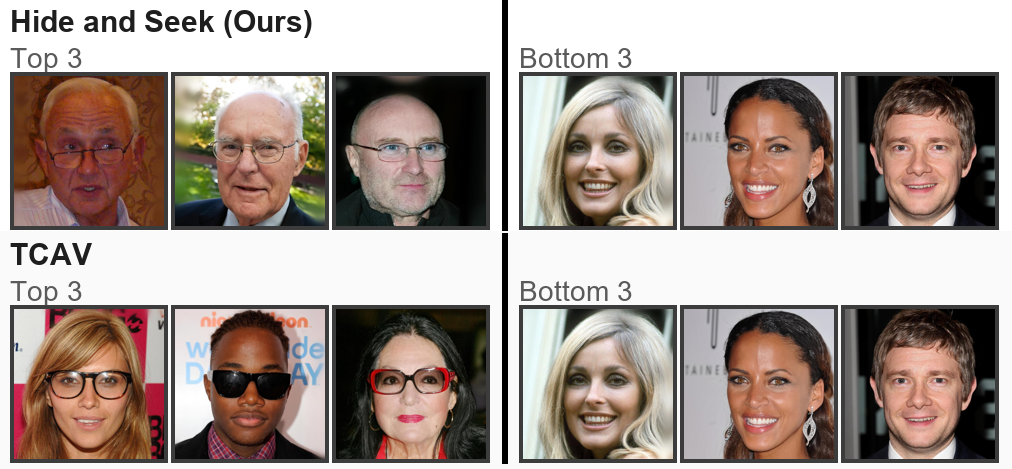}
        \caption{Class: Attractive, Concept: Eyeglasses}
        \label{fig:local_attribution_examples_a}
    \end{subfigure}
    \hfill
    \begin{subfigure}[b]{0.48\linewidth}
        \includegraphics[width=\linewidth]{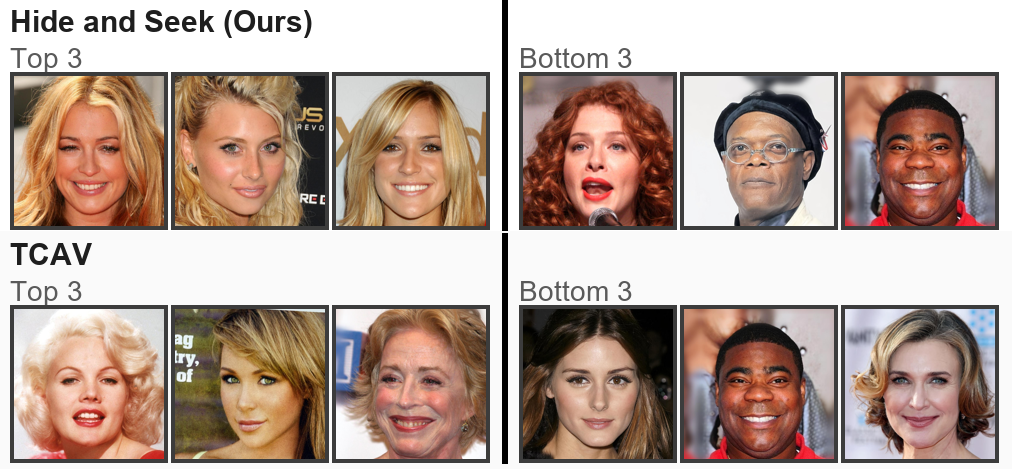}
        \caption{Class: Attractive, Concept: Blond Hair}
        \label{fig:local_attribution_examples_b}
    \end{subfigure}
    \vspace{0.2cm}
    \begin{subfigure}[b]{0.48\linewidth}
        \includegraphics[width=\linewidth]{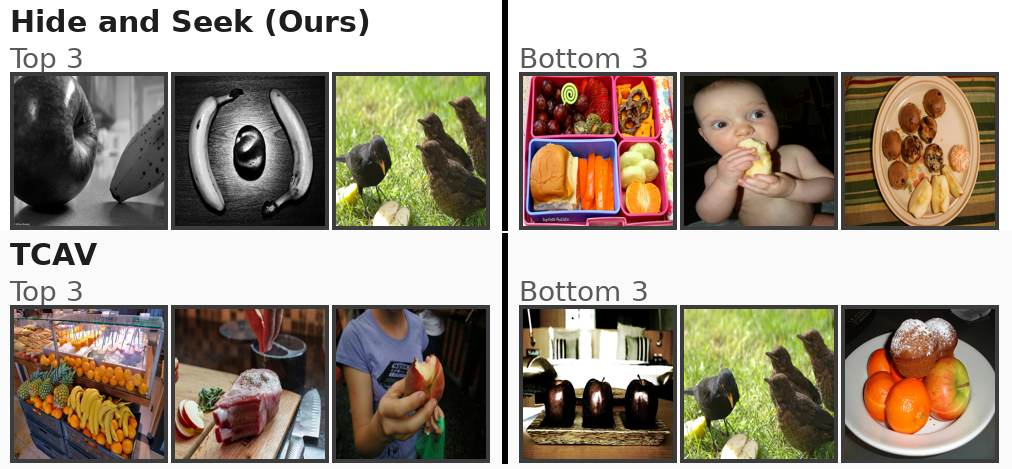}
        \caption{Class: Apple, Concept: Black}
        \label{fig:local_attribution_examples_coco_a}
    \end{subfigure}
    \hfill
    \begin{subfigure}[b]{0.48\linewidth}
        \includegraphics[width=\linewidth]{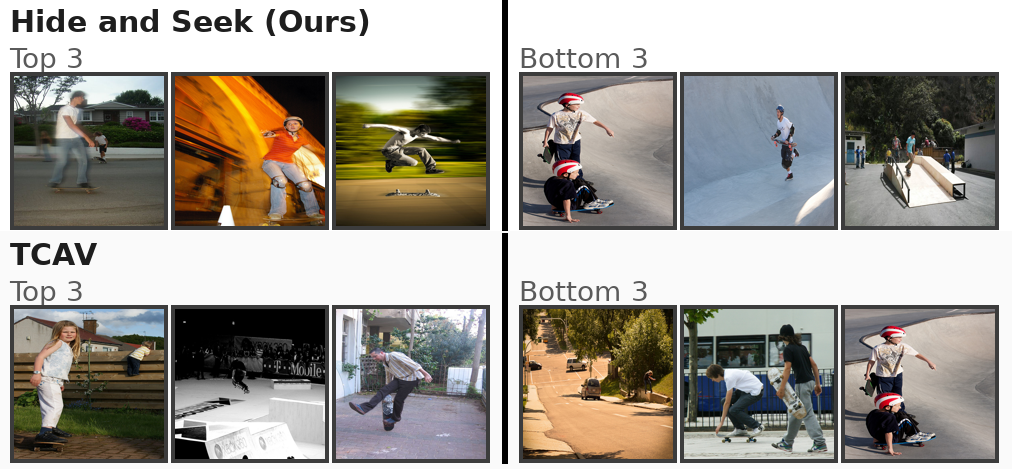}
        \caption{Class: Skateboard, Concept: Blurry}
        \label{fig:local_attribution_examples_coco_b}
    \end{subfigure}
    \vspace{-0.5cm}
    \caption{For each class-concept pair, we ranked images by magnitude of local attribution scores of Hide and Seek (\textit{HnS}) and TCAV. For each method, we visualized the top $3$ and bottom $3$ images. (a)-(b) are for CelebA, (c)-(d) are for COCO-Stuff. We show more examples in Appendix \ref{appendix:Local_Attribution}.
    }
    \label{fig:local_attribution_examples}
    \vspace{-0.3cm}
\end{figure}

In Figure \ref{fig:local_attribution_examples}, the top row shows the images retrieved by \textit{HnS}, while the bottom row shows images by TCAV. For both methods, the top $3$ images (for a class-concept pair) are in the left grid, while the bottom $3$ images are in the right grid.

\textbf{\textit{HnS} is better than TCAV at image retrieval, even though \textit{HnS} has no access to model weights}. In CelebA, for \{class: attractive, concept: eyeglasses\} (Figure \ref{fig:local_attribution_examples_a}), the top $3$ images of both methods had eyeglasses while the bottom $3$ images did not. For \{class: attractive, concept: blond hair\} (Figure \ref{fig:local_attribution_examples_b}), the top $3$ images of both methods had blond hair, while the bottom $3$ images did not. On CelebA, both methods retrieved sensible images. In COCO-Stuff, for \{class: apple, concept: black\} (Figure \ref{fig:local_attribution_examples_coco_a}), all the top $3$ images in \textit{HnS} had an apple and some black-colored elements. None of the top $3$ images in TCAV had any black elements. For \{class: skateboard, concept: blurry\} (Figure \ref{fig:local_attribution_examples_coco_b}), all top $3$ images in \textit{HnS} had a skateboard and visible motion blur. None of the top $3$ images in TCAV had a blurring effect. Only \textit{HnS} retrieved sensible images in COCO-Stuff. 

%% file: sections/05_application.tex
\section{Hide and Seek in Real World Applications}
\subsection{Image Captioning}
\label{subsec:image_captioning}

Previous studies~\citep{DBAC, tang2021mitigating} have shown that image captioning models exhibit gender bias. To identify gender bias, these studies required annotated datasets (in which gender and objects are annotated in the images). We evaluated whether \textit{HnS} can identify bias trends similar to those reported in prior work, without these annotations.

We computed concept attribution on $10$ captioning models: ViT-GPT2~\citep{vit_gpt2}, BLIP~\citep{li2022blip}, Florence-2~\citep{florence}, LLaVA~\citep{liu2023llava}, BakLLaVA~\citep{liu2023improvedllava}, GPT-4o~\citep{achiam2023gpt}, FC~\citep{rennie2017self}, Att2In~\citep{rennie2017self}, Transformer~\citep{Attention_is_all_you_need}, and UpDown~\citep{UpDn}. We evaluated these models on the COCO Captions dataset, which contains $10,780$ images of people (male, female) co-occurring with objects (e.g., tie, ball). 

Given an input image, each model generates a caption and the predicted probability of each word in the caption. In each caption, we identified male-centric words (e.g., he, him) and female-centric words (e.g., she, her). We used a pre-defined list of these words (see Appendix \ref{appendix:Captioning_Experiment} for the word list). We summed the predicted probabilities for the male-centric words ($\sum male$) and then for the female-centric words ($\sum female$). We define a gender score: $\sum male$ $-$ $\sum female$. For \textit{HnS}, we defined the attribution of a concept $c$ as the difference in the gender score when $c$ is present in the image versus when $c$ is hidden. We ignored captions that did not contain any male- or female-centric words. We used the $79$ object names in COCO captions as our concept list.

\begin{table}[htbp]
    \center
    \footnotesize
    \vspace{-0.2cm}
    \caption{Across all models, we identified concepts that frequently occurred in the list of top $20$ concepts with the highest positive attribution (male-centric concepts), and in the list of top $20$ concepts with the highest negative attribution (female-centric concepts). For instance, tie was a male-centric concept in $9$ out of $10$ models.}
    \vspace{-0.2cm}
    \label{tab:captioning_overlap}
    \begin{tabularx}{0.8\textwidth}{YYYY}
        \toprule
    \makecell{\textbf{Male-centric} \\ \textbf{concepts}} & \makecell{\textbf{Number of}\\ \textbf{models}} & \makecell{\textbf{Female-centric} \\ \textbf{concepts}} & \makecell{\textbf{Number of}\\ \textbf{models}}\\
        \midrule
        tie      & 9/10 & cake       & 7/10 \\
        toaster  & 8/10 & hair dryer & 6/10 \\
        backpack & 7/10 & teddy bear & 6/10 \\
        sandwich & 7/10 & carrot     & 6/10 \\
        bottle   & 6/10 & umbrella   & 6/10 \\
        \bottomrule
    \end{tabularx}
    \vspace{-0.4cm}
\end{table}

We identified the concepts that most commonly influenced models to generate male- and female-centric captions. For each model, we ranked $79$ concepts (highest to lowest) using \textit{HnS}'s global attribution scores. We created two lists: (1) male-centric concepts: top $20$ concepts in each model with the highest positive attribution score, and (2) female-centric concepts: top $20$ concepts with the highest negative attribution score. In Table \ref{tab:captioning_overlap}, across the $10$ models, we identified the concepts that frequently appeared in the list of male-centric and female-centric concepts. 

\textbf{\textit{HnS} can identify gender biases in captioning models, without using gender or object annotations}. In Table \ref{tab:captioning_overlap}, we see several bias trends. For instance, for the tie concept, $9/10$ models showed high positive attribution scores. This indicates that when images had a tie, $9/10$ models generated male-centric captions. For the cake concept, $7/10$ models showed high negative attribution scores. When images had a cake, $7/10$ models generated female-centric captions. Similar gender-bias trends have been shown in prior work \citep{DBAC, tang2021mitigating}.  
\subsection{Field Segmentation}
\label{subsec:ftw}

We used \textit{HnS} to understand why PRUE is the current state-of-the-art model for segmenting crop fields ~\citep{muhawenayo2026pruepracticalrecipefield}. PRUE accepts two images as input: window A and window B. Window A is the image of a field area (e.g., a field in Kenya) during the planting season (when crops are sown). Window B is the image of the same field area during the harvest season (when crops have fully grown). Using these two images, PRUE produces a semantic segmentation of the field area, in which each pixel is classified as either ``field'' or ``not field''. To interpret PRUE, we used crop field images from the Fields of the World Benchmark \cite{kerner2025fields}.

We used the $19$ classes in BigEarthNet \citep{sumbul2019bigearthnet} as our concept list (e.g., arable land, marine waters). Since PRUE operates on two images at a time, we hide concept $c$ in three different ways: ($1$)  $c$ is hidden in window A (visible in window B), ($2$) $c$ is hidden in window B (visible in window A), and ($3$) $c$ is hidden in both windows A and B. To compute concept attribution, we passed the original image ($c$ is present in both windows A and B) and its three masked versions ($c$ is hidden in either window A, window B, or both) to PRUE. We computed three attribution scores per concept (one for each masked version). 

\begin{figure*}[htbp]
    \centering
    \vspace{-0.3cm}
    \includegraphics[width=0.8\textwidth]{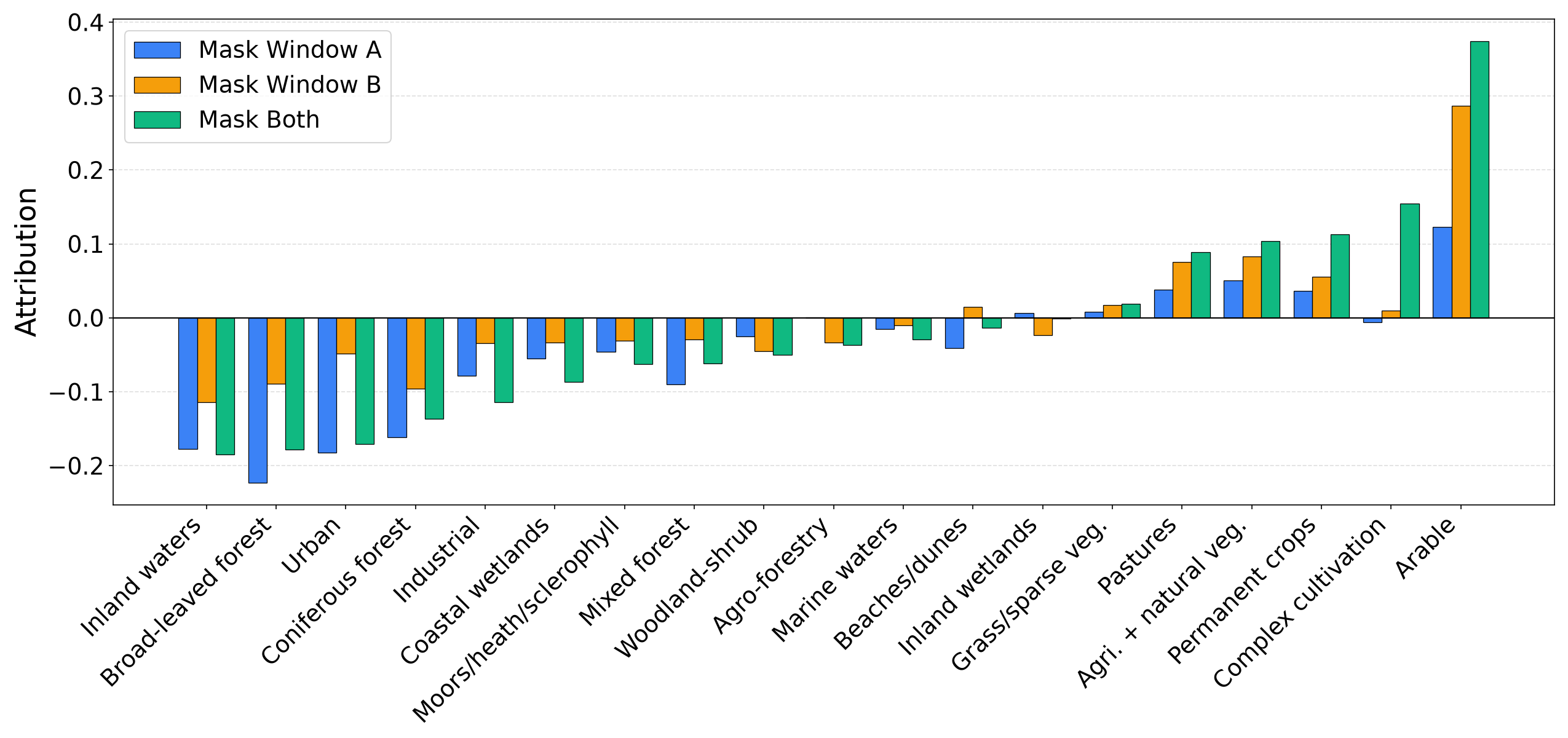}
    \vspace{-0.3cm}
    \caption{\textit{HnS} attributions for PRUE, a model used to segment crop fields. For each concept $c$ (e.g., inland waters), we reported three attribution scores: $(1)$ mask window A, with a blue bar ($c$ is only hidden in window A), $(2)$ mask window B, with a orange bar ($c$ is only hidden in window B), and $(3)$ mask both, with an green bar ($c$ is hidden in both windows). }
    \label{fig:ftw_prue}
    \vspace{-0.2cm}
\end{figure*}

For each version, we computed the attribution score using the difference between the average predicted probability (across all pixels) for the field class when the concept is present and when it is hidden. In Figure \ref{fig:ftw_prue}, for each concept, we reported these three attribution scores: mask window A (blue bar), mask window B (orange bar), and mask both (green bar).  

\textbf{PRUE can not only identify concepts that are useful to segment croplands, but also the concepts that are irrelevant to cropland}. PRUE showed high positive attribution scores for concepts such as arable (farmland) and permanent crops, which are important concepts for croplands. PRUE showed high negative attribution scores for irrelevant concepts like inland waters and urban. We show that these insights are replicated at the country level as well in Appendix \ref{appendix:FTW_country_level}.

\textbf{PRUE understands that complex cultivation patterns look different in the planting season (window A) and the harvest season (window B).} In Figure \ref{fig:ftw_prue}, for most concepts, PRUE showed consistent attribution scores across the three masks: either all attribution scores were non-zero (e.g., inland waters), or all scores were near zero (e.g., marine water). The complex cultivation concept exhibits a different behavior. For the partial masking cases (mask window A and mask window B), the attribution was near zero (blue and orange bars). However, for the full masking case (mask both), the attribution was high (green bar). PRUE has learned that most concepts are static: they should look the same in both windows (e.g., inland waters). If static concepts are masked even in one of the windows, PRUE's outputs are affected. PRUE has also learned that complex cultivation patterns are dynamic: they look different in both windows. If we mask this concept in one of the windows, PRUE's outputs are unaffected.

%% file: sections/06_discussion.tex
\section{Limitations}

\textbf{\textit{HnS} cannot always guarantee concept isolation}. To ensure optimal attribution scores from \textit{HnS}, concept vectors should be sufficiently distinct from one another (concept isolation). While concept isolation is largely ensured by our L\"odwin orthogonalization step, prior work has shown that VLMs like SigLIP can produce similar vectors for two concepts \citep{rawlekar2025efficiently, materzynska2022disentangling}. In such cases, even orthogonalization cannot guarantee concept isolation. Users should be mindful of potential entanglement and avoid supplying similar concepts (e.g., dog and canine). 


\textbf{\textit{HnS} only works for concepts that have spatial meaning.} We cannot compute attribution for abstract concepts (e.g., lighting, happiness) since they cannot be masked. 

%% file: sections/07_conclusion.tex
\section{Conclusion}

We introduced Hide and Seek (\textit{HnS}), a technique for measuring concept attribution in models. Our results show that \textit{HnS} provides the most reliable concept attribution among existing techniques. \textit{HnS} offers a combination of benefits that no existing technique offers: (1) it does not require a curated concept dataset, (2) it can measure attribution for both white and black-box models, (3) it provides both global and local (image-level) attributions, and (4) it can be used to interpret specialized models such as captioning and segmentation models. Although \textit{HnS} provides more accurate concept attribution scores than previous methods, these scores can be further improved with human input. Users can manually inspect the representative set $R_c$ to remove images that do not represent a concept. Users can also avoid using concept names with similar meanings.

%% file: sections/08_appendix.tex
\newpage
\appendix

\input{sections/appendices/compute_costs}

\input{sections/appendices/component_ablation}

\input{sections/appendices/perturbation_methods}

\input{sections/appendices/modality_shift_intuition}

\input{sections/appendices/gdino}

\input{sections/appendices/broden_investigation}

\newpage

\input{sections/appendices/shortcut_experiment}

\input{sections/appendices/dataset_ablation}

\input{sections/appendices/local_attribution}

\input{sections/appendices/captioning_exp_details}

\input{sections/appendices/FTW_county_level}

%% file: sections/appendices/compute_costs.tex
\section{Computational Cost Comparison}
\label{appendix:Computational_Cost}

Hide and Seek (\textit{HnS}) estimates attribution by directly comparing the black-box model output on original and masked images, so unlike a method that fits a single reusable probe, its cost scales with the number of images scored per concept rather than with the size of any training set. We report wall-clock time at the actual scale used throughout this paper: \textit{HnS} retrieves from the full official-val pool (16{,}874 images) with $K{=}50$; PCBM's CLIP-concepts variant is trained on the full official CelebA train partitions (162,770 images); TCAV uses the same exemplar, random, and control set sizes as our main results. Each method's procedure is run twice to reflect both tasks used in our experiments (Attractive, Male).

Tables~\ref{tab:computational_cost_rtx4090} and~\ref{tab:computational_cost_a100} split each method's cost along a different axis: how much of it is spent purely on a black-box-independent encoder, versus how much requires querying the specific black-box model being explained. This split matters because only the second kind of cost must be paid again if the same images and concepts are used to explain a different black-box model: the encoder-side work carries over unchanged, since it never depends on the model being audited.

By this measure, TCAV is the only method with no reusable component at all: a CAV is fit directly on a specific model's internal activations at a specific layer, so its entire cost is black-box-dependent and must be repeated in full for every new model (405.8s on an RTX-4090, 349.5s on an A100). The other four methods all rely on a backbone that is decoupled from the model being explained, so most of their cost carries over, and this pattern holds consistently across both GPUs: black-box-dependent cost accounts for 3.0\% of \textit{HnS}'s total on an RTX-4090 (10.6s of 348.7s) and 2.5\% on an A100 (6.0s of 238.0s); 28.0\% of PCBM (SigLIP)'s on an RTX-4090 (461.5s of 1648.5s) and 31.7\% on an A100 (391.7s of 1236.4s); 38.8\% of PCBM (CLIP-RN50)'s on an RTX-4090 (429.6s of 1108.1s) and 40.3\% on an A100 (348.1s of 864.7s); and 42.0\% of PCBM (ResNet-18)'s on an RTX-4090 (481.6s of 1145.7s) and 36.1\% on an A100 (347.3s of 962.5s). Concretely, after explaining one black-box model, explaining a second model on the same images and concepts would cost \textit{HnS} only another 6-11s, depending on hardware, versus roughly 350-480s for the three PCBM variants, and the full cost again from scratch for TCAV.

\begin{table}[htbp]
    \center
    \caption{Wall-clock cost with an Nvidia A-100 for different techniques, split into encoder-side cost that is reusable across black-box models and cost that depends on the specific black-box model being explained.}
    \label{tab:computational_cost_a100}
    \begin{tabularx}{\textwidth}{lYYY}
        \toprule
        \textbf{Method} & \textbf{Reusable (encoder-side)} & \textbf{Black-Box-Dependent} & \textbf{Total} \\
        \midrule
        Hide and Seek (ours) & \textbf{232.0s} & \ \ \ \textbf{6.0s} & \textbf{238.0s} \\
        TCAV & - & 349.5s & 349.5s \\
        PCBM (SigLIP) & 844.7s & 391.7s & 1236.4s \ \ \ \\
        PCBM (CLIP) & 516.6s & 348.1s & 864.7s\\
        PCBM (ResNet-18) & 615.2s & 347.3s & 962.5s\\
        \bottomrule
    \end{tabularx}
\end{table}

\begin{table}[htbp]
    \center
    \caption{Wall-clock cost with an Nvidia RTX-4090 for different techniques, split into encoder-side cost that is reusable across black-box models and cost that depends on the specific black-box model being explained.}
    \label{tab:computational_cost_rtx4090}
    \begin{tabularx}{\textwidth}{lYYY}
        \toprule
        \textbf{Method} & \textbf{Reusable (encoder-side)} & \textbf{Black-Box-Dependent} & \textbf{Total} \\
        \midrule
        Hide and Seek (ours) & \textbf{338.1s} &\ \textbf{10.6s} & \ \textbf{348.7s} \\
        TCAV & - & 405.8s &\ \  405.8s \\
        PCBM (SigLIP) & 1187.0s & 461.5s & 1648.5s \\
        PCBM (CLIP) & \ \  678.5s & 429.6s & 1108.1s\\
        PCBM (ResNet-18) &\ \  664.1s & 481.6s & 1145.7s\\
        \bottomrule
    \end{tabularx}
\end{table}

%% file: sections/appendices/component_ablation.tex
\section{Component Ablation}
\label{appendix:component_ablation}

\subsection{Orthogonalization and Modality Shift}
\label{appendix:subsec_corrections}

To explore the impact of our vector preprocessing steps (orthogonalization and modality shift correction) across encoders, we ablate them across four encoders (SigLIP, CLIP, OpenCLIP-H and Perception) across two datasets (CelebA and COCO-Stuff) with a Resnet-18 based black box model.

\begin{table}[htbp]
\centering
\caption{Joint ablation of encoder, orthogonalization, and Modality Shift for CelebA dataset with ResNet18 black box classifier.}
\label{tab:celeba-official-train-encoder-orth-modality-ablation}
\begin{tabular}{cc}
\begin{subtable}[t]{0.45\textwidth}
\centering
\caption{SigLIP}
\begin{tabular}{lcc}
\toprule
 & \multicolumn{2}{c}{Orthogonalize} \\
\cmidrule(lr){2-3}
Modality Shift & On & Off \\
\midrule
On & \textbf{0.615} & 0.526 \\
Off & 0.613 & 0.525 \\
\bottomrule
\end{tabular}
\end{subtable} &
\begin{subtable}[t]{0.45\textwidth}
\centering
\caption{CLIP}
\begin{tabular}{lcc}
\toprule
 & \multicolumn{2}{c}{Orthogonalize} \\
\cmidrule(lr){2-3}
Modality Shift & On & Off \\
\midrule
On & 0.237 & 0.371 \\
Off & 0.257 & \textbf{0.409} \\
\bottomrule
\end{tabular}
\end{subtable} \\[1em]
\begin{subtable}[t]{0.45\textwidth}
\centering
\caption{OpenCLIP-H}
\begin{tabular}{lcc}
\toprule
 & \multicolumn{2}{c}{Orthogonalize} \\
\cmidrule(lr){2-3}
Modality Shift & On & Off \\
\midrule
On & \textbf{0.354} & 0.349 \\
Off & 0.322 & 0.186 \\
\bottomrule
\end{tabular}
\end{subtable} &
\begin{subtable}[t]{0.45\textwidth}
\centering
\caption{Perception}
\begin{tabular}{lcc}
\toprule
 & \multicolumn{2}{c}{Orthogonalize} \\
\cmidrule(lr){2-3}
Modality Shift & On & Off \\
\midrule
On & 0.498 & \textbf{0.573} \\
Off & 0.544 & 0.544 \\
\bottomrule
\end{tabular}
\end{subtable}
\end{tabular}
\end{table}

\begin{table}[htbp]
\centering
\caption{Joint ablation of encoder, orthogonalization, and Modality Shift for COCO-Stuff dataset with ResNet18 black box classifier.}
\label{tab:coco_stuff-official-train-encoder-orth-modality-ablation}
\begin{tabular}{cc}
\begin{subtable}[t]{0.45\textwidth}
\centering
\caption{SigLIP}
\begin{tabular}{lcc}
\toprule
 & \multicolumn{2}{c}{Orthogonalize} \\
\cmidrule(lr){2-3}
Modality Shift & On & Off \\
\midrule
On & 0.315 & 0.300 \\
Off & \textbf{0.318} & 0.274 \\
\bottomrule
\end{tabular}
\end{subtable} &
\begin{subtable}[t]{0.45\textwidth}
\centering
\caption{CLIP}
\begin{tabular}{lcc}
\toprule
 & \multicolumn{2}{c}{Orthogonalize} \\
\cmidrule(lr){2-3}
Modality Shift & On & Off \\
\midrule
On &  0.322 & \textbf{0.331} \\
Off & 0.309 & 0.272 \\
\bottomrule
\end{tabular}
\end{subtable} \\[1em]
\begin{subtable}[t]{0.45\textwidth}
\centering
\caption{OpenCLIP-H}
\begin{tabular}{lcc}
\toprule
 & \multicolumn{2}{c}{Orthogonalize} \\
\cmidrule(lr){2-3}
Modality Shift & On & Off \\
\midrule
On & 0.180 & \textbf{0.197} \\
Off & 0.172 & 0.156 \\
\bottomrule
\end{tabular}
\end{subtable} &
\begin{subtable}[t]{0.45\textwidth}
\centering
\caption{Perception}
\begin{tabular}{lcc}
\toprule
 & \multicolumn{2}{c}{Orthogonalize} \\
\cmidrule(lr){2-3}
Modality Shift & On & Off \\
\midrule
On & 0.223 & \textbf{0.245} \\
Off & 0.226 & 0.192 \\
\bottomrule
\end{tabular}
\end{subtable}
\end{tabular}
\end{table}

Table~\ref{tab:celeba-official-train-encoder-orth-modality-ablation} and Table~\ref{tab:coco_stuff-official-train-encoder-orth-modality-ablation} report, for each encoder, the correlation between our attribution scores and ground truth under all four combinations of orthogonalization and modality shift correction switched on or off.

Neither preprocessing step is uniformly helpful: the best configuration is encoder-specific, and in some cases, dataset-specific as well. On CelebA, SigLIP and OpenCLIP-H reach their best correlation with both orthogonalization and modality shift enabled (0.615 and 0.354, respectively), while CLIP is best with both disabled (0.409) and Perception is best with only modality shift enabled (0.573). On COCO-Stuff, three of the four encoders (CLIP, OpenCLIP-H, and Perception) reach their best correlation with modality shift enabled and orthogonalization disabled (0.331, 0.197, and 0.245), while SigLIP is the outlier, preferring the opposite combination, modality shift disabled and orthogonalization enabled (0.318).

\subsection{Encoder Ablation}
\label{appendix:subsec_encoder_ablations}

We summarize the results for cross-encoder ablations by taking the best-performing preprocessing combination for each encoder.

\begin{table}[htbp]
\centering
\caption{Each encoder's own best setting (over the orthogonalize x modality-correction grid)}
\label{tab:encoder-best-celeba-coco-stuff}
\begin{tabular}{lcccc}
\toprule
Dataset & SigLIP & CLIP & OpenCLIP-H & Perception \\
\midrule
CelebA & \textbf{0.633} & 0.409 & 0.354 & 0.573 \\
COCO-Stuff & 0.318 & \textbf{0.331} & 0.197 & 0.245 \\
\bottomrule
\end{tabular}
\end{table}

Table \ref{tab:encoder-best-celeba-coco-stuff} shows that the SigLIP encoder clearly outperforms other encoders for the CelebA datasets. In fact, the worst-case SigLIP performance ($\rho = 0.525$ on CelebA) is still competitive with the best-case performance of the next-best encoder, the Perception encoder ($0.573$). For the COCO-Stuff dataset, CLIP appears to be the best-performing model ($\rho=0.331$), with SigLIP closely following behind ($\rho=0.318$). 

Given that the trade-offs are minimal in the case of COCO-Stuff, we recommend using SigLIP with orthogonality and demeaning, as it significantly outperforms other methods on CelebA.

\subsection{K-Value}
\label{appendix:subsec_k_ablation}

The present set size $K$ determines how many images from the retrieval pool are pulled into a concept's positive set via top-$K$ cosine similarity to the concept vector. This choice trades off two competing risks: too small a $K$ gives the concept vector too few supporting images to reliably estimate the concept's attribution, while too large a $K$ pulls in images that are only weakly related to the concept, diluting the current set with noise.

To choose a default value, we sweep $K \in \{15, 30, 50, 75, 100\}$ and measure its effect on attribution quality, reported as Spearman correlation with ground truth on CelebA and COCO-Stuff. Table~\ref{tab:k-demean-true-ablation-celeba-coco-stuff} shows that correlation is highest at $K=50$ on both datasets, and declines as $K$ moves away from it in either direction (more steeply toward smaller $K$ on COCO-Stuff than toward larger $K$). The two datasets agree on $K=50$ as the optimal value, which we adopt as the default.

\begin{table}[htbp]
\centering
\caption{K (retrieval pool size) ablation.}
\label{tab:k-demean-true-ablation-celeba-coco-stuff}
\begin{tabular}{lccccc}
\toprule
Dataset & K=15 & K=30 & K=50 & K=75 & K=100 \\
\midrule
CelebA & 0.603 & 0.607 & \textbf{0.615} & 0.604 & 0.594 \\
COCO-Stuff & 0.269 & 0.297 & \textbf{0.315} & 0.311 & 0.305 \\
\bottomrule
\end{tabular}
\end{table}

\subsection{Alpha Value}
\label{appendix:subsec_alpha_ablation}

The concept localization step determines which patches of an image belong to a concept by thresholding each patch's similarity to the concept vector at $\tau_c = \mu_c + \alpha\sigma_c$, where $\mu_c$ and $\sigma_c$ are the mean and standard deviation of similarity scores pooled across the concept's present set. This threshold controls a precision/recall trade-off in the resulting mask: a larger $\alpha$ produces a smaller, more selective mask that isolates only the most strongly concept-aligned patches, while a smaller $\alpha$ produces a larger, more inclusive mask that risks capturing background or context patches unrelated to the concept. Too small a mask under-perturbs the concept and understates its effect, while too large a mask perturbs unrelated content and adds noise to the attribution.

\begin{figure*}[htbp]
    \centering
    \includegraphics[width=0.75\textwidth]{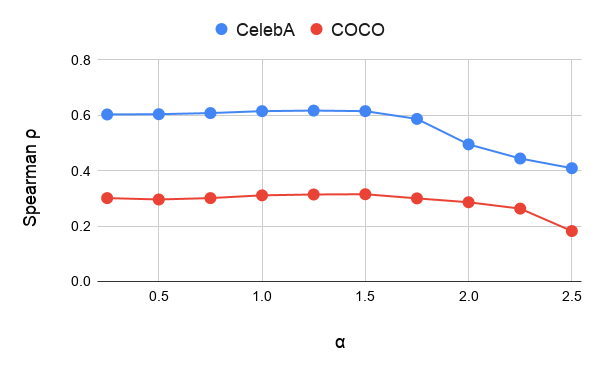}
    \caption{Spearman $\rho$ reported on CelebA and COCO-stuff dataset with ResNet-18 black box model and SigLIP encoder. We observe that for both datasets, the curve remains relatively flat until $\alpha=1.75$, after which it declines sharply. As such, we suggest keeping the default $\alpha=1.5$ for real-world usage.}
    \label{fig:alpha_ablation}
\end{figure*}

To choose a default value, we sweep $\alpha$ and measure its effect on attribution quality, reported as Spearman correlation with ground truth on CelebA and COCO-Stuff, using a ResNet-18 black-box model and the SigLIP encoder (Figure~\ref{fig:alpha_ablation}). In both datasets, the correlation is relatively stable across a wide range of $\alpha$, then declines sharply once $\alpha$ exceeds 1.75, consistent with the threshold becoming so strict that masks shrink to the point where they no longer capture enough of the concept to produce a reliable perturbation. Based on this, we set $\alpha = 1.5$ as the default for real-world use, since it sits comfortably inside the stable region while leaving headroom before the sharp decline.

\subsection{Prompt Variation}

For VLM-based models, prompt ensembling can help improve zero-shot accuracy~\citep{CLIP}. To test whether similar observations hold for our method, we evaluate 3 prompt variations and their ensemble. All other parameters are kept the same as the global attribution experiment in Section \ref{subsec:Masking_Experiment}.

The three prompt templates that we use are: "a photo of {concept}", "a photo showing {concept}", and "an image of {concept}". For CelebA, we rephrase each "{concept}" as "a person with {concept}" for better clarity. I.e., the concept "blond hair", can be templated as "a photo of a person with blond hair". For ensembling, we take the average embedding from all three templates.

\begin{table}[t]
\centering
\caption{Prompt ablation: 3 phrasing templates and ensemble, for \textit{HnS} on CelebA classifier and COCO-Stuff dataset. Class-pooled Spearman $\rho$ is reported against ground truth annotations.}
\label{tab:prompt-ablation-celeba-coco-stuff}
\begin{tabular}{lcc}
\toprule
Prompt & CelebA & COCO-Stuff \\
\midrule
a photo of {concept} & 0.621 & 0.308 \\
a photo showing {concept} & \textbf{0.634} & 0.311 \\
an image of {concept} & 0.605 & 0.306 \\
Ensemble  & 0.615 & \textbf{0.315} \\
\bottomrule
\end{tabular}
\end{table}

Table \ref{tab:prompt-ablation-celeba-coco-stuff} shows that for both datasets, prompt variations do not cause a significant difference in the resulting $\rho$ value compared to other ablated parameters. We also do not see any clear ``best" prompting strategy. As such, we defer to existing literature and suggest using prompt ensembling.

\newpage

%% file: sections/appendices/perturbation_methods.tex
\section{Perturbation Methods}
\label{appendix:perturbation_methods}

Given a concept's localized binary mask $M_i \in \{0,1\}^{H\times W}$, each perturbation strategy $f \in \mathcal{F}$ first computes a fully-filled version of the \emph{entire} image, $\mathrm{fill}_f(x)$, and then composites it with the original image using $M_i$ as a per-pixel selector:

\begin{equation}
    \tilde{x}_{f} = M_i \odot \mathrm{fill}_f(x) + (1-M_i) \odot x.
\end{equation}

Only pixels inside $M_i$ are ever replaced; the rest of the image is
untouched. We use seven strategies, chosen to span a range of ways a region can be made concept-uninformative without introducing the same
confound every time:

\begin{description}
    \item[\texttt{blur}] Gaussian-blurs the \emph{whole} image
    ($\sigma{=}20$px), then composites the blurred version within $M_i$. Because the blur is computed over the full image before compositing, a pixel near $M_i$'s boundary can still incorporate some low-frequency content bled in from just outside the mask through the blur kernel -- blur destroys fine detail and structure but does not fully erase a region's own coarse color statistics (its mean color typically survives largely intact). This is a common choice in the deletion-metric literature (e.g.\ RISE \citep{petsiuk2018rise}) specifically because it avoids the out-of-distribution ``hole'' a constant fill can create, while still meaningfully degrading the region.

    \item[\texttt{zero\_fill}] Replaces the region with solid black
    $(0,0,0)$. Erases color and structure completely, at the cost of
    creating a flat, unnatural patch that can itself be salient to a
    model independent of the concept being removed.

    \item[\texttt{white\_fill}] The complementary constant fill, solid
    white $(255,255,255)$ -- included specifically so that concepts whose real color is closest to black (e.g.\ dark hair, dark plumage) are not the only ones facing maximal fill-color contrast; \texttt{white\_fill} gives the same maximal-contrast treatment to light-colored concepts.

    \item[\texttt{mean\_fill}] Fills with a single constant color
    corresponding to the ImageNet normalization mean, $(124, 116, 104)$ in 8-bit RGB. A neutral, dataset-standard color chosen to be information-free under the same normalization the encoder itself expects, rather than an arbitrary constant.

    \item[\texttt{hue\_shift}] Rotates only the region's hue by
    $180^\circ$ in HSV space (each color's exact complementary hue) while leaving saturation and value -- i.e.\ luminance-driven edges, texture, and shape -- untouched. This is a surgical, color-only perturbation: unlike \texttt{blur} (which leaves much of the true color intact) or \texttt{zero\_fill}/\texttt{white\_fill} (which erases color and structure together), \texttt{hue\_shift} isolates color specifically, letting the perturbation-selection step choose a color-only edit when that is what best matches a concept's own direction, and a structure-destroying edit otherwise. Has no analog for single-band (non-RGB) rasters, since a lone reflectance band has no hue. Therefore, omitted for the FTW/PRUE application.

    \item[\texttt{zero\_fill\_noise}] Fills with i.i.d.\ Gaussian noise ($\mathcal{N}(0, 20)$ on the $0$-$255$ scale, clipped to valid range). A flat, constant fill is unlike any real photographic region. Even a genuinely uniform surface has natural sensor and texture noise, so this targets the same confound as \texttt{zero\_fill} while avoiding an artificially perfect-flat patch.

    \item[\texttt{noise\_then\_blur}] The same Gaussian noise draw as
    \texttt{zero\_fill\_noise}, subsequently Gaussian-blurred with the
    same $\sigma$ as \texttt{blur}. This combines the erasure completeness of noise with the smooth, photographically-plausible appearance of a blurred patch. Thus, it is content-free without either \texttt{zero\_fill\_noise}'s visible high-frequency static or \texttt{blur}'s own tendency to leak the original region's low-frequency content.
\end{description}

\paragraph{Multispectral (FTW) analog.} For the FTW/PRUE application
(Section~\ref{fig:ftw_prue}), each of the four Sentinel-2 bands
(R, G, B, NIR) is perturbed independently using the same six strategies (all but \texttt{hue\_shift}). Because raw Sentinel-2 digital numbers do not share RGB's fixed $0$-$255$ scale or ImageNet's fixed mean, the \texttt{zero\_fill}/ \texttt{white\_fill}/ \texttt{mean\_fill} constants and the noise strategies' standard deviation are each derived from that specific band's own observed minimum, maximum, and mean, rather than a fixed constant. Filled values are clipped to the band's valid integer range before the perturbed raster is written back to its original data type, to avoid
integer wraparound (e.g.\ a negative noise value silently overflowing to a near-maximum value under an unsigned encoding).

%% file: sections/appendices/modality_shift_intuition.tex
\section{Modality Shift Intuition}
\label{appendix:Modality_Shift_Intuition}

\begin{figure*}[htbp]
    \centering
    \includegraphics[width=0.8\textwidth]{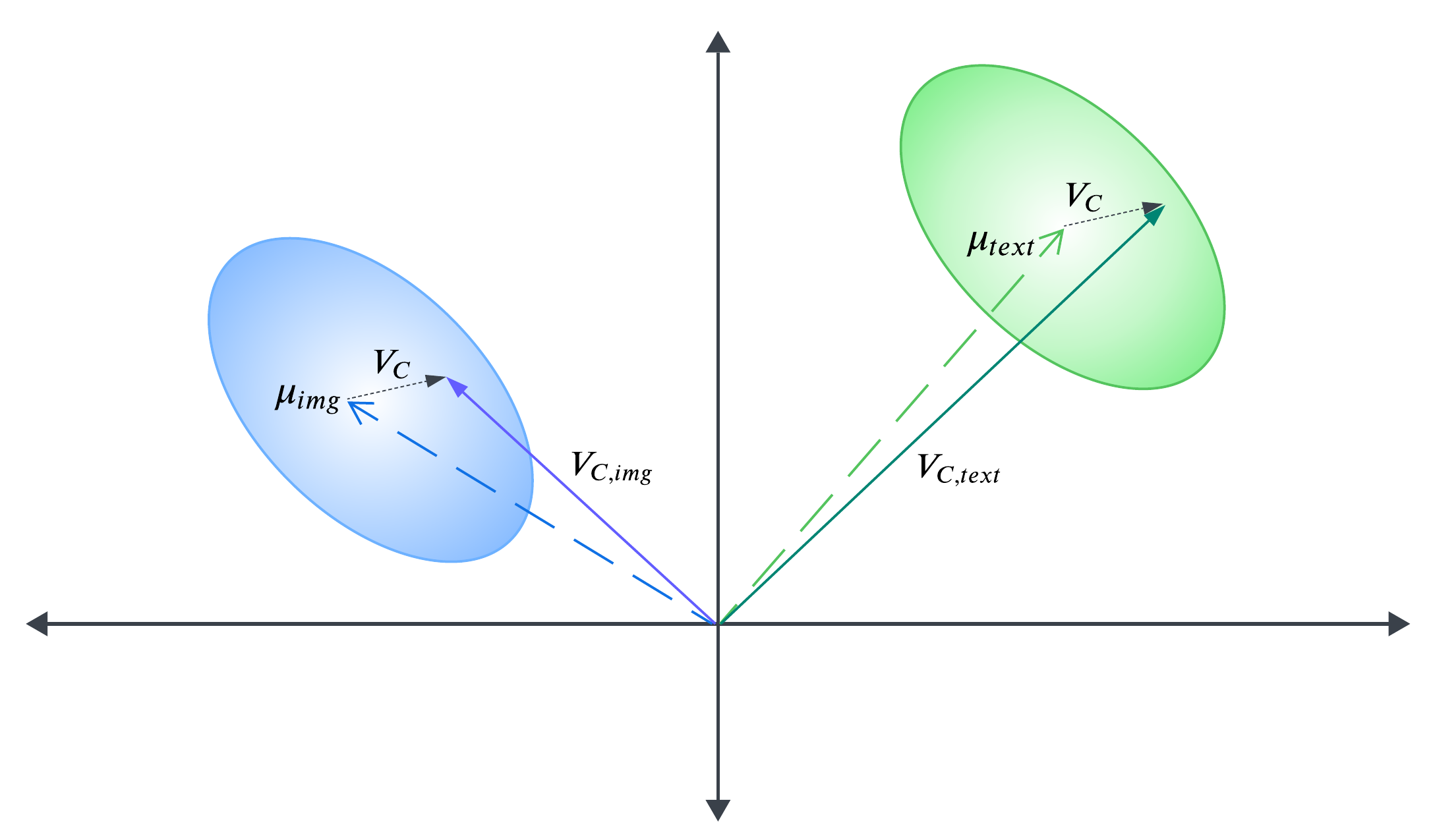}
    \caption{Modality Shift: image embeddings (blue) and text embeddings (green) occupy two separate regions, each offset from the origin by its own center, $\mu_{img}$ and $\mu_{text}$ respectively, rather than being centered at zero. For a given concept $C$, the full image embedding $V_{C,img}$ and text embedding $V_{C,text}$ (solid arrows from the origin) each decompose into their ellipsoid's center plus a concept-specific offset $V_C$ (dotted) that is shared between the two modalities. Because $\mu_{img} \neq \mu_{text}$, directly comparing $V_{C,img}$ and $V_{C,text}$ conflates this modality offset with the concept-specific signal we actually want, motivating the modality shift correction.}
    \label{fig:modality_shift}
\end{figure*}

\cite{levi2025double} demonstrates that in the CLIP embedding space, prior to normalization, the image and text embedding exist in two different ellipsoids that are not centered at the origin, see Figure \ref{fig:modality_shift}. Let's denote the centers of the image and text ellipsoids by $\mu_{img}$ and $\mu_{text}$, respectively. As such, the CLIP embeddings for the text "an image of a dog" can be modeled as 
\[    
V_{dog, text} = \mu_{text} + V_{dog}
\]

Similarly, we can model the embeddings for an actual image of a dog as:
\[
V_{dog, img} = \mu_{img} + V_{dog}
\]

Let's say we have an image database of $n$ images: $\{x_{1}, x_{2}, ..., x_{n}\}$. We can define their image embeddings as $\{V_{x_{1}, img},V_{x_{2}, img}, ..., V_{x_{n},img}\}$. This can be further simplified to $\{\mu_{img} + V_{x_{1}}, \mu_{img} + V_{x_{2}}, ..., \mu_{img} + V_{x_{n}}\}$.

To find the representative set $R_{c}$ for the concept dog, we must compute the cosine similarity ($V_{dog,text}\cdot V_{x_i,img}$) between the image embeddings in the dataset and the text embedding for the phrase "an image of a dog". This dot product expands to four terms:
\[
V_{dog,text} \cdot V_{x_i,img} = (\mu_{text} + V_{dog}) \cdot (\mu_{img} + V_{x_i}) 
\]
\[
V_{dog,text} \cdot V_{x_i,img} = \underbrace{V_{dog} \cdot V_{x_i}}_{\text{what we want}} + \underbrace{V_{dog} \cdot \mu_{img} + \mu_{text} \cdot \mu_{img} + \mu_{text} \cdot V_{x_i}}_{\text{contamination}}
\]

Only the first term, $V_{dog} \cdot V_{x_i}$, reflects whether image $x_i$ actually aligns with the dog-specific direction. The remaining three terms are contamination from the two ellipsoids' offsets. 

Modality shift correction aims to removes this contamination. We first estimate $\mu_{text}$ empirically, as the average text embedding over a large, diverse set of generic words. We then subtract our estimated $\hat{\mu}_{text}$ from each concept's own text embedding:
\[
V_{dog,text} - \hat{\mu}_{text} \approx V_{dog}
\]

Using this new approximation of $V_{dog}$, we compute the similarity as:
\[
V_{dog} \cdot V_{x_i,img} = V_{dog} \cdot (\mu_{img} + V_{x_i})
\]

\[
V_{dog} \cdot V_{x_i,img} = \underbrace{V_{dog} \cdot V_{x_i}}_{\text{what we want}} + \underbrace{V_{dog} \cdot \mu_{img}}_{\text{contamination}}
\]

Note that the new contamination term $V_{dog} \cdot \mu_{img}$ is actually a constant term independent of the image $x_{i}$. Therefore, it acts as a constant offset in the cosine similarity computed across all images in the image dataset. As such, the relative ranking remains unchanged. This gives us a reliable representative set $R_{c}$.

%% file: sections/appendices/gdino.tex
\section{COCO-Stuff Ground Truth via Grounded Detection and Segmentation}
\label{appendix:coco_stuff_ground_truth}

Unlike CelebA, COCO-Stuff images have no pre-existing per-concept segmentation masks we can reuse to compute ground-truth attribution: knowing that an image contains a "dog" and a "bicycle" does not tell us \emph{where} each one is. To get a real, causal ground-truth signal for this track, we detect and segment each Broden concept directly within COCO's own images, then measure the effect of physically removing it on a trained classifier's predictions.

\paragraph{Concept vocabulary.} We use the same 143 active Broden concepts as elsewhere in this paper (the full 170-concept set minus the 27 concepts dropped in Appendix~\ref{appendix:Concept_Retrieval}), so the ground-truth vocabulary stays consistent across tracks.

\paragraph{Detector threshold calibration.} Grounding DINO, our open-vocabulary detector, requires a box confidence threshold and a text-matching threshold. Rather than choosing these by inspection, we calibrate them against real ground truth: we take the subset of Broden concepts whose names exactly match one of COCO's 80 official class names, sample 1,000 COCO train-set images containing at least one instance of an overlapping class, and sweep all 16 combinations of box threshold $\in \{0.30, 0.35, 0.40, 0.45\}$ and text threshold $\in \{0.25, 0.30, 0.35, 0.40\}$. For each combination, we run Grounding DINO on all 1,000 images and compare its predicted boxes against COCO's real human-annotated instance boxes for the same concepts, computing mean IoU, recall at IoU $\geq 0.5$, and precision at IoU $\geq 0.5$. The best combination, box threshold $= 0.30$ and text threshold $= 0.30$ (mean IoU $= 0.6708$), is used for all ground-truth generation.

\paragraph{Detection and deduplication.} For each COCO train-set image containing at least one of the 20 target classes, we query Grounding DINO once with all 143 concepts batched into a single caption, returning a bounding box and matched concept name for every detection. Because a single object instance can trigger multiple overlapping boxes, we apply non-maximum suppression (IoU $= 0.5$) separately for each concept, which collapses duplicate detections of the same instance while keeping genuinely distinct instances of that concept in the image.

\paragraph{Segmentation.} Each surviving box is passed to the Segment Anything Model (SAM) as a box prompt, producing one instance mask per detection. When a concept appears multiple times in an image, we take the union of moving that concept, meaning we remove every instance of it simultaneously.

\paragraph{Masking and scoring.} We compute the original image's class probabilities once per image with our trained ResNet-18 COCO-Stuff classifier, then compare against a masked variant per concept: the unioned SAM instance mask filled black. This gives us $\Delta p$ and $\Delta\text{logit}$ (original minus masked) for all 20 target classes.

%% file: sections/appendices/broden_investigation.tex
\section{Are Broden Concepts Reliable}
\label{appendix:Concept_Retrieval}

The Broden concept bank we use for COCO-Stuff (Section~\ref{subsec:Masking_Experiment}) is a repackaged, image-crop version of the original Broden dataset~\citep{Broden}. The concept bank was introduced in \citep{Broden_filtered} but is distributed alongside PCBM's own codebase~\citep{PCBM-h}. 

Due to inconsistent results in our initial experiments, we wanted to run a sanity check on the Broden dataset. We aimed to identify the Broden concepts that our SigLIP encoder struggled to recognize. For this purpose, we aimed to classify the positive and negative sets for each concept by projecting the images on the concept vector $\vv^{*}_{c}$ in SigLIP embedding space. The intuition being, if SigLIP cannot express the concept, the image representation along the vector $\vv_{c}^{*}$ will not be a good feature to classify the positive and negative set.

But a manual inspection of the worst-performing concepts revealed a different pattern. Instead of SigLIP failing to express the concept, we found that the concept annotations were incorrect. Below, we list the wrongly annotated concepts, along with some example mislabeled positive images for the concept.

\textbf{Mislabled concepts}: \textit{air\_conditioner}, \textit{apron}, \textit{awning}, \textit{bathroom\_s}, \textit{beak}, \textit{bedroom\_s}, \textit{blotchy}, \textit{bumper}, \textit{can}, \textit{candlestick}, \textit{cap}, \textit{clock}, \textit{dining\_room\_s}, \textit{drinking\_glass}, \textit{eye}, \textit{eyebrow}, \textit{faucet}, \textit{figurine}, \textit{handle}, \textit{inside\_arm}, \textit{jar}, \textit{knob}, \textit{loudspeaker}, \textit{mouse}, \textit{mouth}, \textit{outside\_arm}, and \textit{street\_s}. 

We exclude these 27 concepts entirely rather than attempting to repair them; the remaining 143 concepts are the ones used throughout Section~\ref{subsec:Masking_Experiment}.

\begin{figure}[htbp]
    \centering
    \begin{subfigure}[b]{0.48\linewidth}
        \includegraphics[width=\linewidth]{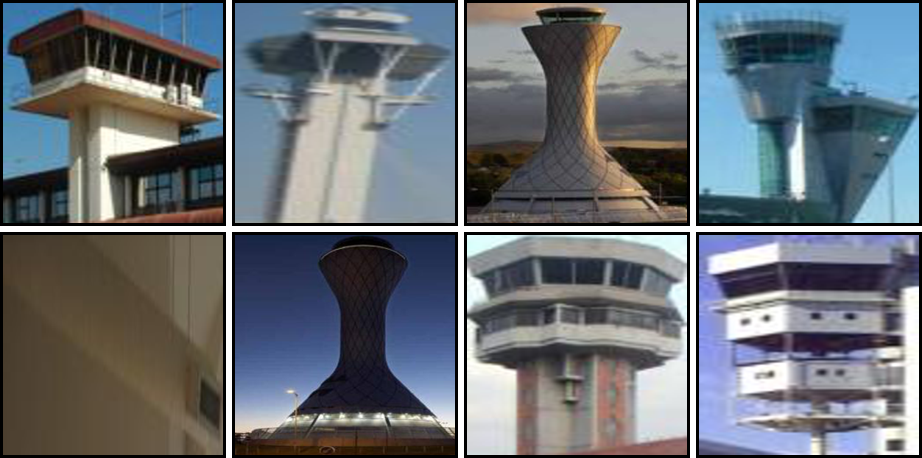}
        \caption{\textit{air conditioner}}
    \end{subfigure}
    \hfill
    \begin{subfigure}[b]{0.48\linewidth}
        \includegraphics[width=\linewidth]{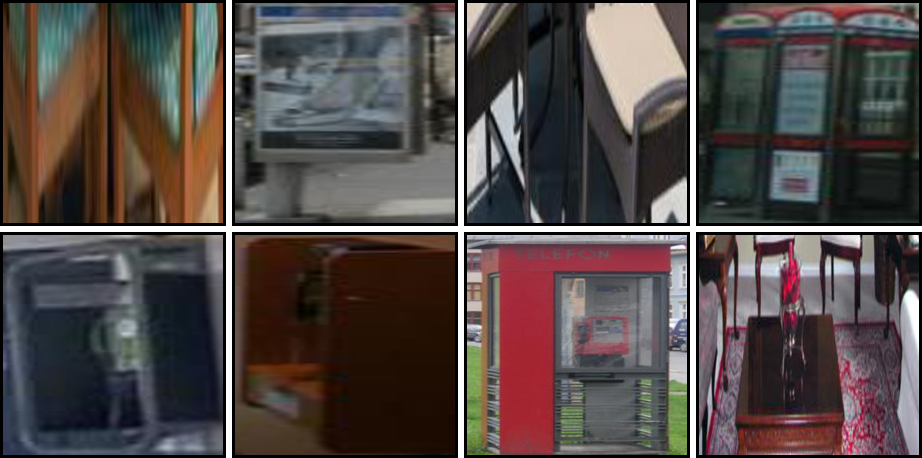}
        \caption{\textit{apron}}
    \end{subfigure}
    \caption{A random sample of images labeled as each concept, shown to illustrate that the labels are unreliable.}
    \label{fig:broden_dropped_air_conditioner_apron}
\end{figure}

\begin{figure}[htbp]
    \centering
    \begin{subfigure}[b]{0.48\linewidth}
        \includegraphics[width=\linewidth]{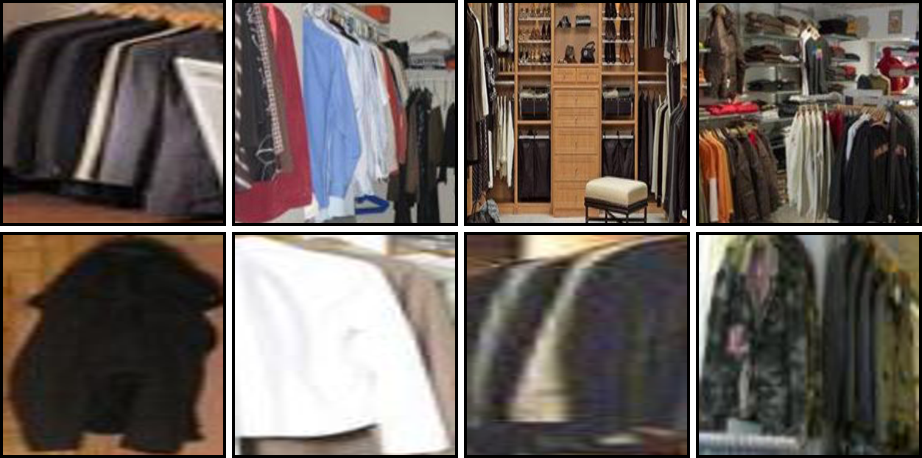}
        \caption{\textit{awning}}
    \end{subfigure}
    \hfill
    \begin{subfigure}[b]{0.48\linewidth}
        \includegraphics[width=\linewidth]{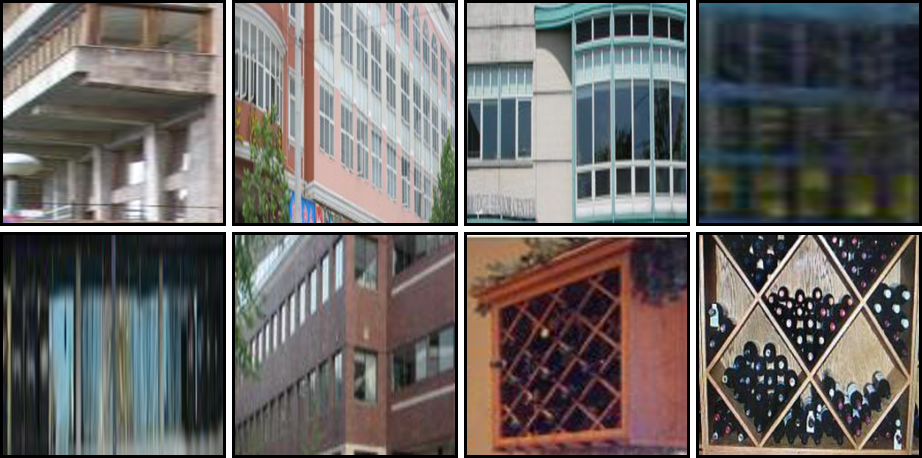}
        \caption{\textit{bathroom (scene)}}
    \end{subfigure}
    \caption{A random sample of images labeled as each concept, shown to illustrate that the labels are unreliable.}
    \label{fig:broden_dropped_awning_bathroom_s}
\end{figure}

\begin{figure}[htbp]
    \centering
    \begin{subfigure}[b]{0.48\linewidth}
        \includegraphics[width=\linewidth]{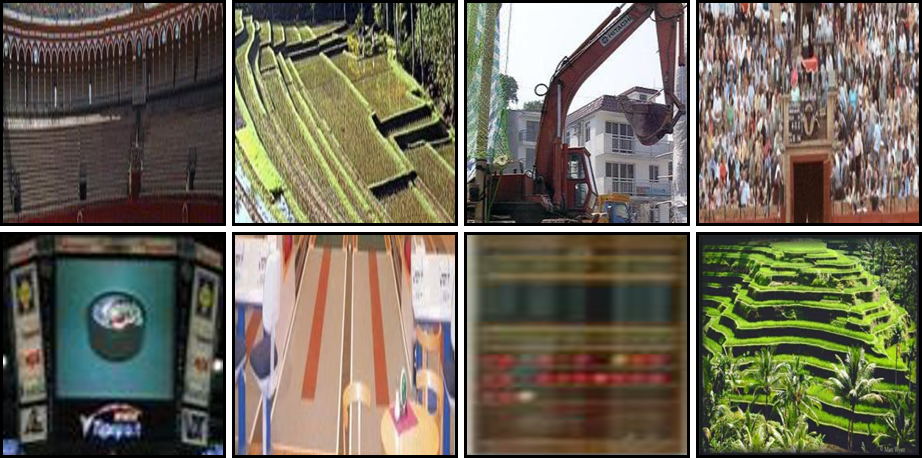}
        \caption{\textit{beak}}
    \end{subfigure}
    \hfill
    \begin{subfigure}[b]{0.48\linewidth}
        \includegraphics[width=\linewidth]{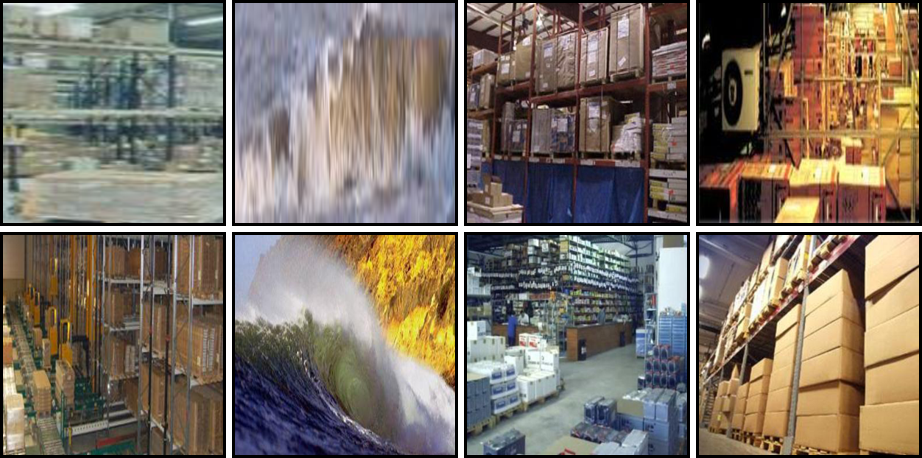}
        \caption{\textit{bedroom (scene)}}
    \end{subfigure}
    \caption{A random sample of images labeled as each concept, shown to illustrate that the labels are unreliable.}
    \label{fig:broden_dropped_beak_bedroom_s}
\end{figure}

\begin{figure}[htbp]
    \centering
    \begin{subfigure}[b]{0.48\linewidth}
        \includegraphics[width=\linewidth]{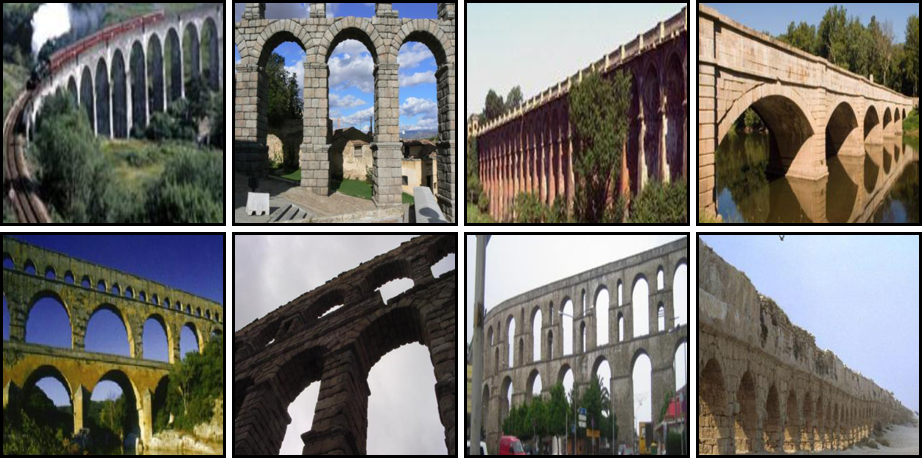}
        \caption{\textit{blotchy}}
    \end{subfigure}
    \hfill
    \begin{subfigure}[b]{0.48\linewidth}
        \includegraphics[width=\linewidth]{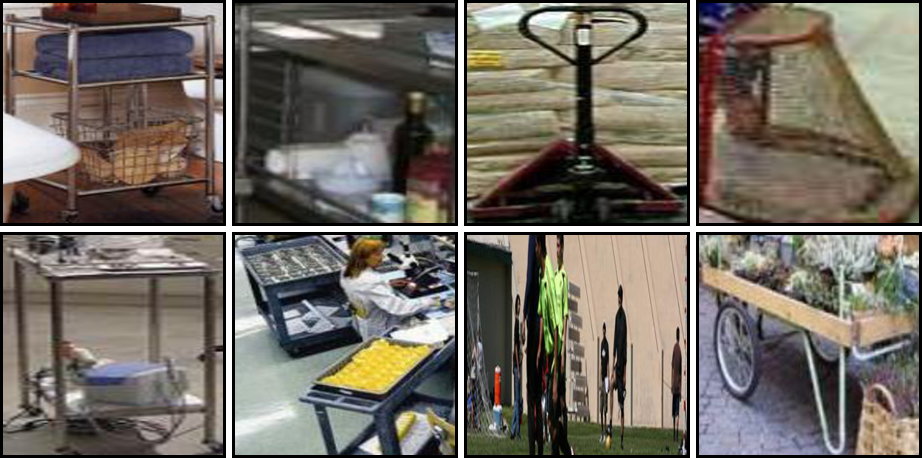}
        \caption{\textit{bumper}}
    \end{subfigure}
    \caption{A random sample of images labeled as each concept, shown to illustrate that the labels are unreliable.}
    \label{fig:broden_dropped_blotchy_bumper}
\end{figure}

\begin{figure}[htbp]
    \centering
    \begin{subfigure}[b]{0.48\linewidth}
        \includegraphics[width=\linewidth]{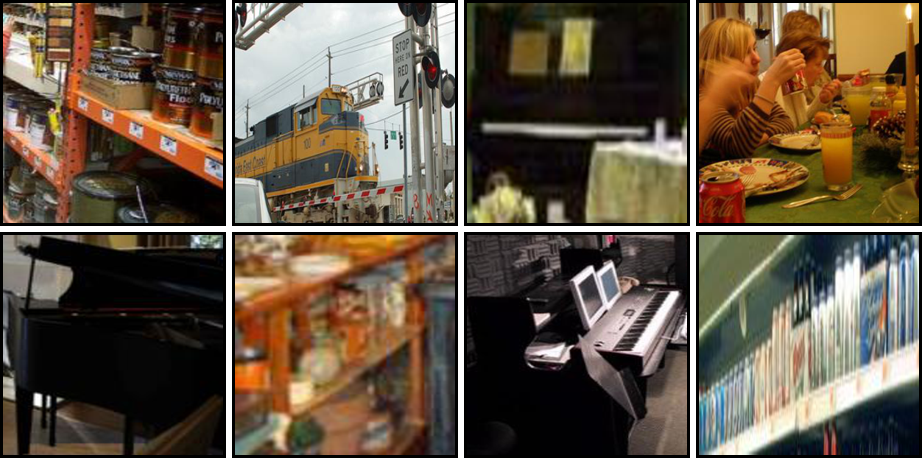}
        \caption{\textit{can}}
    \end{subfigure}
    \hfill
    \begin{subfigure}[b]{0.48\linewidth}
        \includegraphics[width=\linewidth]{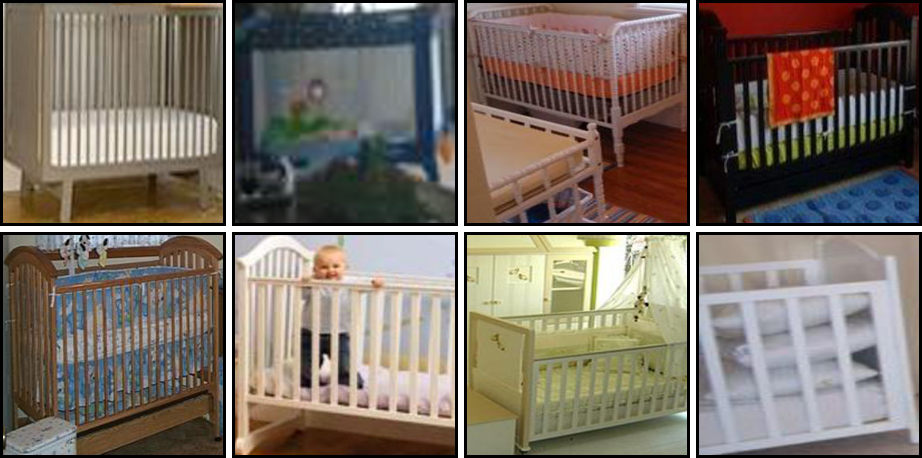}
        \caption{\textit{candlestick}}
    \end{subfigure}
    \caption{A random sample of images labeled as each concept, shown to illustrate that the labels are unreliable.}
    \label{fig:broden_dropped_can_candlestick}
\end{figure}

\begin{figure}[htbp]
    \centering
    \begin{subfigure}[b]{0.48\linewidth}
        \includegraphics[width=\linewidth]{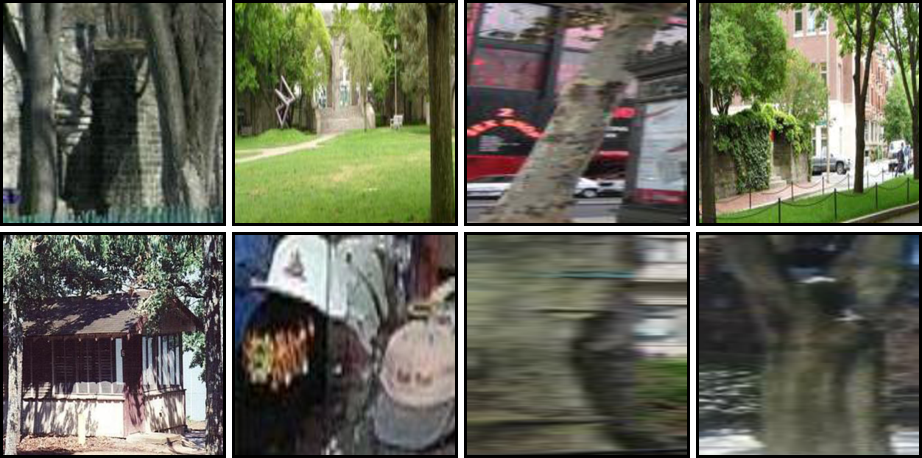}
        \caption{\textit{cap}}
    \end{subfigure}
    \hfill
    \begin{subfigure}[b]{0.48\linewidth}
        \includegraphics[width=\linewidth]{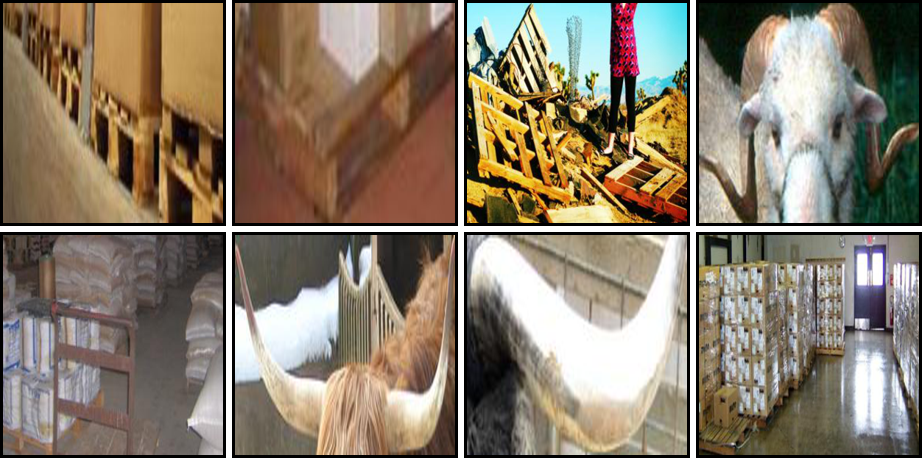}
        \caption{\textit{clock}}
    \end{subfigure}
    \caption{A random sample of images labeled as each concept, shown to illustrate that the labels are unreliable.}
    \label{fig:broden_dropped_cap_clock}
\end{figure}

\begin{figure}[htbp]
    \centering
    \begin{subfigure}[b]{0.48\linewidth}
        \includegraphics[width=\linewidth]{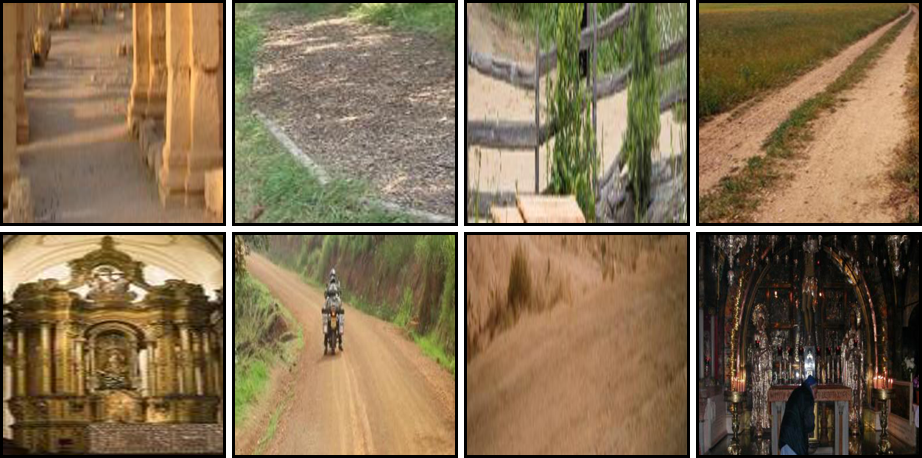}
        \caption{\textit{dining room (scene)}}
    \end{subfigure}
    \hfill
    \begin{subfigure}[b]{0.48\linewidth}
        \includegraphics[width=\linewidth]{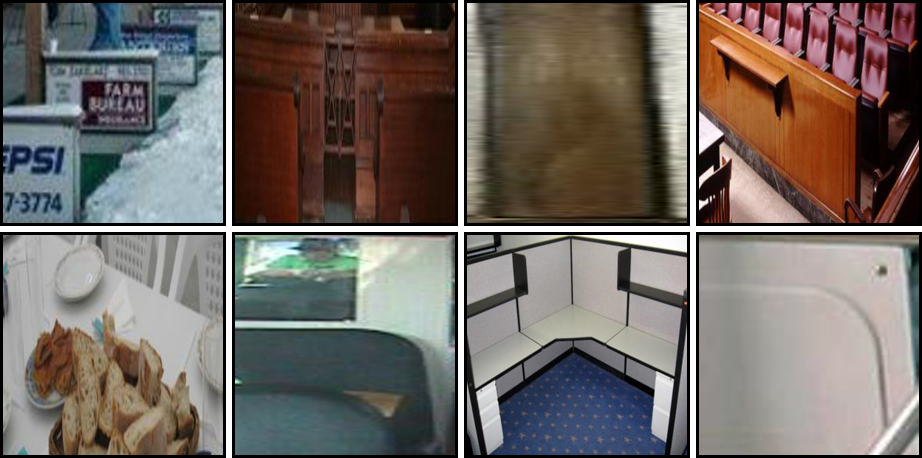}
        \caption{\textit{drinking glass}}
    \end{subfigure}
    \caption{A random sample of images labeled as each concept, shown to illustrate that the labels are unreliable.}
    \label{fig:broden_dropped_dining_room_s_drinking_glass}
\end{figure}

\begin{figure}[htbp]
    \centering
    \begin{subfigure}[b]{0.48\linewidth}
        \includegraphics[width=\linewidth]{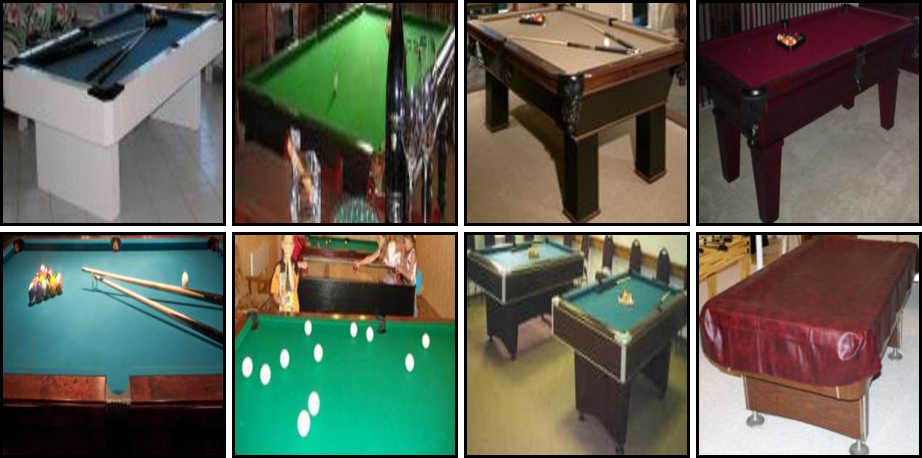}
        \caption{\textit{eye}}
    \end{subfigure}
    \hfill
    \begin{subfigure}[b]{0.48\linewidth}
        \includegraphics[width=\linewidth]{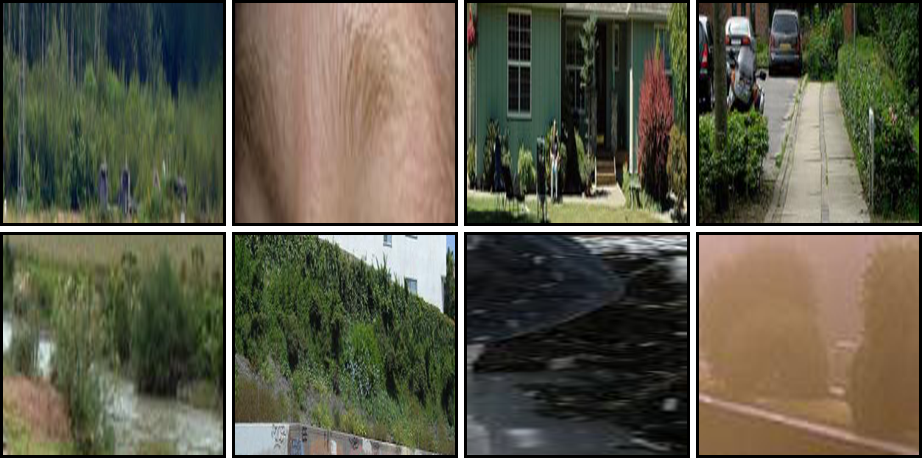}
        \caption{\textit{eyebrow}}
    \end{subfigure}
    \caption{A random sample of images labeled as each concept, shown to illustrate that the labels are unreliable.}
    \label{fig:broden_dropped_eye_eyebrow}
\end{figure}

\begin{figure}[htbp]
    \centering
    \begin{subfigure}[b]{0.48\linewidth}
        \includegraphics[width=\linewidth]{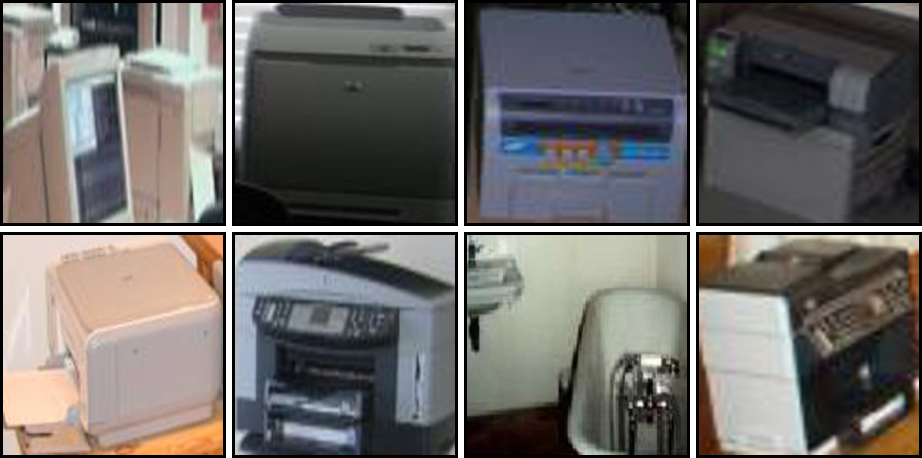}
        \caption{\textit{faucet}}
    \end{subfigure}
    \hfill
    \begin{subfigure}[b]{0.48\linewidth}
        \includegraphics[width=\linewidth]{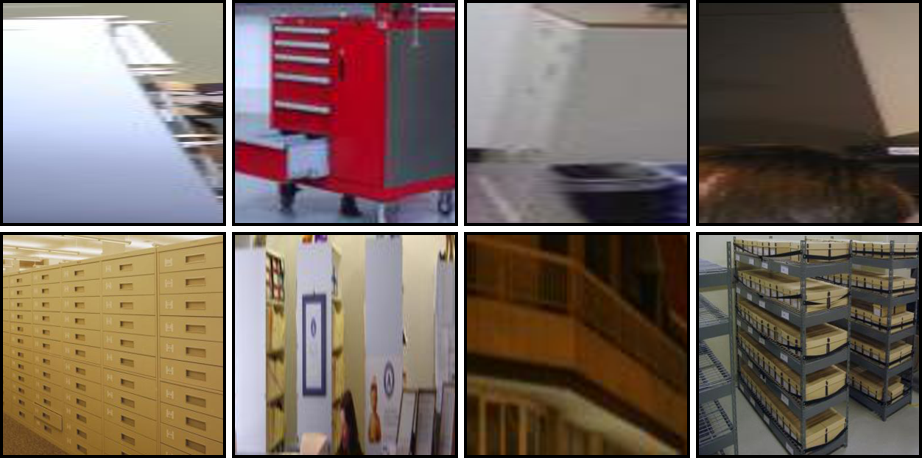}
        \caption{\textit{figurine}}
    \end{subfigure}
    \caption{A random sample of images labeled as each concept, shown to illustrate that the labels are unreliable.}
    \label{fig:broden_dropped_faucet_figurine}
\end{figure}

\begin{figure}[htbp]
    \centering
    \begin{subfigure}[b]{0.48\linewidth}
        \includegraphics[width=\linewidth]{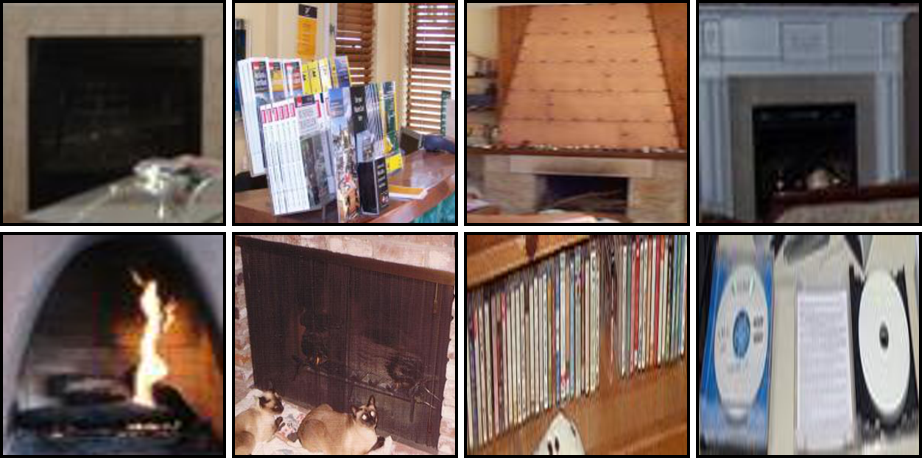}
        \caption{\textit{handle}}
    \end{subfigure}
    \hfill
    \begin{subfigure}[b]{0.48\linewidth}
        \includegraphics[width=\linewidth]{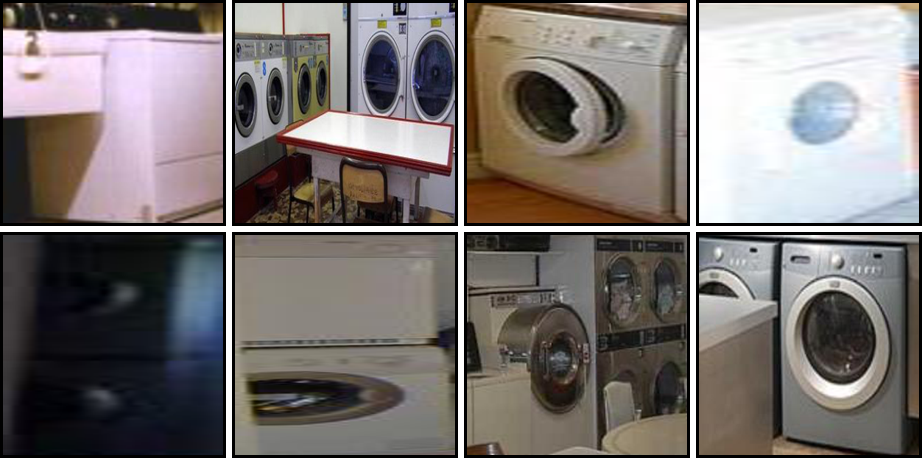}
        \caption{\textit{inside arm}}
    \end{subfigure}
    \caption{A random sample of images labeled as each concept, shown to illustrate that the labels are unreliable.}
    \label{fig:broden_dropped_handle_inside_arm}
\end{figure}

\begin{figure}[htbp]
    \centering
    \begin{subfigure}[b]{0.48\linewidth}
        \includegraphics[width=\linewidth]{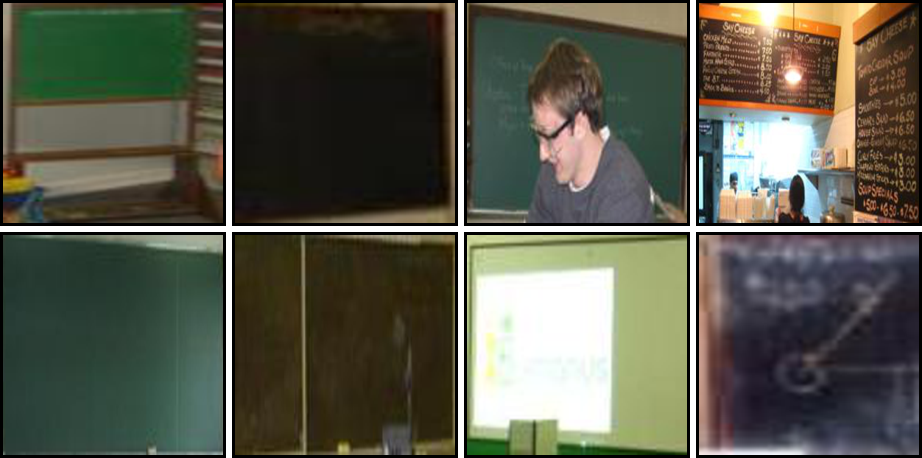}
        \caption{\textit{jar}}
    \end{subfigure}
    \hfill
    \begin{subfigure}[b]{0.48\linewidth}
        \includegraphics[width=\linewidth]{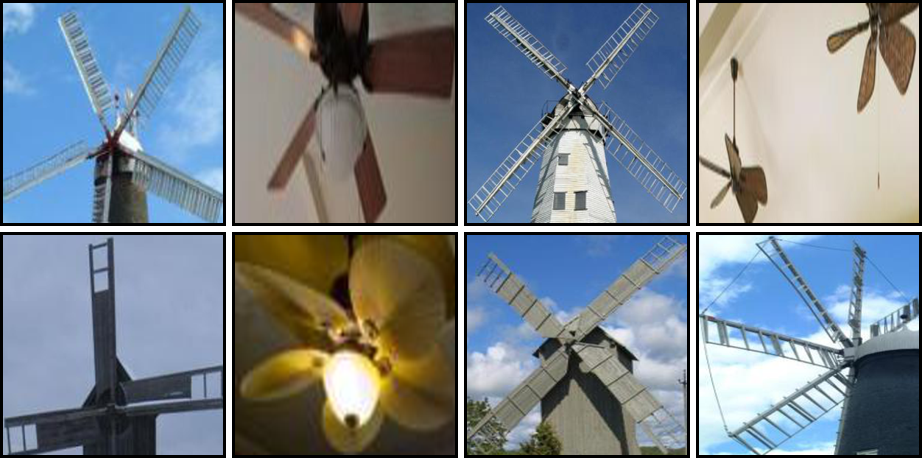}
        \caption{\textit{knob}}
    \end{subfigure}
    \caption{A random sample of images labeled as each concept, shown to illustrate that the labels are unreliable.}
    \label{fig:broden_dropped_jar_knob}
\end{figure}

\begin{figure}[htbp]
    \centering
    \begin{subfigure}[b]{0.48\linewidth}
        \includegraphics[width=\linewidth]{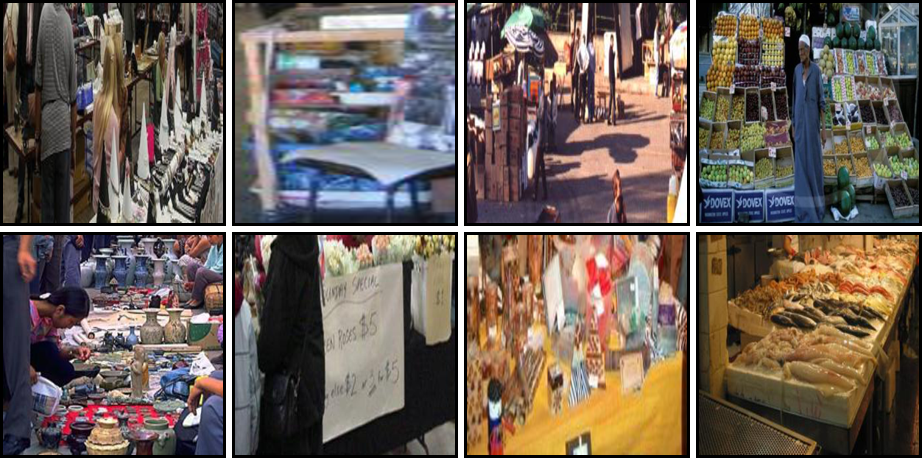}
        \caption{\textit{loudspeaker}}
    \end{subfigure}
    \hfill
    \begin{subfigure}[b]{0.48\linewidth}
        \includegraphics[width=\linewidth]{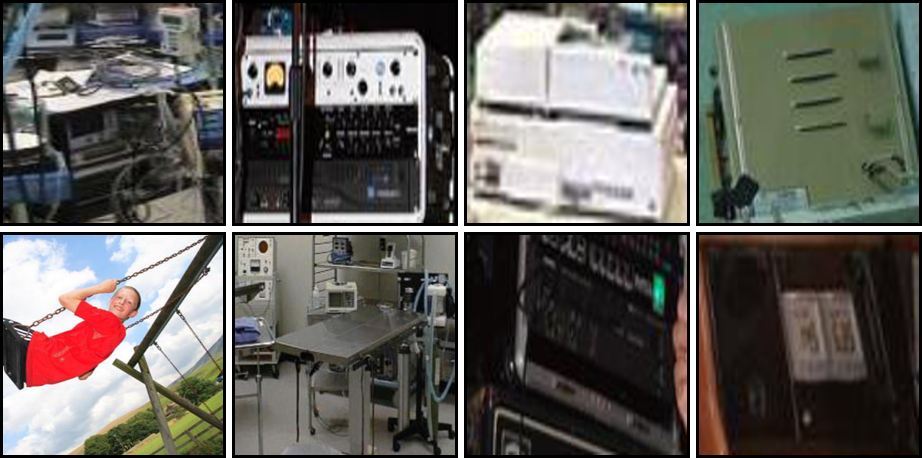}
        \caption{\textit{mouse}}
    \end{subfigure}
    \caption{A random sample of images labeled as each concept, shown to illustrate that the labels are unreliable.}
    \label{fig:broden_dropped_loudspeaker_mouse}
\end{figure}

\begin{figure}[htbp]
    \centering
    \begin{subfigure}[b]{0.48\linewidth}
        \includegraphics[width=\linewidth]{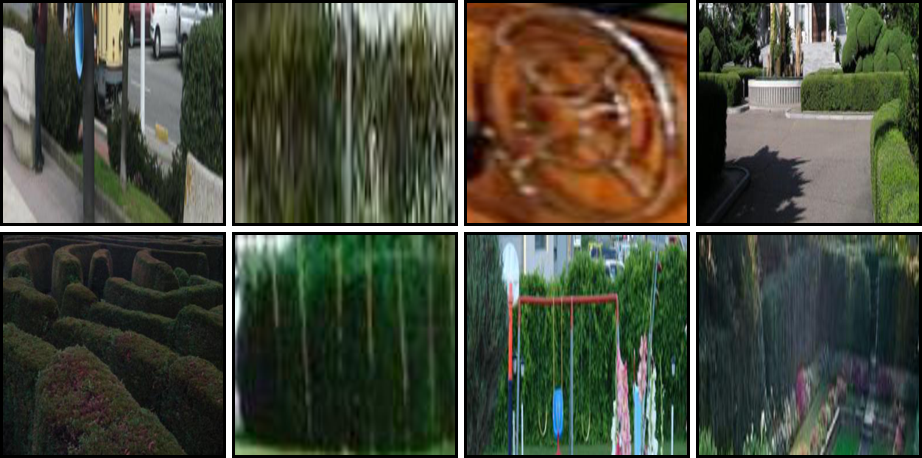}
        \caption{\textit{mouth}}
    \end{subfigure}
    \hfill
    \begin{subfigure}[b]{0.48\linewidth}
        \includegraphics[width=\linewidth]{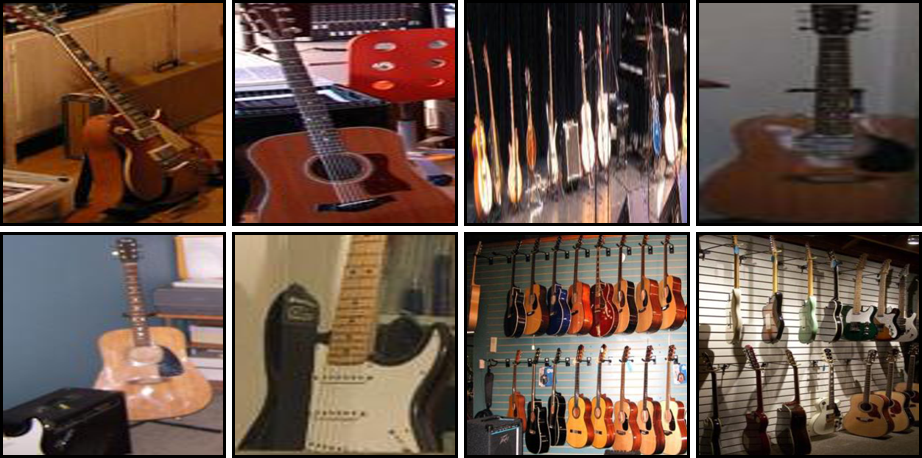}
        \caption{\textit{outside arm}}
    \end{subfigure}
    \caption{A random sample of images labeled as each concept, shown to illustrate that the labels are unreliable.}
    \label{fig:broden_dropped_mouth_outside_arm}
\end{figure}

\begin{figure}[htbp]
    \centering
    \begin{subfigure}[b]{0.48\linewidth}
        \includegraphics[width=\linewidth]{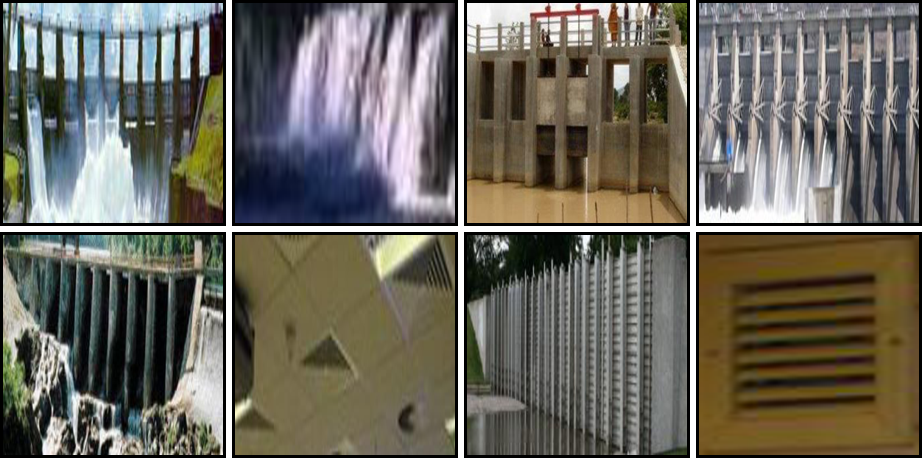}
        \caption{\textit{street (scene)}}
    \end{subfigure}
    \caption{A random sample of images labeled as each concept, shown to illustrate that the labels are unreliable.}
    \label{fig:broden_dropped_street_s}
\end{figure}

%% file: sections/appendices/shortcut_experiment.tex
\section{Shortcut Experiment}
\label{appendix:Shortcut_Experiment}

To test whether an attribution method's output reflects a model's actual reliance on real semantic concepts, rather than being an artifact of the attribution mechanism itself, we construct a controlled shortcut-learning setup similar to the shortcut experiment in \cite{DPA}. We inject a synthetic, class-correlated visual cue (a fixed-position solid-color block, one color per class) into a fraction $s \in \{0\%, 10\%, 20\%, ..., 100\%\}$ of each class's training images, and train a separate classifier $b_s$ at each shortcut rate $s$. As $s$ increases, $b_s$ can increasingly solve the task by relying solely on the injected shortcut rather than on actual image concepts.

\begin{figure*}[htbp]
    \vspace{-0.2cm}
    \centering
    \begin{subfigure}[b]{0.48\textwidth}
        \centering
        \includegraphics[width=\textwidth]{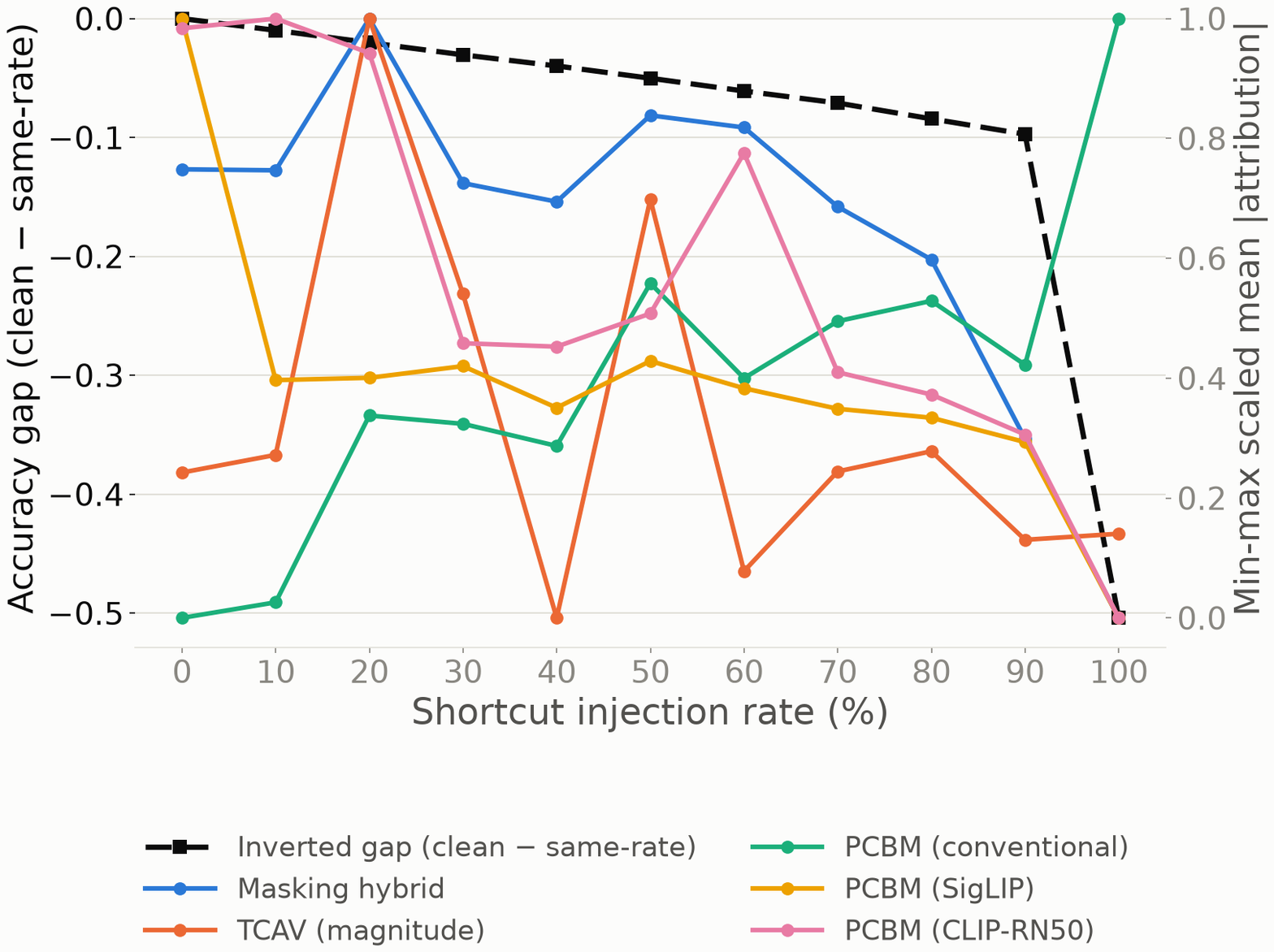}
        \caption{CelebA}
        \label{fig:celeba_shortcut_sub}
    \end{subfigure}
    \hfill
    \begin{subfigure}[b]{0.48\textwidth}
        \centering
        \includegraphics[width=\textwidth]{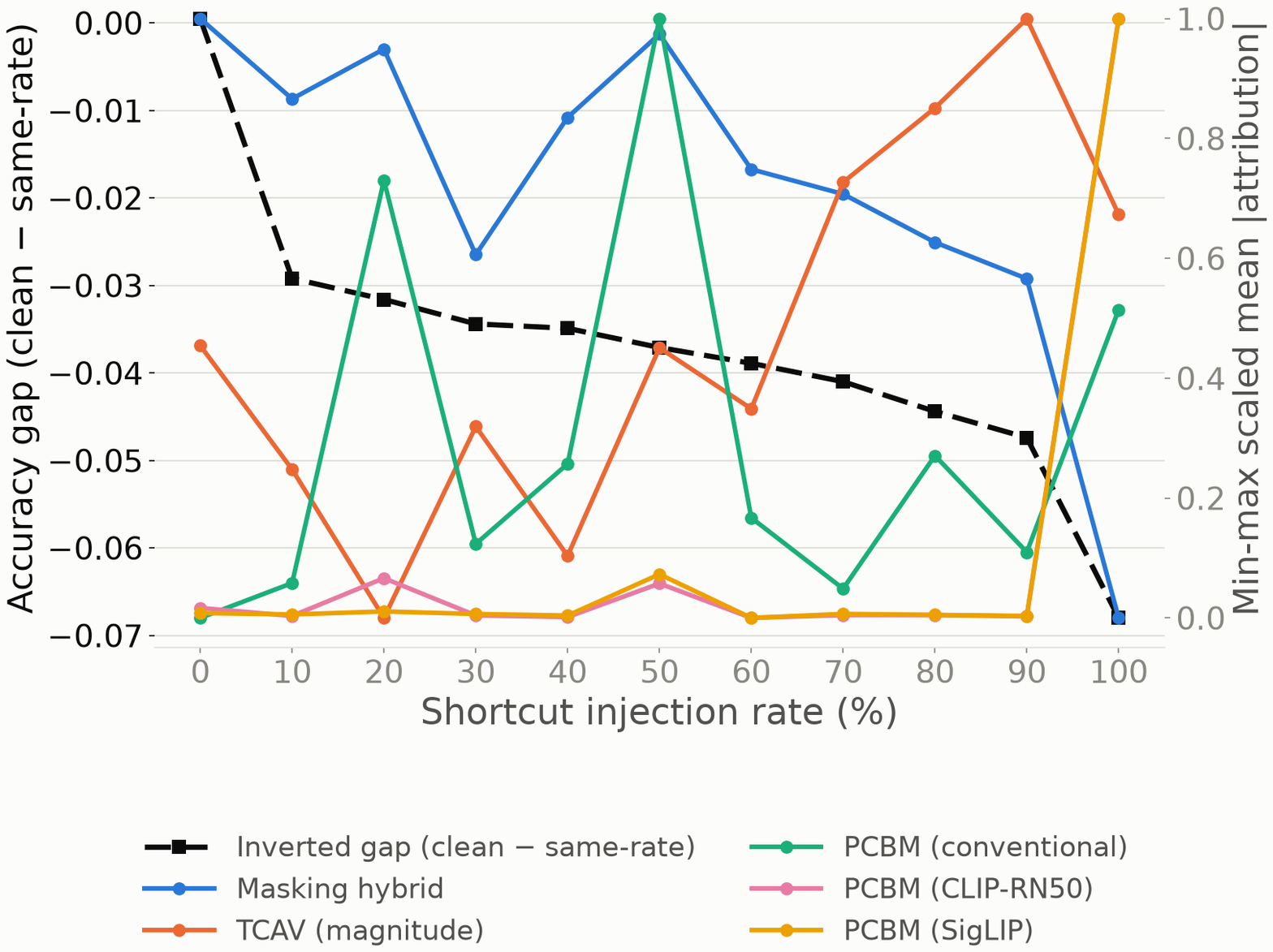}
        \caption{COCO-Stuff}
        \label{fig:coco_stuff_shortcut_sub}
    \end{subfigure}
    \caption{
        The accuracy gap $\Delta(s)$ is shown in the black dotted line (left axis, real scale). Average absolute concept attributions for each method are shown on the right-hand axis, min-max-normalized for visual comparability.
    }
    \vspace{-0.4cm}
    \label{fig:celeba_shortcut}
\end{figure*}

We quantify this growing reliance with the accuracy gap $\Delta(s) = \mathrm{acc}_{\mathrm{clean}}(b_s) - \mathrm{acc}_{s}(b_s)$, the difference between $b_s$'s accuracy on a clean, cue-free validation set and its accuracy on a validation set carrying the \emph{same} injection rate $s$ it was trained on. $\Delta(s)$ starts near zero at $s{=}0\%$ (no shortcut available, so the two accuracies coincide) and grows increasingly negative as $s$ rises, reflecting the classifier's growing preference for the shortcut over actual concepts (black dotted line in Figure~\ref{fig:celeba_shortcut}).

For each $b_s$, we compute every method's attribution score for the same fixed bank of 26 concepts for CelebA and 103 concepts for COCO-Stuff, and average their absolute magnitudes into a single per-rate summary, $\overline{|\mathrm{Attr}|}(s)$ (min-max normalized across rates in Figure~\ref{fig:celeba_shortcut} for visual comparability only. The correlations in Table~\ref{tab:shortcut_experiment} are computed
on the raw, unnormalized values, which Pearson correlation is invariant to, regardless. Table~\ref{tab:shortcut_experiment} reports the Pearson correlation between $\overline{|\mathrm{Attr}|}(s)$ and $b_s$'s own clean accuracy across the 4 shortcut rates: a method that correctly detects declining reliance on real concepts should show a strong \emph{positive} correlation (attribution magnitude falls as clean accuracy falls).

\begin{table*}[htbp]
    \center
    \caption{Pearson correlation between the accuracy gap $\Delta(s)$ and mean absolute concept attributions across the 11 shortcut rates ($s \in \{0\%,10\%,\ldots,100\%\}$).}
    \label{tab:shortcut_experiment}
    \begin{tabularx}{0.7\textwidth}{P{4cm}YY}
        \toprule
        \textbf{Metric} & \textbf{COCO-Stuff} & \textbf{CelebA}\\
        \midrule
        TCAV & -0.42 & \ 0.27\\
        PCBM (Resnet-18) & -0.28 & -0.84\\
        PCBM (CLIP-RN50) & -0.62 & \ 0.71\\
        PCBM (SigLIP) & -0.63 & \ 0.66\\
        Hide and Seek (ours) & \ \textbf{0.82} & \ \textbf{0.85}\\
        \bottomrule
    \end{tabularx}
\end{table*}

\paragraph{Results:} Hide and Seek (\textit{HnS}) shows a strong, statistically significant positive correlation on both datasets (CelebA: $r{=}0.85$, $p{<}0.001$; COCO-Stuff: $r{=}0.82$, $p{<}0.01$), correctly detecting the model's growing reliance on the injected shortcut as a substitute for real concepts in both settings. TCAV, by contrast, shows no significant relationship on either dataset ($r{=}0.27$, $p{=}0.42$ on CelebA; $r{=}-0.42$, $p{=}0.20$ on COCO-Stuff) and even flips the sign between them.

For PCBM, the CAV-fitted, whole-image ResNet-18 variant shows a strong \emph{negative} correlation on CelebA ($r{=}-0.84$, $p{<}0.01$) and a weaker, non-significant negative correlation on COCO-Stuff ($r{=}-0.28$, $p{=}0.40$): its mean attribution magnitude spikes at $s{=}100\%$ on CelebA, where the classifier's own clean-set accuracy has collapsed and PCBM's surrogate, produces arbitrarily large weights rather than decline. The two VLM PCBM variants disagree with \emph{themselves} across datasets: CLIP-RN50 is positive on CelebA ($r{=}0.71$, $p{=}0.01$) but negative on COCO-Stuff ($r{=}-0.62$, $p{=}0.04$); SigLIP shows the same reversal ($r{=}0.66$, $p{=}0.03$ on CelebA vs.\ $r{=}-0.63$, $p{=}0.04$ on COCO-Stuff), despite identical construction on a different backbone. Taken together, none of PCBM's three variants reliably tracks a model's actual reliance on real concepts across datasets, unlike \textit{HnS}, which is the only method in this comparison to show a consistent, significant, correctly-signed correlation on both.

%% file: sections/appendices/dataset_ablation.tex
\section{Dataset Ablation}
\label{appendix:Dataset_Ablation}

A question users might ask is: Does the retrieval pool need to be drawn from the same dataset, or can we extend it with other datasets? To test this, we re-run \textit{HnS}'s global attribution using two out-of-distribution face datasets, LFW and UTKFace, as the retrieval pool in place of CelebA, keeping the ground truth, the classifier, and every other component of the pipeline unchanged.

Table~\ref{tab:pooled_across_tasks} reports the resulting agreement with ground truth, averaged across the Attractive and Male tasks.

\begin{table*}[htbp]
    \center
    \caption{Concept attribution agreement with the official-train classifier's masking ground truth, averaged across the Attractive and Male tasks (previously\_used ground truth).}
    \label{tab:pooled_across_tasks}
    \begin{tabularx}{\textwidth}{P{4cm}YYY}
        \toprule
        \textbf{Method} & \textbf{Attractive} & \textbf{Male} & \textbf{Average}\\
        \midrule
        TCAV & 0.478 & 0.160 & 0.319 \\
        PCBM (ResNet-18) & 0.131 & 0.228 & 0.180 \\
        PCBM (CLIP-RN50) & 0.267 & 0.032 & 0.149 \\
        PCBM (SigLIP) & 0.320 & 0.096 & 0.208 \\
        Hide and Seek (UTKFace) & 0.501 & 0.426 & 0.463 \\
        Hide and Seek (LFW) & 0.641 & 0.502 & 0.571 \\
        Hide and Seek (CelebA) & \textbf{0.702} & \textbf{0.528} & \textbf{0.615} \\
        \bottomrule
    \end{tabularx}
\end{table*}

\begin{figure}[htbp]
    \centering
    \begin{subfigure}[t]{0.48\textwidth}
        \centering
        \includegraphics[width=\textwidth]{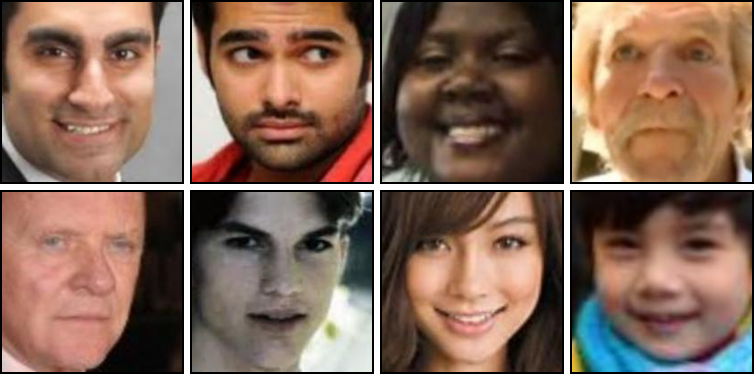}
        \caption{UTKFace}
        \label{fig:utkface_examples}
    \end{subfigure}
    \hfill
    \begin{subfigure}[t]{0.49\textwidth}
        \centering
        \includegraphics[width=\textwidth]{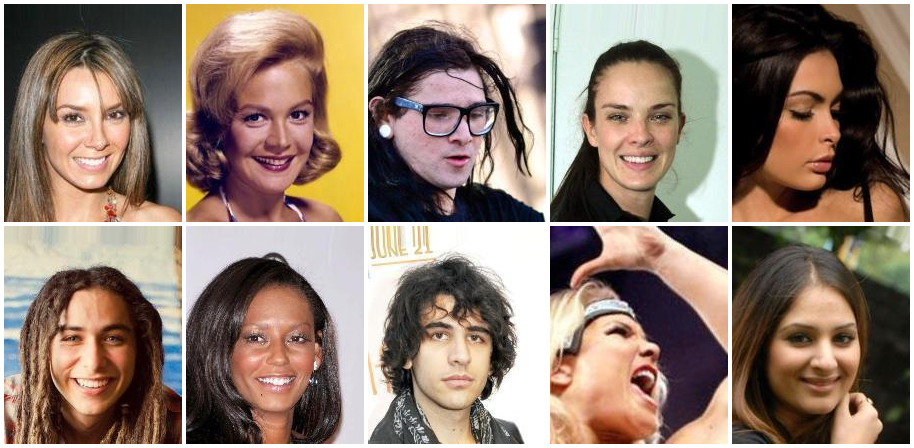}
        \caption{CelebA}
        \label{fig:celeba_examples}
    \end{subfigure}
    \caption{Example images from UTKFace (left) and CelebA (right). UTKFace's crops are tight to the face, leaving little to no visible neck or hairline, so concepts like \textit{Bald}, \textit{Necklace}, and \textit{Receding Hairline} are frequently absent from the frame entirely. CelebA's native crops are looser, consistently including hair, neck, and shoulders, so these same concepts remain visible.}
    \label{fig:dataset_crop_examples}
\end{figure}

\textit{HnS}'s agreement with the ground truth declines when retrieving from an out-of-distribution pool: the average correlation drops from 0.615 on CelebA to 0.571 on LFW and 0.463 on UTKFace. Even so, both out-of-distribution variants remain clearly ahead of every baseline; specifically, \textit{HnS} (UTKFace) still comfortably outperforms TCAV, the strongest baseline (0.463 vs.\ 0.319). \textit{HnS}'s own retrieval pool matters for its absolute performance, but its advantage over existing methods is not an artifact of sharing a dataset with the ground truth.

One thing worth noting is that for UTK Face, the images are closely cropped to the face. The crops are so close in fact that concepts like \textit{Bald}, \textit{Necklace}, \textit{Wearing Hat}, etc., are often not present in the images themselves. As such, when attributing to such concepts, UTK-Face is at a natural disadvantage. Below are some example images from UTK-Face.

%% file: sections/appendices/local_attribution.tex
\section{Local Attribution across Concepts}
\label{appendix:Local_Attribution}

For a representative sample of concepts, we rank images by $|\mathrm{score}|$ and show the three highest-ranked (\textbf{Top-3 Present}) alongside the three lowest-ranked (\textbf{Top-3 Absent}), for both \textit{HnS} and TCAV, following the format introduced in Section~\ref{subsec:local_attribution_experiment}. We show 10 concepts for CelebA's Male task, and 10 (task class, concept) pairs for COCO-Stuff.

\begin{figure}[htbp]
    \centering
    \begin{subfigure}[b]{0.48\linewidth}
        \includegraphics[width=\linewidth]{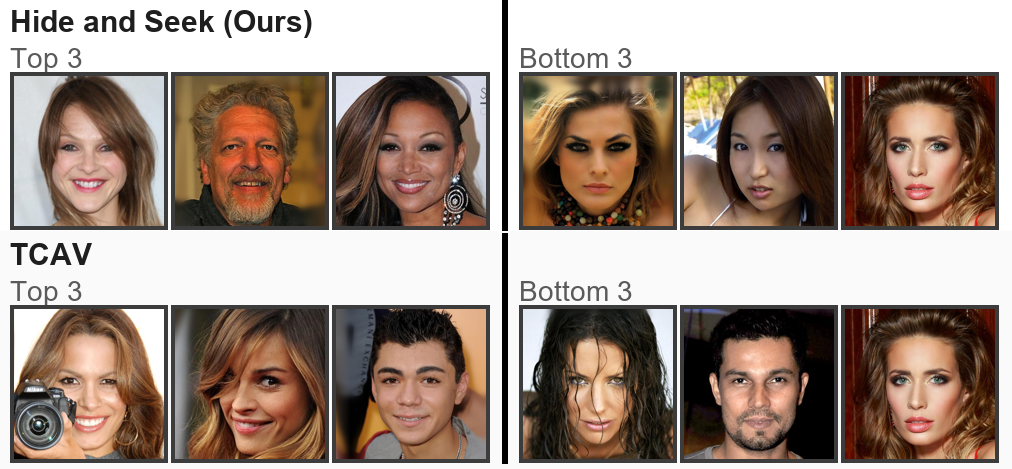}
        \caption{Smiling}
    \end{subfigure}
    \hfill
    \begin{subfigure}[b]{0.48\linewidth}
        \includegraphics[width=\linewidth]{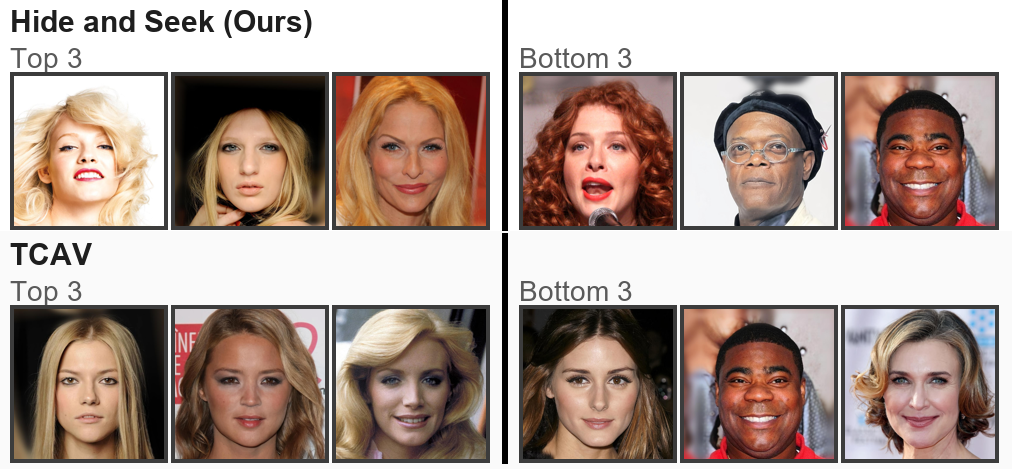}
        \caption{Blond Hair}
    \end{subfigure}
    \caption{Top-3 (highest $|\mathrm{score}|$) and bottom-3 (lowest $|\mathrm{score}|$) Male-task images for HnS and TCAV, for each concept shown.}
    \label{fig:local_attribution_sample_Mouth_Slightly_Open_Wearing_Necklace}
\end{figure}

\begin{figure}[htbp]
    \centering
    \begin{subfigure}[b]{0.48\linewidth}
        \includegraphics[width=\linewidth]{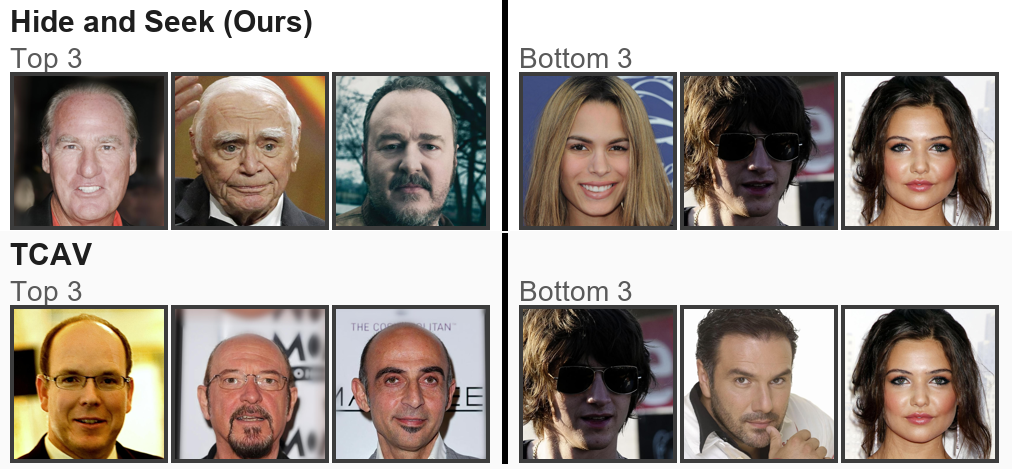}
        \caption{Bald}
    \end{subfigure}
    \hfill
    \begin{subfigure}[b]{0.48\linewidth}
        \includegraphics[width=\linewidth]{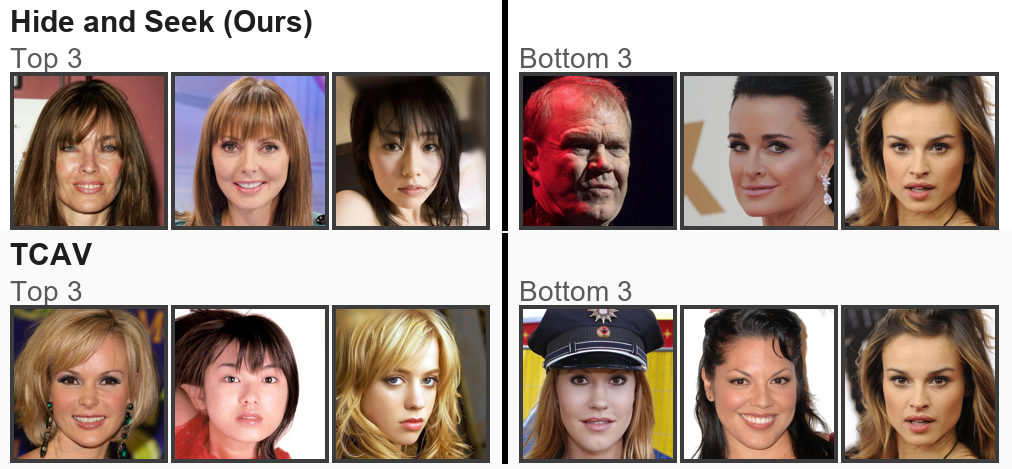}
        \caption{Bangs}
    \end{subfigure}
    \caption{Top-3 (highest $|\mathrm{score}|$) and bottom-3 (lowest $|\mathrm{score}|$) Male-task images for HnS and TCAV, for each concept shown.}
    \label{fig:local_attribution_sample_Narrow_Eyes_Bags_Under_Eyes}
\end{figure}

\begin{figure}[htbp]
    \centering
    \begin{subfigure}[b]{0.48\linewidth}
        \includegraphics[width=\linewidth]{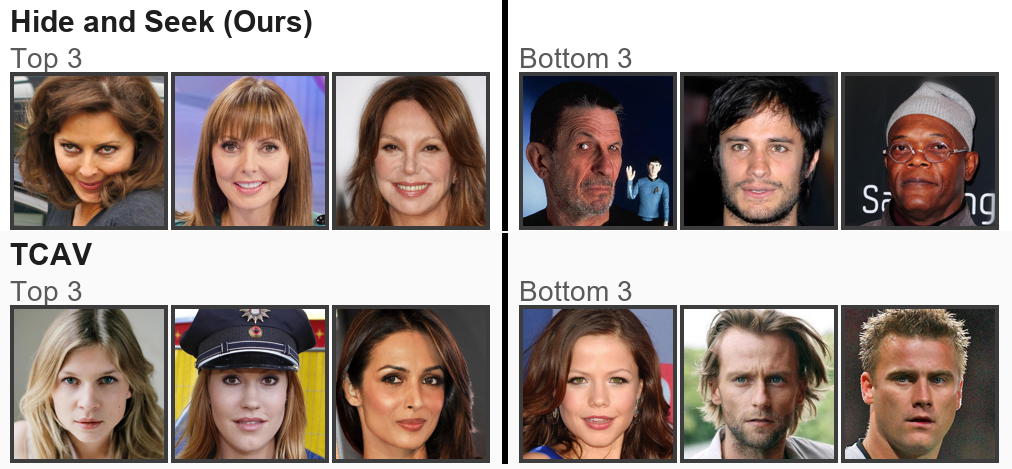}
        \caption{Brown Hair}
    \end{subfigure}
    \hfill
    \begin{subfigure}[b]{0.48\linewidth}
        \includegraphics[width=\linewidth]{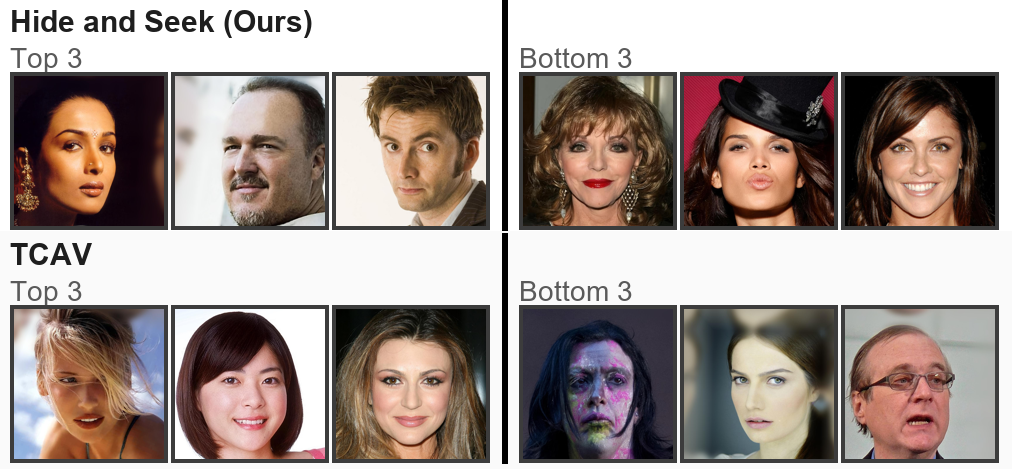}
        \caption{Receding Hairline}
    \end{subfigure}
    \caption{Top-3 (highest $|\mathrm{score}|$) and bottom-3 (lowest $|\mathrm{score}|$) Male-task images for \textit{HnS} and TCAV, for each concept shown.}
    \label{fig:local_attribution_sample_Brown_Hair_Receding_Hairline}
\end{figure}

\begin{figure}[htbp]
    \centering
    \begin{subfigure}[b]{0.48\linewidth}
        \includegraphics[width=\linewidth]{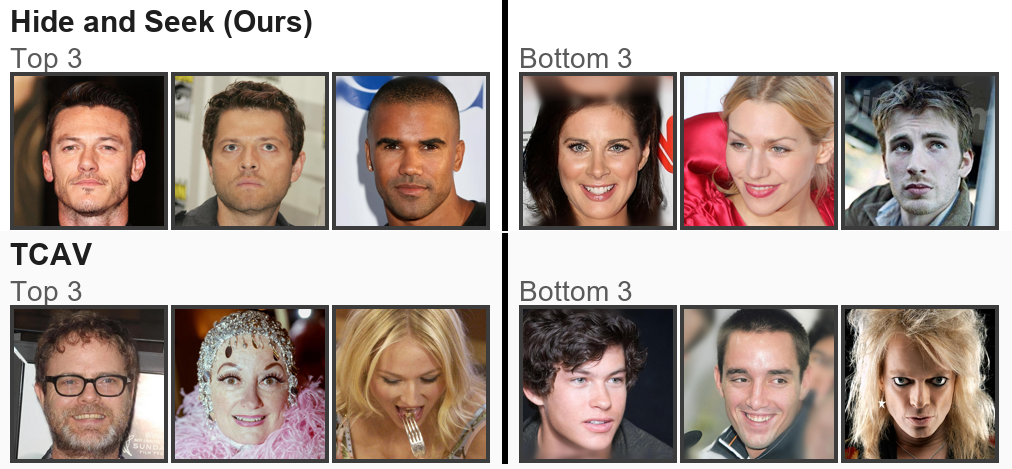}
        \caption{Pointy Nose}
    \end{subfigure}
    \hfill
    \begin{subfigure}[b]{0.48\linewidth}
        \includegraphics[width=\linewidth]{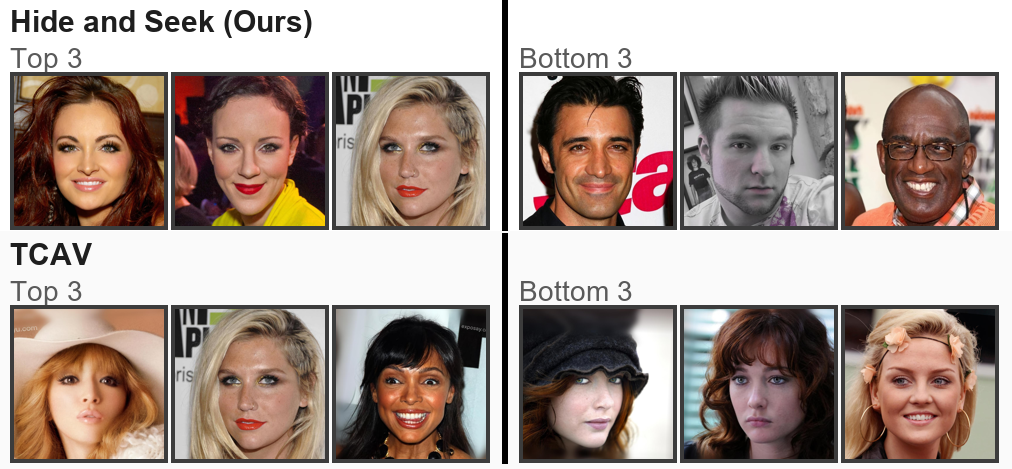}
        \caption{Wearing Lipstick}
    \end{subfigure}
    \caption{Top-3 (highest $|\mathrm{score}|$) and bottom-3 (lowest $|\mathrm{score}|$) Male-task images for \textit{HnS} and TCAV, for each concept shown.}
    \label{fig:local_attribution_sample_Pointy_Nose_Wearing_Necktie}
\end{figure}

\begin{figure}[htbp]
    \centering
    \begin{subfigure}[b]{0.48\linewidth}
        \includegraphics[width=\linewidth]{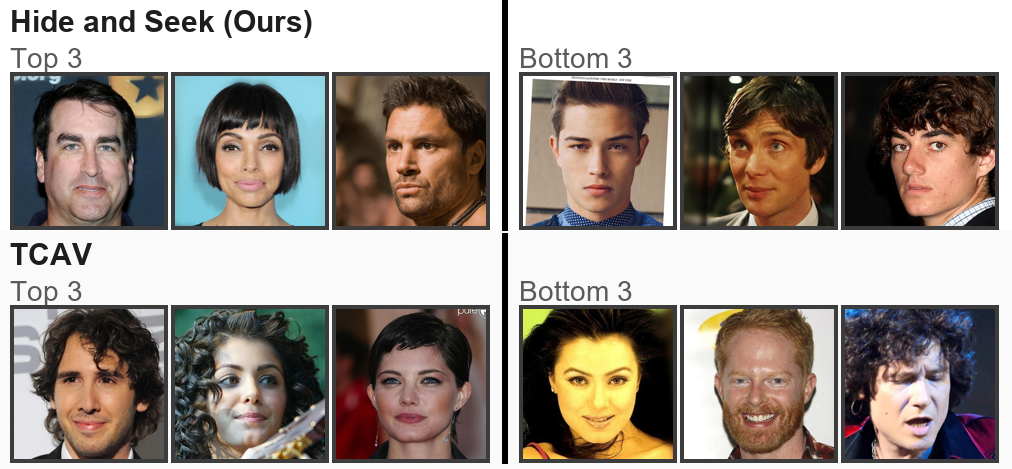}
        \caption{Bushy Eyebrows}
    \end{subfigure}
    \hfill
    \begin{subfigure}[b]{0.48\linewidth}
        \includegraphics[width=\linewidth]{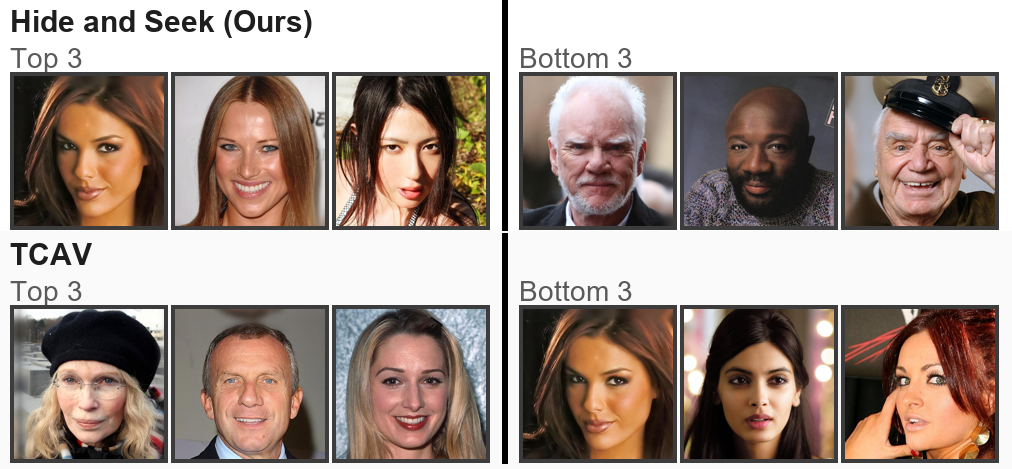}
        \caption{Straight Hair}
    \end{subfigure}
    \caption{Top-3 (highest $|\mathrm{score}|$) and bottom-3 (lowest $|\mathrm{score}|$) Male-task images for \textit{HnS} and TCAV, for each concept shown.}
    \label{fig:local_attribution_sample_Bushy_Eyebrows_Straight_Hair}
\end{figure}

\begin{figure}[htbp]
    \centering
    \begin{subfigure}[b]{0.48\linewidth}
        \includegraphics[width=\linewidth]{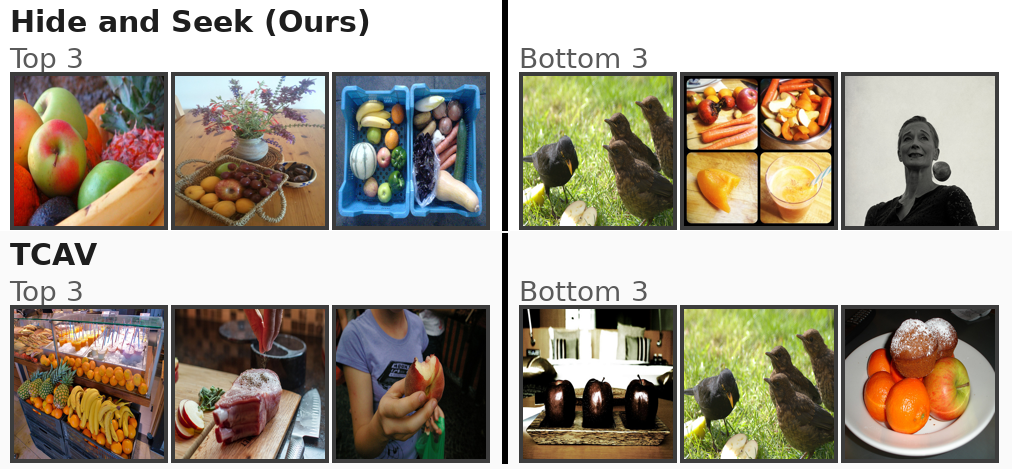}
        \caption{apple / basket}
    \end{subfigure}
    \hfill
    \begin{subfigure}[b]{0.48\linewidth}
        \includegraphics[width=\linewidth]{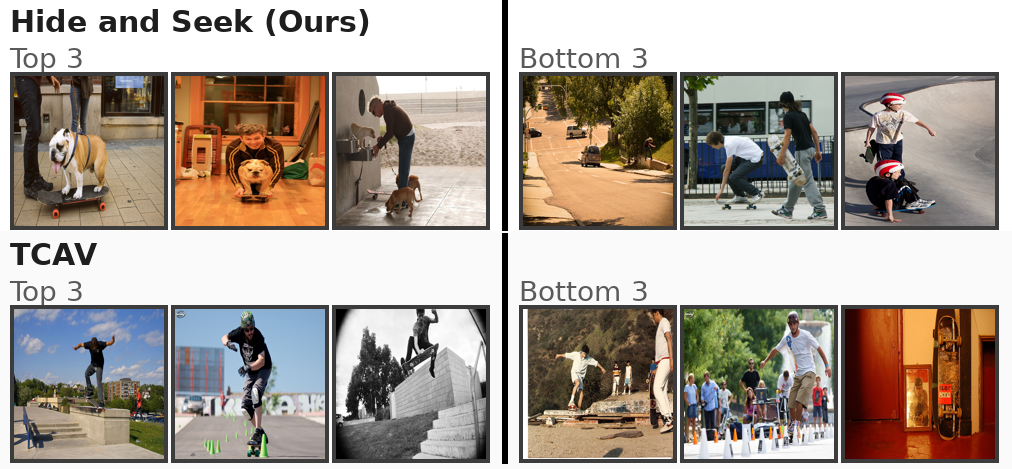}
        \caption{skateboard / dog}
    \end{subfigure}
    \caption{Top-3 (highest $|\mathrm{score}|$) and bottom-3 (lowest $|\mathrm{score}|$) images for \textit{HnS} and TCAV, for each (task class, concept) pair shown.}
    \label{fig:local_attribution_coco_apple_basket_black}
\end{figure}

\begin{figure}[htbp]
    \centering
    \begin{subfigure}[b]{0.48\linewidth}
        \includegraphics[width=\linewidth]{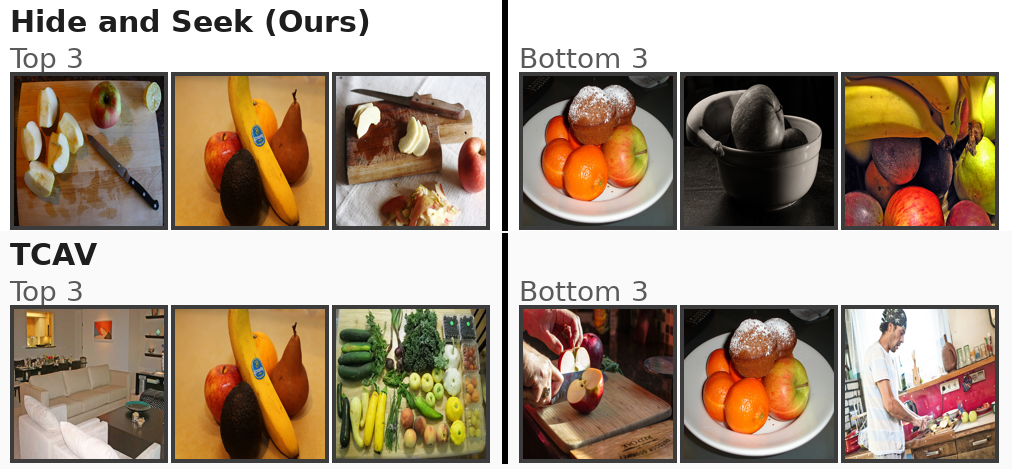}
        \caption{apple / board}
    \end{subfigure}
    \hfill
    \begin{subfigure}[b]{0.48\linewidth}
        \includegraphics[width=\linewidth]{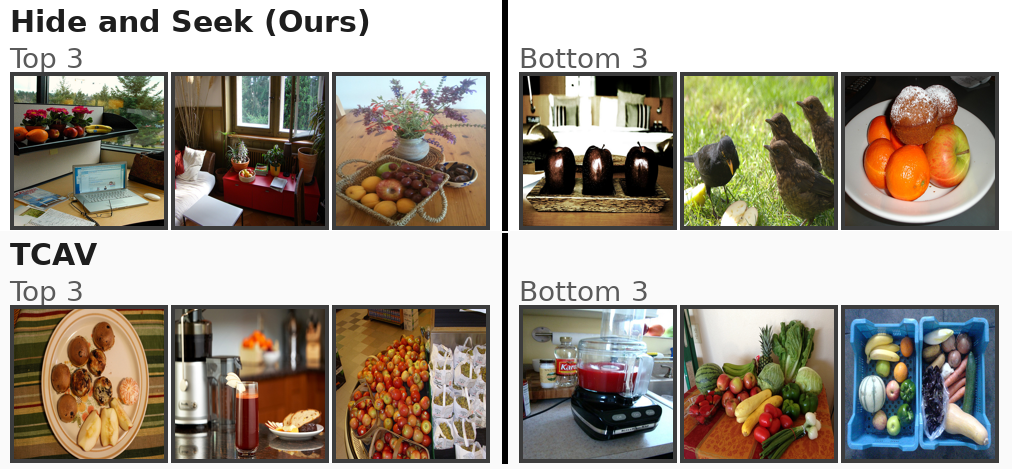}
        \caption{apple / flowerpot}
    \end{subfigure}
    \caption{Top-3 (highest $|\mathrm{score}|$) and bottom-3 (lowest $|\mathrm{score}|$) images for \textit{HnS} and TCAV, for each (task class, concept) pair shown.}
    \label{fig:local_attribution_coco_apple_board_flowerpot}
\end{figure}

\begin{figure}[htbp]
    \centering
    \begin{subfigure}[b]{0.48\linewidth}
        \includegraphics[width=\linewidth]{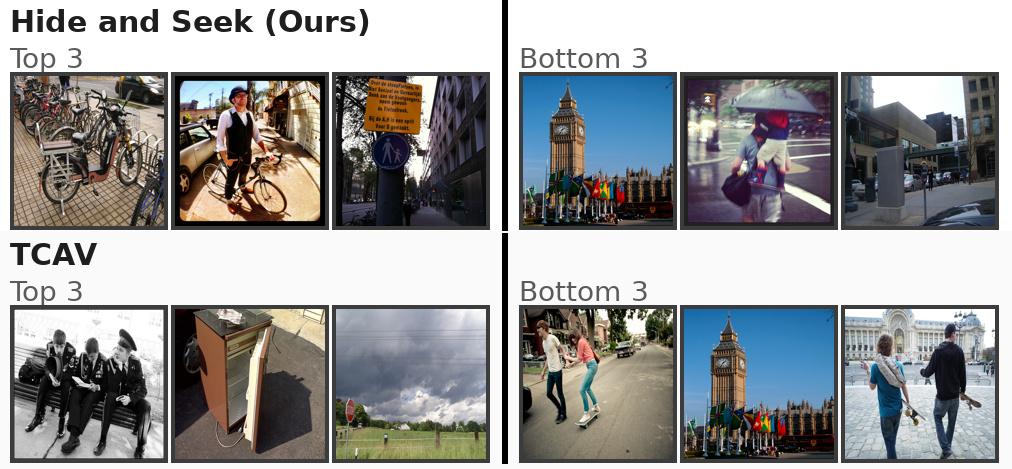}
        \caption{car / bicycle}
    \end{subfigure}
    \hfill
    \begin{subfigure}[b]{0.48\linewidth}
        \includegraphics[width=\linewidth]{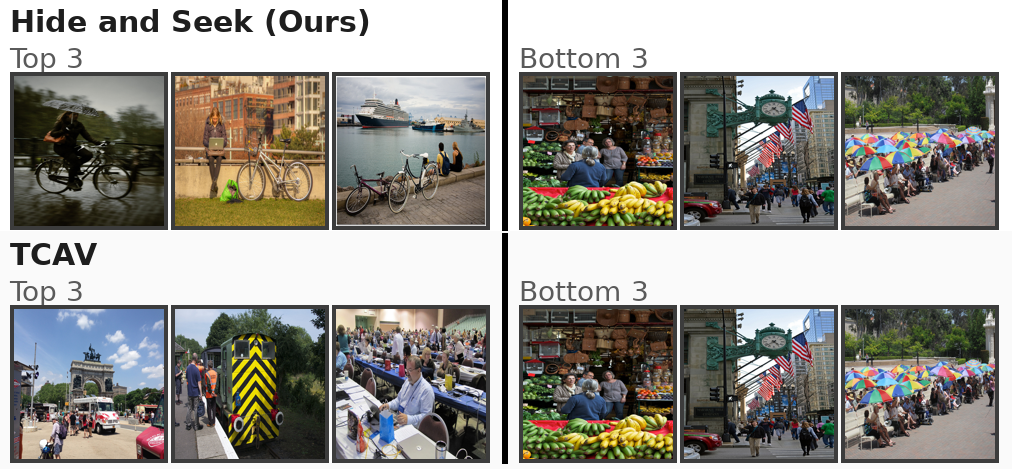}
        \caption{handbag / bicycle}
    \end{subfigure}
    \caption{Top-3 (highest $|\mathrm{score}|$) and bottom-3 (lowest $|\mathrm{score}|$) images for \textit{HnS} and TCAV, for each (task class, concept) pair shown.}
    \label{fig:local_attribution_coco_car_bicycle_handbag_bicycle}
\end{figure}

\begin{figure}[htbp]
    \centering
    \begin{subfigure}[b]{0.48\linewidth}
        \includegraphics[width=\linewidth]{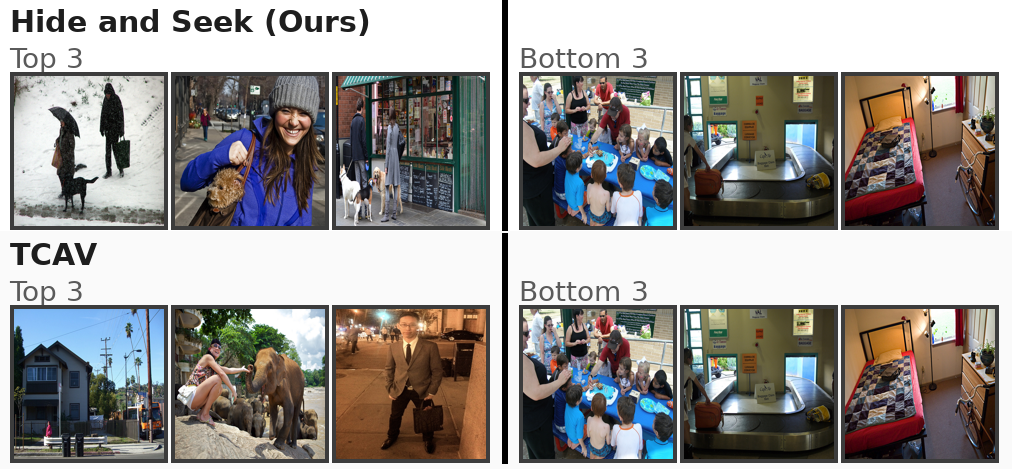}
        \caption{handbag / dog}
    \end{subfigure}
    \hfill
    \begin{subfigure}[b]{0.48\linewidth}
        \includegraphics[width=\linewidth]{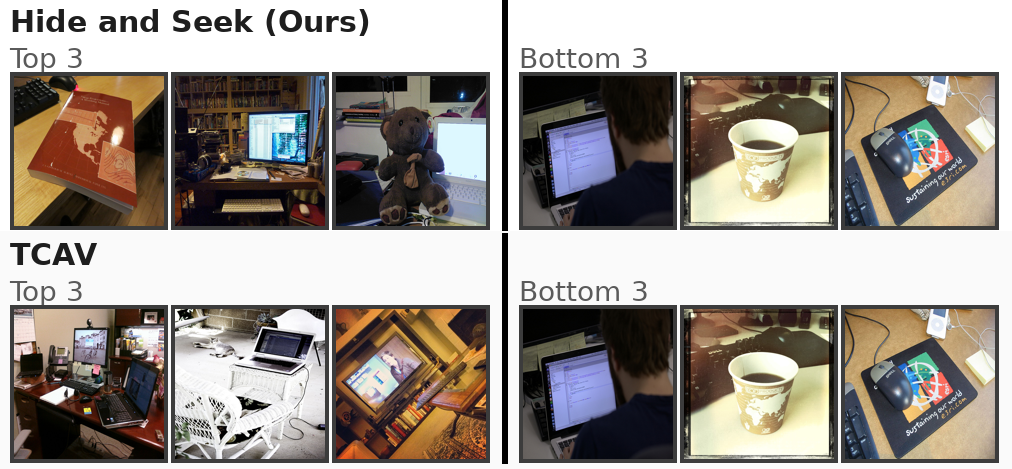}
        \caption{keyboard / book}
    \end{subfigure}
    \caption{Top-3 (highest $|\mathrm{score}|$) and bottom-3 (lowest $|\mathrm{score}|$) images for \textit{HnS} and TCAV, for each (task class, concept) pair shown.}
    \label{fig:local_attribution_coco_handbag_dog_keyboard_book}
\end{figure}

\begin{figure}[htbp]
    \centering
    \begin{subfigure}[b]{0.48\linewidth}
        \includegraphics[width=\linewidth]{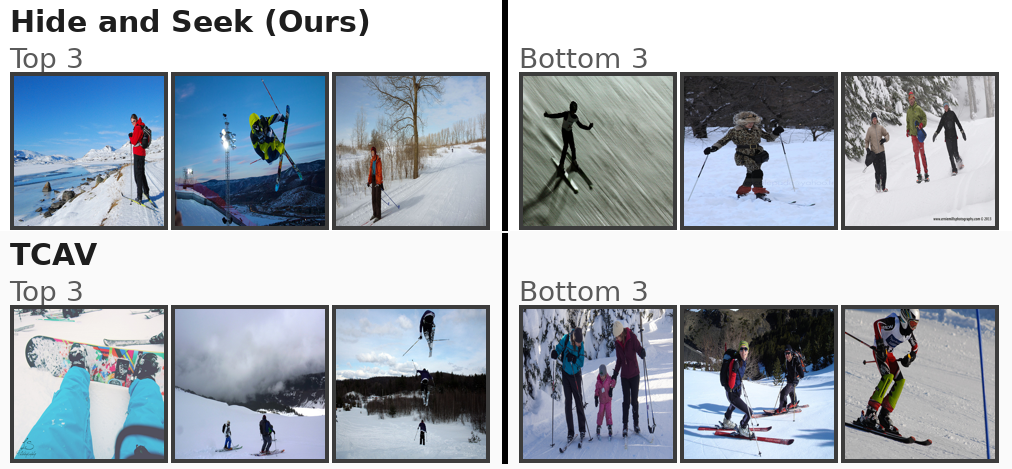}
        \caption{skis / blue}
    \end{subfigure}
    \hfill
    \begin{subfigure}[b]{0.48\linewidth}
        \includegraphics[width=\linewidth]{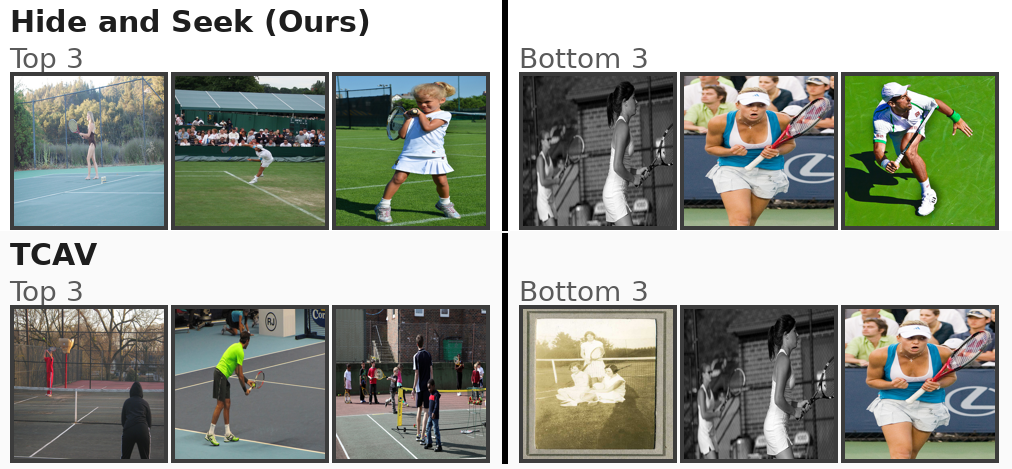}
        \caption{tennis racket / field}
    \end{subfigure}
    \caption{Top-3 (highest $|\mathrm{score}|$) and bottom-3 (lowest $|\mathrm{score}|$) images for \textit{HnS} and TCAV, for each (task class, concept) pair shown.}
    \label{fig:local_attribution_coco_skis_blue_tennis_racket_field}
\end{figure}

%% file: sections/appendices/captioning_exp_details.tex
\section{Captioning Experiment Details}
\label{appendix:Captioning_Experiment}

\subsection{Gender Word Set}

To score a caption's gender lean without any masking or intervention, we sum each caption model's token probabilities over a fixed set of masculine- and feminine-coded words (Section~\ref{subsec:image_captioning}). We reuse the gender word lists from ~\citep{hirota2022quantifying} rather than defining a new list. 

The masculine word list is \{man, men, male, father, gentleman, boy, uncle, husband, actor, prince, waiter, he, his, him\}, and the corresponding feminine word list is \{woman, women, female, mother, lady, girl, aunt, wife, actress, princess, waitress, she, her, hers\}.

\subsection{Bias Breakdown}

The pooled correlation results in Section~\ref{subsec:image_captioning} establish that our attribution scores relate to independent bias measures on average, across all 10 caption models and 79 concepts. To make this more concrete, we look at individual concepts: for each concept, we check whether models agree on its attribution sign (masculine- or feminine-leaning), then identify the masculine-leaning and feminine-leaning concepts with the strongest cross-model consensus.

\begin{figure}[htbp]
    \center
    \includegraphics[width=0.48\linewidth]{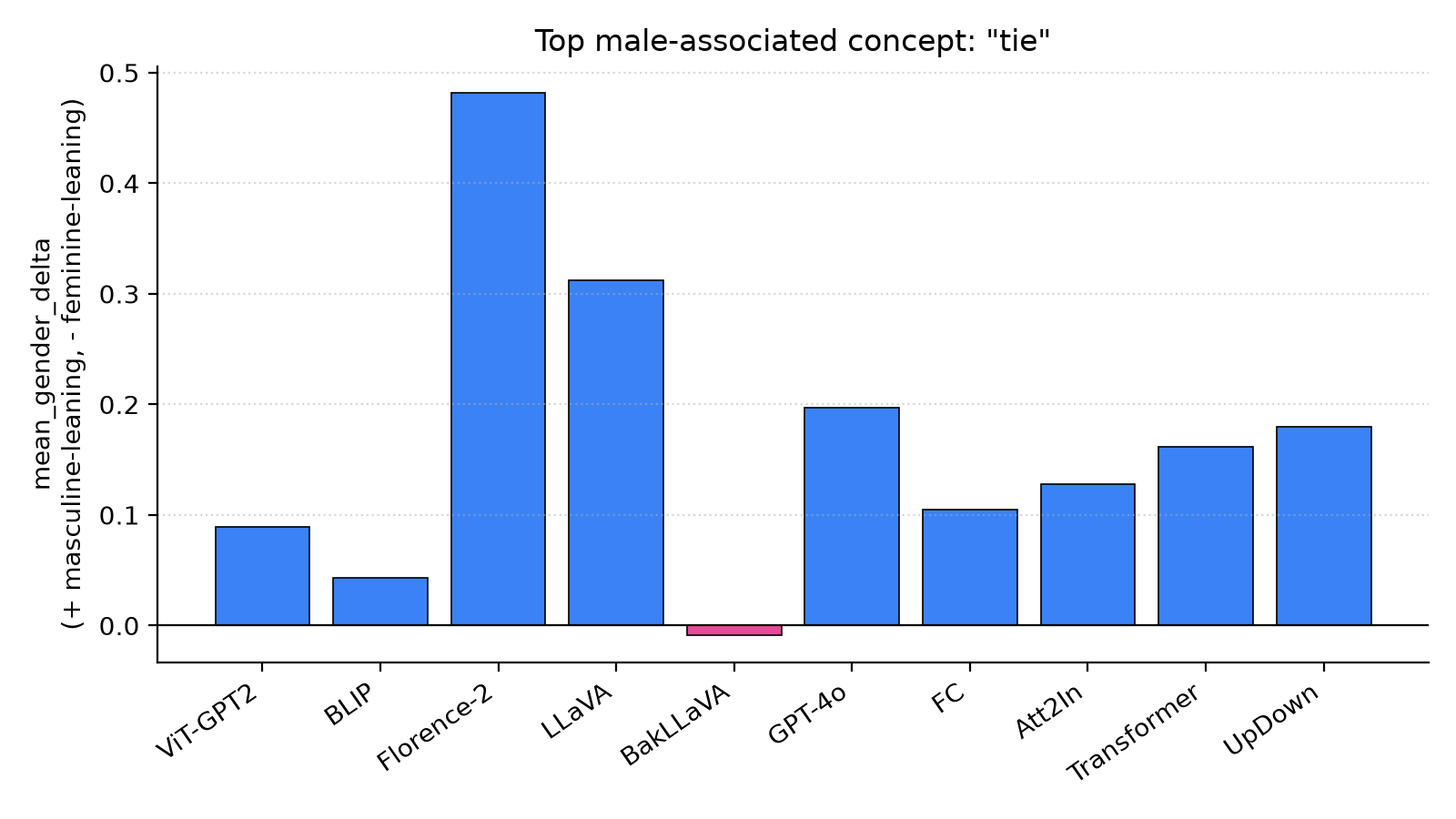}
    \includegraphics[width=0.48\linewidth]{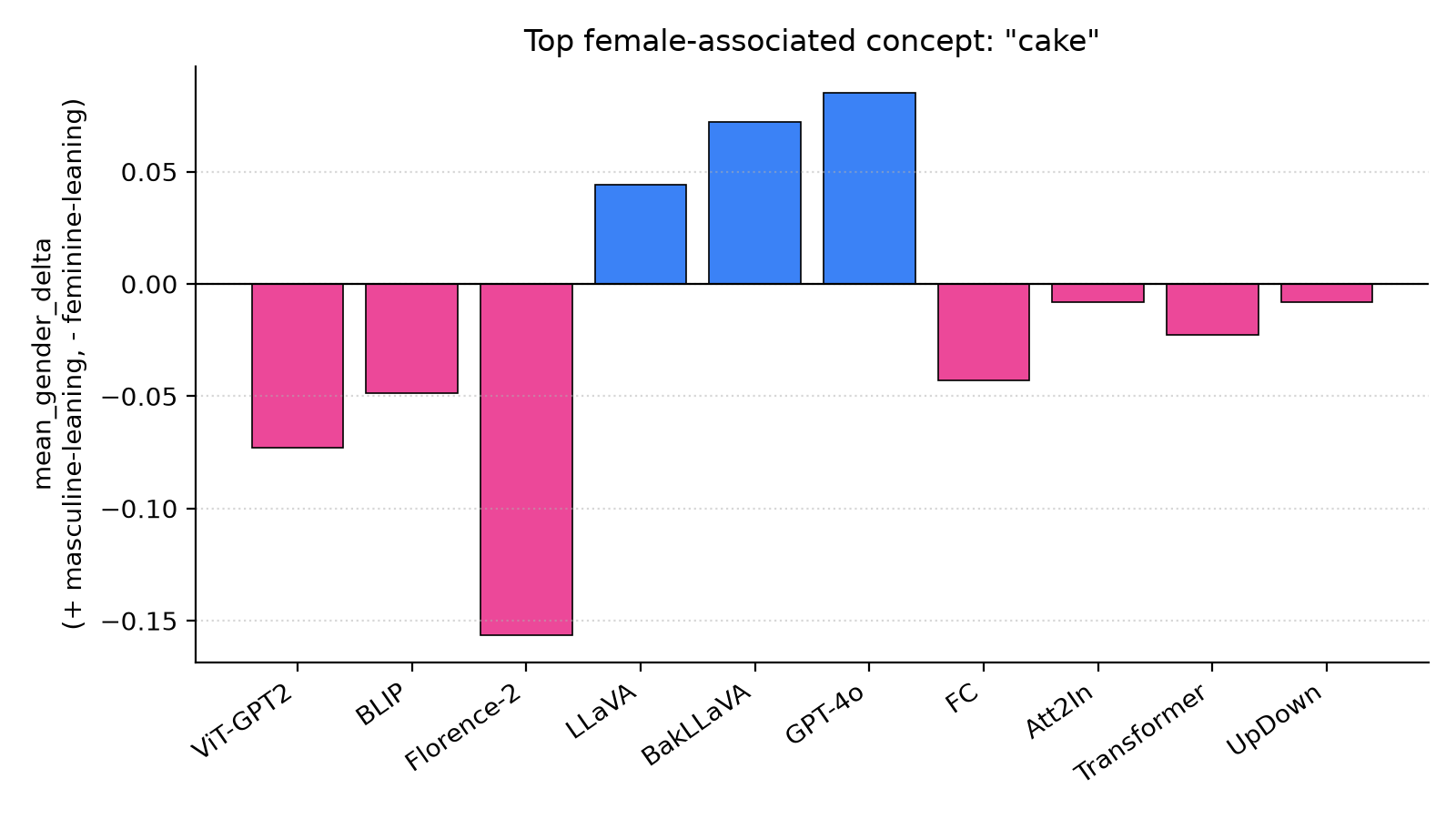}
    \caption{Per-model attribution for the top consensus masculine-leaning concept (\textit{tie}, left) and top consensus feminine-leaning concept (\textit{cake}, right). Bar color shows sign (blue = masculine-leaning, pink = feminine-leaning). 9/10 models agree on \textit{tie}'s masculine direction (BakLLaVA is the lone, near-zero exception); 7/10 models agree on \textit{cake}'s feminine direction, with the 3 disagreeing models (LLaVA, BakLLaVA, GPT-4o) all showing a positive value instead.}
    \label{fig:captioning_top_concepts}
\end{figure}

Figure~\ref{fig:captioning_top_concepts} shows the per-model attribution for these two concepts, \textit{tie} (top consensus masculine-leaning) and \textit{cake} (top consensus feminine-leaning). The consensus is strong but not universal: 9 of 10 models agree \textit{tie} is masculine-leaning, with BakLLaVA the lone exception, and its attribution there is close to zero rather than reversed. For \textit{cake}, 7 of 10 models agree on the feminine-leaning direction, while the 3 disagreeing models (LLaVA, BakLLaVA, GPT-4o) all attribute a positive (masculine-leaning) value instead. This pattern, most models agreeing on a concept's gender lean with a small minority disagreeing, rather than the vote being evenly split, is what we would expect if the attribution scores are picking up a real, shared bias across models.

%% file: sections/appendices/FTW_county_level.tex
\section{FTW Country Level}
\label{appendix:FTW_country_level}

The field segmentation experiment (Section~\ref{subsec:ftw}) uses the whole FTW dataset as the retrieval pool. This can hide real variation between sub-populations: a concept the model relies on heavily in one country may matter little in another. To check this, we break the same attribution down by country rather than pooling it.

To ensure diversity in the image database, we select countries with at least 2,000 observations. For each country with at least 2,000 matched dual-window image pairs (Austria, Corsica, Croatia, Denmark, Estonia, Finland, France, Latvia, Lithuania, Netherlands, Slovakia, Slovenia, Spain, and Sweden), we rerun the Concept Mask using the images of a single country as the retrieval pool.

\begin{figure*}[htbp]
    \centering
    \includegraphics[width=\textwidth]{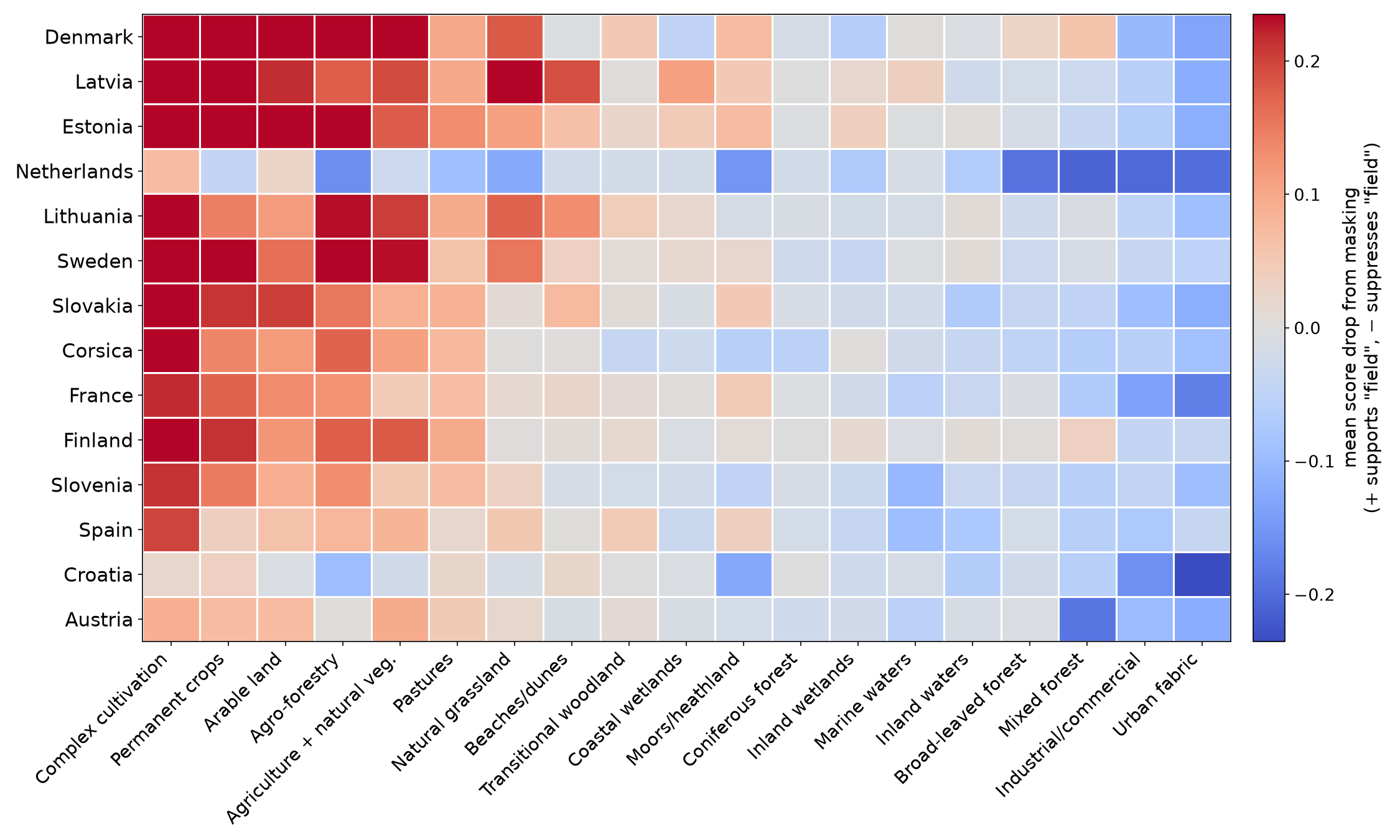}
    \caption{Hide and Seek's land-cover concept attributions for the FTW field-segmentation model, across the 14 countries with sufficient retrieved locations. Each cell is the mean drop in the model's
    predicted field score when that concept's most-associated region is masked (normalized by 255), colored on a diverging scale: red indicates the concept supports a ``field'' prediction (e.g., Arable land), blue indicates it suppresses one (e.g., Urban fabric). Concepts are ordered left to right by their mean effect across countries; countries are ordered top to bottom by mean $|\text{effect}|$. The agriculture-supports/urban-suppresses gradient holds consistently across most countries.}
    \label{fig:ftw_country_heatmap}
\end{figure*}

Figure~\ref{fig:ftw_country_heatmap} shows the resulting country-by-concept attribution grid. A consistent pattern holds across nearly every country: agricultural concepts (\textit{Arable land}, \textit{Permanent crops}, \textit{Complex cultivation}) support the field prediction, while built and forested concepts (\textit{Urban fabric}, \textit{Industrial or commercial units}, \textit{Mixed forest}) suppress it. This consistency across 14 countries is itself evidence that Concept Mask's attributions reflect a real property of the model rather than an artifact specific to one country's data.

%% file: iclr2027_conference.bib
@inproceedings{Score_CAM,
  title={Score-CAM: Score-weighted visual explanations for convolutional neural networks},
  author={Wang, Haofan and Wang, Zifan and Du, Mengnan and Yang, Fan and Zhang, Zijian and Ding, Sirui and Mardziel, Piotr and Hu, Xia},
  booktitle={Proceedings of the IEEE/CVF conference on computer vision and pattern recognition workshops},
  pages={24--25},
  year={2020}
}

@inproceedings{IntegratedGradients,
  title={Axiomatic attribution for deep networks},
  author={Sundararajan, Mukund and Taly, Ankur and Yan, Qiqi},
  booktitle={International conference on machine learning},
  pages={3319--3328},
  year={2017},
  organization={PMLR}
}

@inproceedings{TCAV,
  title={Interpretability beyond feature attribution: Quantitative testing with concept activation vectors (tcav)},
  author={Kim, Been and Wattenberg, Martin and Gilmer, Justin and Cai, Carrie and Wexler, James and Viegas, Fernanda and others},
  booktitle={International conference on machine learning},
  pages={2668--2677},
  year={2018},
  organization={PMLR}
}

@inproceedings{PACE,
  title={Probabilistic Conceptual Explainers: Trustworthy Conceptual Explanations for Vision Foundation Models},
  author={Wang, Hengyi and Tan, Shiwei and Wang, Hao},
  booktitle={International Conference on Machine Learning},
  pages={51502--51522},
  year={2024},
  organization={PMLR}
}

@article{ConceptSHAP,
  title={On completeness-aware concept-based explanations in deep neural networks},
  author={Yeh, Chih-Kuan and Kim, Been and Arik, Sercan and Li, Chun-Liang and Pfister, Tomas and Ravikumar, Pradeep},
  journal={Advances in neural information processing systems},
  volume={33},
  pages={20554--20565},
  year={2020}
}

@article{conceptWhiten,
  title={Concept whitening for interpretable image recognition},
  author={Chen, Zhi and Bei, Yijie and Rudin, Cynthia},
  journal={Nature Machine Intelligence},
  volume={2},
  number={12},
  pages={772--782},
  year={2020},
  publisher={Nature Publishing Group UK London}
}

@inproceedings{conceptGradient,
  title={Concept Gradient: Concept-based Interpretation Without Linear Assumption.},
  author={Bai, Andrew and Yeh, Chih-Kuan and Ravikumar, Pradeep and Lin, YC and Hsieh, Cho-Jui},
  booktitle={International Conference on Learning Representation},
  year={2023},
  organization={ICLR}
}

@inproceedings{ConeSHAP,
author = {Li, Jiahui and Kuang, Kun and Li, Lin and Chen, Long and Zhang, Songyang and Shao, Jian and Xiao, Jun},
title = {Instance-wise or Class-wise? A Tale of Neighbor Shapley for Concept-based Explanation},
year = {2021},
isbn = {9781450386517},
publisher = {Association for Computing Machinery},
address = {New York, NY, USA},
url = {https://doi.org/10.1145/3474085.3475337},
doi = {10.1145/3474085.3475337},
booktitle = {Proceedings of the 29th ACM International Conference on Multimedia},
pages = {3664–3672},
numpages = {9},
location = {Virtual Event, China},
series = {MM '21}
}

@InProceedings{CBM,
  title = 	 {Concept Bottleneck Models},
  author =       {Koh, Pang Wei and Nguyen, Thao and Tang, Yew Siang and Mussmann, Stephen and Pierson, Emma and Kim, Been and Liang, Percy},
  booktitle = 	 {Proceedings of the 37th International Conference on Machine Learning},
  pages = 	 {5338--5348},
  year = 	 {2020},
  editor = 	 {III, Hal Daumé and Singh, Aarti},
  volume = 	 {119},
  series = 	 {Proceedings of Machine Learning Research},
  month = 	 {13--18 Jul},
  publisher =    {PMLR},
  url = 	 {https://proceedings.mlr.press/v119/koh20a.html}
}

@inproceedings{PCBM-h,
  title={Post-hoc Concept Bottleneck Models},
  author={Yuksekgonul, Mert and Wang, Maggie and Zou, James},
  booktitle={The Eleventh International Conference on Learning Representations},
  year={2023}
}

@misc{tapli2026rethinkingconceptbottleneckmodels,
      title={Rethinking Concept Bottleneck Models: From Pitfalls to Solutions}, 
      author={Merve Tapli and Quentin Bouniot and Wolfgang Stammer and Zeynep Akata and Emre Akbas},
      year={2026},
      eprint={2603.05629},
      archivePrefix={arXiv},
      primaryClass={cs.CV},
      url={https://arxiv.org/abs/2603.05629}, 
}

@inproceedings{ProbabilisticCBMs,
  title={Probabilistic Concept Bottleneck Models},
  author={Kim, Eunji and Jung, Dahuin and Park, Sangha and Kim, Siwon and Yoon, Sungroh},
  booktitle={International Conference on Machine Learning},
  pages={16521--16540},
  year={2023},
  organization={PMLR}
}

@article{StochasticCBMs,
  title={Stochastic concept bottleneck models},
  author={Vandenhirtz, Moritz and Laguna, Sonia and Marcinkevi{\v{c}}s, Ri{\v{c}}ards and Vogt, Julia E},
  journal={Advances in Neural Information Processing Systems},
  volume={37},
  pages={51787--51810},
  year={2024}
}

@inproceedings{CLIP,
  title={Learning transferable visual models from natural language supervision},
  author={Radford, Alec and Kim, Jong Wook and Hallacy, Chris and Ramesh, Aditya and Goh, Gabriel and Agarwal, Sandhini and Sastry, Girish and Askell, Amanda and Mishkin, Pamela and Clark, Jack and others},
  booktitle={International conference on machine learning},
  pages={8748--8763},
  year={2021},
  organization={PmLR}
}

@inproceedings{levi2025double,
  title={The Double-Ellipsoid Geometry of CLIP},
  author={Levi, Meir Yossef and Gilboa, Guy},
  booktitle={International Conference on Machine Learning},
  pages={33999--34019},
  year={2025},
  organization={PMLR}
}

@inproceedings{time,
  title={Text-to-image models for counterfactual explanations: a black-box approach},
  author={Jeanneret, Guillaume and Simon, Lo{\"\i}c and Jurie, Fr{\'e}d{\'e}ric},
  booktitle={Proceedings of the IEEE/CVF Winter Conference on Applications of Computer Vision},
  pages={4757--4767},
  year={2024}
}

@inproceedings{abid2022meaningfully,
  title={Meaningfully debugging model mistakes using conceptual counterfactual explanations},
  author={Abid, Abubakar and Yuksekgonul, Mert and Zou, James},
  booktitle={International Conference on Machine Learning},
  pages={66--88},
  year={2022},
  organization={PMLR}
}

@inproceedings{jeanneret2023adversarial,
  title={Adversarial counterfactual visual explanations},
  author={Jeanneret, Guillaume and Simon, Lo{\"\i}c and Jurie, Fr{\'e}d{\'e}ric},
  booktitle={Proceedings of the IEEE/CVF Conference on Computer Vision and Pattern Recognition},
  pages={16425--16435},
  year={2023}
}

@inproceedings{jeanneret2022diffusion,
  title={Diffusion models for counterfactual explanations},
  author={Jeanneret, Guillaume and Simon, Lo{\"\i}c and Jurie, Fr{\'e}d{\'e}ric},
  booktitle={Proceedings of the Asian conference on computer vision},
  pages={858--876},
  year={2022}
}

@inproceedings{DPA,
title={{DPA}: A one-stop metric to measure bias amplification in classification datasets},
author={Bhanu Tokas and Rahul Nair and Hannah Kerner},
booktitle={The Thirty-ninth Annual Conference on Neural Information Processing Systems},
year={2025},
url={https://openreview.net/forum?id=efuY2yYC1U}
}

@InProceedings{DBAC,
    author    = {Nair, Rahul and Tokas, Bhanu and Kerner, Hannah},
    title     = {A Woman with a Knife or A Knife with a Woman? Measuring Directional Bias Amplification in Image Captions},
    booktitle = {Proceedings of the IEEE/CVF Winter Conference on Applications of Computer Vision (WACV)},
    month     = {March},
    year      = {2026},
    pages     = {255-264}
}

@inproceedings{tang2021mitigating,
  title={Mitigating gender bias in captioning systems},
  author={Tang, Ruixiang and Du, Mengnan and Li, Yuening and Liu, Zirui and Zou, Na and Hu, Xia},
  booktitle={Proceedings of the Web Conference 2021},
  pages={633--645},
  year={2021}
}

@inproceedings{hirota2022quantifying,
  title={Quantifying societal bias amplification in image captioning},
  author={Hirota, Yusuke and Nakashima, Yuta and Garcia, Noa},
  booktitle={2022 IEEE/CVF Conference on Computer Vision and Pattern Recognition (CVPR)},
  pages={13440--13449},
  year={2022},
  organization={IEEE}
}

@inproceedings{lee2020maskgan, 
  title={Maskgan: Towards diverse and interactive facial image manipulation},
  author={Lee, Cheng-Han and Liu, Ziwei and Wu, Lingyun and Luo, Ping},
  booktitle={2020 IEEE/CVF Conference on Computer Vision and Pattern Recognition (CVPR)},
  pages={5548--5557},
  year={2020},
  organization={IEEE}
}

@inproceedings{singh2020don,
  title={Don’t judge an object by its context: Learning to overcome contextual bias},
  author={Singh, Krishna Kumar and Mahajan, Dhruv and Grauman, Kristen and Lee, Yong Jae and Feiszli, Matt and Ghadiyaram, Deepti},
  booktitle={2020 IEEE/CVF Conference on Computer Vision and Pattern Recognition (CVPR)},
  pages={11067--11075},
  year={2020},
  organization={IEEE}
}

@InProceedings{Broden_filtered,
  title = 	 {Meaningfully debugging model mistakes using conceptual counterfactual explanations},
  author =       {Abid, Abubakar and Yuksekgonul, Mert and Zou, James},
  booktitle = 	 {Proceedings of the 39th International Conference on Machine Learning},
  pages = 	 {66--88},
  year = 	 {2022},
  editor = 	 {Chaudhuri, Kamalika and Jegelka, Stefanie and Song, Le and Szepesvari, Csaba and Niu, Gang and Sabato, Sivan},
  volume = 	 {162},
  series = 	 {Proceedings of Machine Learning Research},
  month = 	 {17--23 Jul},
  publisher =    {PMLR},
  url = 	 {https://proceedings.mlr.press/v162/abid22a.html}
}

@inproceedings{Broden,
  title={Net2vec: Quantifying and explaining how concepts are encoded by filters in deep neural networks},
  author={Fong, Ruth and Vedaldi, Andrea},
  booktitle={Proceedings of the IEEE conference on computer vision and pattern recognition},
  pages={8730--8738},
  year={2018}
}

@inproceedings{kerner2025fields,
  title={Fields of the world: A machine learning benchmark dataset for global agricultural field boundary segmentation},
  author={Kerner, Hannah and Chaudhari, Snehal and Ghosh, Aninda and Robinson, Caleb and Ahmad, Adeel and Choi, Eddie and Jacobs, Nathan and Holmes, Chris and Mohr, Matthias and Dodhia, Rahul and others},
  booktitle={Proceedings of the AAAI Conference on Artificial Intelligence},
  volume={39},
  pages={28151--28159},
  year={2025}
}

@article{sumbul2019bigearthnet,
  title={Bigearthnet: A large-scale benchmark archive for remote sensing image understanding},
  author={Sumbul, Gencer and Charfuelan, Marcela and Demir, Beg{\"u}m and Markl, Volker},
  journal={arXiv preprint arXiv:1902.06148},
  year={2019}
}

@inproceedings{zhifei2017cvpr,
  title={Age Progression/Regression by Conditional Adversarial Autoencoder},
  author={Zhang, Zhifei and Song, Yang and Qi, Hairong},
  booktitle={IEEE Conference on Computer Vision and Pattern Recognition (CVPR)},
  year={2017},
  organization={IEEE}
}

@article{rawlekar2025efficiently,
  title={Efficiently Disentangling CLIP for Multi-Object Perception},
  author={Rawlekar, Samyak and Cai, Yujun and Wang, Yiwei and Yang, Ming-Hsuan and Ahuja, Narendra},
  journal={arXiv preprint arXiv:2502.02977},
  year={2025}
}

@inproceedings{li2022blip,
  title={Blip: Bootstrapping language-image pre-training for unified vision-language understanding and generation},
  author={Li, Junnan and Li, Dongxu and Xiong, Caiming and Hoi, Steven},
  booktitle={International conference on machine learning},
  pages={12888--12900},
  year={2022},
  organization={PMLR}
}

@article{achiam2023gpt,
  title={Gpt-4 technical report},
  author={Achiam, Josh and Adler, Steven and Agarwal, Sandhini and Ahmad, Lama and Akkaya, Ilge and Aleman, Florencia Leoni and Almeida, Diogo and Altenschmidt, Janko and Altman, Sam and Anadkat, Shyamal and others},
  journal={arXiv preprint arXiv:2303.08774},
  year={2023}
}

@INPROCEEDINGS{vit_gpt2,
  author={Mishra, Swapneel and Seth, Saumya and Jain, Shrishti and Pant, Vasudev and Parikh, Jolly and Jain, Rachna and Islam, Sardar M.N.},
  booktitle={2024 2nd International Conference on Advancement in Computation \& Computer Technologies (InCACCT)}, 
  title={Image Caption Generation using Vision Transformer and GPT Architecture}, 
  year={2024},
  volume={},
  number={},
  pages={1-6},
  doi={10.1109/InCACCT61598.2024.10551257}
}

@misc{liu2023llava,
      title={Visual Instruction Tuning}, 
      author={Liu, Haotian and Li, Chunyuan and Wu, Qingyang and Lee, Yong Jae},
      publisher={NeurIPS},
      year={2023},
}

@misc{liu2023improvedllava,
      title={Improved Baselines with Visual Instruction Tuning}, 
      author={Liu, Haotian and Li, Chunyuan and Li, Yuheng and Lee, Yong Jae},
      publisher={arXiv:2310.03744},
      year={2023},
}

@article{florence, 
    title={Florence-2: Advancing a unified representation for a variety of vision tasks}, 
    author={Xiao, Bin and Wu, Haiping and Xu, Weijian and Dai, Xiyang and Hu, Houdong and Lu, Yumao and Zeng, Michael and Liu, Ce and Yuan, Lu}, 
    journal={arXiv preprint arXiv:2311.06242}, 
    year={2023} 
}

@inproceedings{rennie2017self,
  title={Self-critical sequence training for image captioning},
  author={Rennie, Steven J and Marcheret, Etienne and Mroueh, Youssef and Ross, Jerret and Goel, Vaibhava},
  booktitle={Proceedings of the IEEE conference on computer vision and pattern recognition},
  pages={7008--7024},
  year={2017}
}

@inproceedings{Attention_is_all_you_need,
 author = {Vaswani, Ashish and Shazeer, Noam and Parmar, Niki and Uszkoreit, Jakob and Jones, Llion and Gomez, Aidan N and Kaiser, \L ukasz and Polosukhin, Illia},
 booktitle = {Advances in Neural Information Processing Systems},
 editor = {I. Guyon and U. Von Luxburg and S. Bengio and H. Wallach and R. Fergus and S. Vishwanathan and R. Garnett},
 pages = {},
 publisher = {Curran Associates, Inc.},
 title = {Attention is All you Need},
 url = {https://proceedings.neurips.cc/paper_files/paper/2017/file/3f5ee243547dee91fbd053c1c4a845aa-Paper.pdf},
 volume = {30},
 year = {2017}
}

@INPROCEEDINGS {UpDn,
    author = {P. Anderson and X. He and C. Buehler and D. Teney and M. Johnson and S. Gould and L. Zhang},
    booktitle = {2018 IEEE/CVF Conference on Computer Vision and Pattern Recognition (CVPR)},
    title = {Bottom-Up and Top-Down Attention for Image Captioning and Visual Question Answering},
    year = {2018},
    volume = {},
    issn = {},
    pages = {6077-6086},
    doi = {10.1109/CVPR.2018.00636},
    url = {https://doi.ieeecomputersociety.org/10.1109/CVPR.2018.00636},
    publisher = {IEEE Computer Society},
    address = {Los Alamitos, CA, USA},
    month = {jun}
}

@inproceedings{resnet,
  title={Deep residual learning for image recognition. 2016 IEEE Conf},
  author={He, Kaiming and Zhang, Xiangyu and Ren, Shaoquing and Sun, Jian},
  booktitle={Comput. Vis. Pattern Recognit},
  pages={770--778},
  year={2015}
}

@misc{muhawenayo2026pruepracticalrecipefield,
      title={PRUE: A Practical Recipe for Field Boundary Segmentation at Scale}, 
      author={Gedeon Muhawenayo and Caleb Robinson and Subash Khanal and Zhanpei Fang and Isaac Corley and Alexander Wollam and Tianyi Gao and Leonard Strnad and Ryan Avery and Lyndon Estes and Ana M. Tárano and Nathan Jacobs and Hannah Kerner},
      year={2026},
      eprint={2603.27101},
      archivePrefix={arXiv},
      primaryClass={cs.CV},
      url={https://arxiv.org/abs/2603.27101}, 
}

@article{ren2024grounded,
  title={Grounded sam: Assembling open-world models for diverse visual tasks},
  author={Ren, Tianhe and Liu, Shilong and Zeng, Ailing and Lin, Jing and Li, Kunchang and Cao, He and Chen, Jiayu and Huang, Xinyu and Chen, Yukang and Yan, Feng and others},
  journal={arXiv preprint arXiv:2401.14159},
  year={2024}
}

@article{petsiuk2018rise,
  title={Rise: Randomized input sampling for explanation of black-box models},
  author={Petsiuk, Vitali and Das, Abir and Saenko, Kate},
  journal={arXiv preprint arXiv:1806.07421},
  year={2018}
}

@article{dosovitskiy2020image,
  title={An image is worth 16x16 words: Transformers for image recognition at scale},
  author={Dosovitskiy, Alexey},
  journal={arXiv preprint arXiv:2010.11929},
  year={2020}
}

@inproceedings{liu2022convnet,
  title={A convnet for the 2020s},
  author={Liu, Zhuang and Mao, Hanzi and Wu, Chao-Yuan and Feichtenhofer, Christoph and Darrell, Trevor and Xie, Saining},
  booktitle={2022 IEEE/CVF conference on computer vision and pattern recognition (CVPR)},
  pages={11966--11976},
  year={2022},
  organization={IEEE}
}

@inproceedings{meister2023gender,
  title={Gender artifacts in visual datasets},
  author={Meister, Nicole and Zhao, Dora and Wang, Angelina and Ramaswamy, Vikram V and Fong, Ruth and Russakovsky, Olga},
  booktitle={2023 IEEE/CVF International Conference on Computer Vision (ICCV)},
  pages={4814--4825},
  year={2023},
  organization={IEEE}
}

@inproceedings{materzynska2022disentangling,
  title={Disentangling visual and written concepts in clip},
  author={Materzy{\'n}ska, Joanna and Torralba, Antonio and Bau, David},
  booktitle={2022 IEEE/CVF Conference on Computer Vision and Pattern Recognition (CVPR)},
  pages={16389--16398},
  year={2022},
  organization={IEEE}
}
